\documentclass{article}
\usepackage{PRIMEarxiv}

\usepackage[utf8]{inputenc}
\usepackage[T1]{fontenc}    

\usepackage{graphicx}%
\usepackage{multirow}%
\usepackage{amsmath,amssymb,amsfonts}%
\usepackage{amsthm}%
\usepackage{mathrsfs}%
\usepackage[title]{appendix}%
\usepackage{xcolor}%
\usepackage{textcomp}%
\usepackage{manyfoot}%
\usepackage{booktabs}%
\usepackage{hyperref}
\usepackage{algorithm}%
\usepackage{nicefrac}       
\usepackage{algorithmicx}%
\usepackage{algpseudocode}%
\usepackage{listings}%
\usepackage{mathtools}
\graphicspath{{media/}}     
\usepackage{lipsum}
\usepackage{fancyhdr}       
\usepackage{tcolorbox}
\usepackage{url}
\usepackage[numbers,sort&compress]{natbib}

\usepackage{tikz} 
\usetikzlibrary{shapes.geometric, arrows, positioning} 

\tikzstyle{mainblock} = [rectangle, draw, fill=black!10, text centered, rounded corners, minimum height=3em]
\tikzstyle{subblock} = [rectangle, draw, fill=orange!30, text centered, rounded corners, minimum height=2.5em]
\tikzstyle{connector} = [draw, -latex']

\usepackage{forest}

\theoremstyle{plain}
\newtheorem{theorem}{Theorem}[section]

\newtheorem{corollary}[theorem]{Corollary}
\theoremstyle{definition}
\newtheorem{definition}[theorem]{Definition}

\theoremstyle{remark}
\newtheorem{remark}[theorem]{Remark}

\title{A Theoretical Survey on Foundation Models}

\author{
Shi Fu \\
Nanyang Technological University\\
University of Science and Technology of China \\
fs311@mail.ustc.edu.cn
\And
Yuzhu Chen \\
University of Science and Technology of China \\
cyzkrau@mail.ustc.edu.cn
\And
Yingjie Wang \\
Nanyang Technological University \\
yingjiewang1201@gmail.com
\And
Dacheng Tao \\
Nanyang Technological University \\
dacheng.tao@gmail.com
}

\begin{document}

\maketitle

\begin{abstract}
Understanding the inner mechanisms of black-box foundation models (FMs) is essential yet challenging in artificial intelligence and its applications. Over the last decade, the long-running focus has been on their explainability, leading to the  development of post-hoc explainable methods to rationalize the specific decisions already made by black-box FMs. However, these explainable methods have certain limitations in terms of faithfulness and resource requirement. Consequently, a new class of interpretable methods should be considered to unveil the underlying mechanisms of FMs in an accurate, comprehensive,  heuristic, and resource-light way.  This survey aims to review those interpretable methods that comply with the aforementioned principles and have been successfully applied to FMs. These methods are deeply rooted in machine learning theory, covering the analysis of generalization performance, expressive capability, and dynamic behavior. They provide a thorough interpretation of the entire workflow of FMs, ranging from the inference capability and training dynamics to their ethical implications. Ultimately, drawing upon these interpretations, this review identifies the next frontier research directions for FMs.
\end{abstract}

\newpage
\setcounter{tocdepth}{2}
\tableofcontents

\newpage

\section{Introduction}\label{sec1}
 Foundation Models (FMs), emerging as a new paradigm in deep learning, are fundamentally changing the landscape of artificial intelligence (AI). Unlike traditional models trained for specific tasks, FMs leverage massive and diverse datasets for training, which allows them to handle various downstream tasks through approaches like supervised fine-tuning (SFT) \cite{Bommasani2021FoundationModels}, Reinforcement Learning from Human Feedback (RLHF) \cite{xu2020preference}, Retrieval Augmented Generation (RAG) \cite{gao2023retrieval},  prompting engineering \cite{LiuYFJHN23,wei2022chain} or continual learning \cite{shi2024continual}.  In  natural language processing, most  FMs under autoregressive structure (e.g., GPT-3 \cite{Brown-NeurIPS-language-2020}, PaLM \cite{Aakanksha-JMLR-language-2023}, or Chinchilla \cite{Hoffmann-Chinchilla}) are established to output the next token given a sequence. To learn the distribution of images conditioned on textual input, efforts have been made to propose  text-to-image models, such as DALL·E \cite{Ramesh}. Video FMs, trained on video data, potentially combined with text or audio, can be categorized based on their pretraining goals: generative models like VideoBERT \cite{sun2019videobert} focus on creating new video content, while discriminative models like VideoCLIP \cite{xu2021videoclip} excel at recognizing and classifying video elements. Hybrid approaches like TVLT \cite{tang2022tvlt} combine these strengths for more versatile tasks. FMs have the potential to revolutionize a vast array of industries. Consider ChatGPT, a versatile chatbot developed by OpenAI, which has already made strides in enhancing customer support, virtual assistants, and voice-activated devices \cite{Achiam2023GPT4TR}. As FMs become more integrated into various products, their deployments will be scaled to accommodate a growing user base.  This trend is evident in the growing list of companies, such as Microsoft with Bing Chat and Google with Bard, all planning to deploy similar FMs-powered products, further expanding the reach of this technology.

Despite of rapid development of FMs,  there are also ongoing concerns that continue to arise. The underlying causes behind certain emerging properties  in FMs remain unclear. \cite{wei2022emergent} firstly consider a focused definition of emergent abilities of FMs, ``an ability is emergent if it is not present in smaller models but is present in larger models''. They conduct sufficient empirical studies to show emergent abilities can span a variety of large-scale language models, task types, and experimental scenarios.  On the contrary, serveral works claim that emergent abilities only appear for specific metrics, not for model families on particular tasks, and that changing the metric causes the emergence phenomenon to disappear \cite{schaeffer2024emergent}. Moreover, prompting strategies can elicit chains of thought (CoT) for a reasoning task, suggesting a potential for reasoning abilities of FMs \cite{wei2022chain}. However, recent critiques by Arkoudas highlight limitations in current evaluation methods, arguing that GPT-4's occasional displays of analytical prowess may not translate to true reasoning capabilities \cite{arkoudas2023gpt}. 

`All that glitters is not gold', as potential risks gradually emerge behind the prosperous phenomenon. The phenomenon of “hallucination” has garnered widespread public attention, referring to the tendency of FMs to generate text that sounds convincing, even if it is fabricated or misleading, as discussed in recent research \cite{bang2023multitask}.  Additionally, studies have shown that adding a series of specific meaningless tokens can lead to security failures, resulting in the unlimited leakage of sensitive content \cite{zou2023universal}. These issues serve as stark reminders that we still lack a comprehensive understanding of the intricate workings of FMs, hindering our ability to fully unleash their potential while mitigating potential risks.
 
\subsection{Historical Notes on Explainability and Interpretability}
The term ``explainability'' refers to the ability to understand the decision-making process of an AI model. Specifically, most explainable methods concentrates on providing users with post-hoc explanations on the specific decisions already made by the black-box models \cite{Rudin_stopexplainability,Zhao-XLLM-survey,luo2024understanding}.    Recently,  several survey studies   review  local and global explainable methods  for FMs  \cite{Zhao-XLLM-survey,luo2024understanding,changdan_interpretability_2024}. Local methods, like feature attribution \cite{ribeiro2016should,lundberg2017unified,mohebbi2021exploring}, attention-based explanation \cite{abnar2020quantifying,hoover2019exbert,yeh2023attentionviz}, and example-based explanation \cite{jin2019bert,ross2020explaining}, focus on understanding the model's decisions for specific instances. Conversely, global methods, such as probing-based explanation \cite{petroni2019language,li2024inference}, neuron activation explanation \cite{bills2023language,antverg2021pitfalls}, and mechanistic explanation \cite{chughtai2023toy,meng2022locating}, aim to  delve into the broader picture, revealing what input variables significantly influence the model as a whole. More specifically, these explanation techniques demonstrate excellent performance in exploring the differences in how humans and FMs work. For instance, \cite{li2024machines} utilize perturbation based explainable approach, SHAP (SHapley Additive exPlanations) \cite{lundberg2017unified}, to investigate the influence of code tokens on summary generation, and find no evidence of a statistically significant relationship between  attention mechanisms in FMs  and human programmers.  Endeavors are also being made to propose explainable methods tailored specifically  for FMs. \cite{zhao2023explaining} propose a novel framework to provide impact-aware explanations, which are robust to feature changes and influential to the model’s predictions. Recently, a GPT-4 based explainable method is utilized to explain the behavior of neurons in GPT-2 \cite{bills2023language}.

However, as discussed in Secton \ref{sec1-1}, these explainable methods have certain limitations in terms of faithfulness and resource requirement. Numerous researchers have endeavored to explore the term ``interpretability''. \cite{Doshi_interpretability} define the interpretability of a machine learning system as ``the ability to explain or present, in understandable terms to a human.''  \cite{Lipton_interpretability} highlight, interpretability and explainability seek to answer different questions: `How does the model work?' versus `What else can the model tell me?'. \cite{Rudin_stopexplainability} states that interpretable machine learning emphasizes creating inherently interpretable models, whereas explainable methods aim to understand existing black-box models through  post-hoc explanations. Following this definition, recent efforts have focused on developing inherently interpretable FMs based on symbolic or causal representations \cite{arenas2021foundations, rajendran2024learning,cunnington2024role}. However, 
a prevalent reality in the field of FMs is that the most widely used and high-performing FMs remain black-box in nature.

\begin{figure}[htb]
    \centering
    \includegraphics[width=1\linewidth]{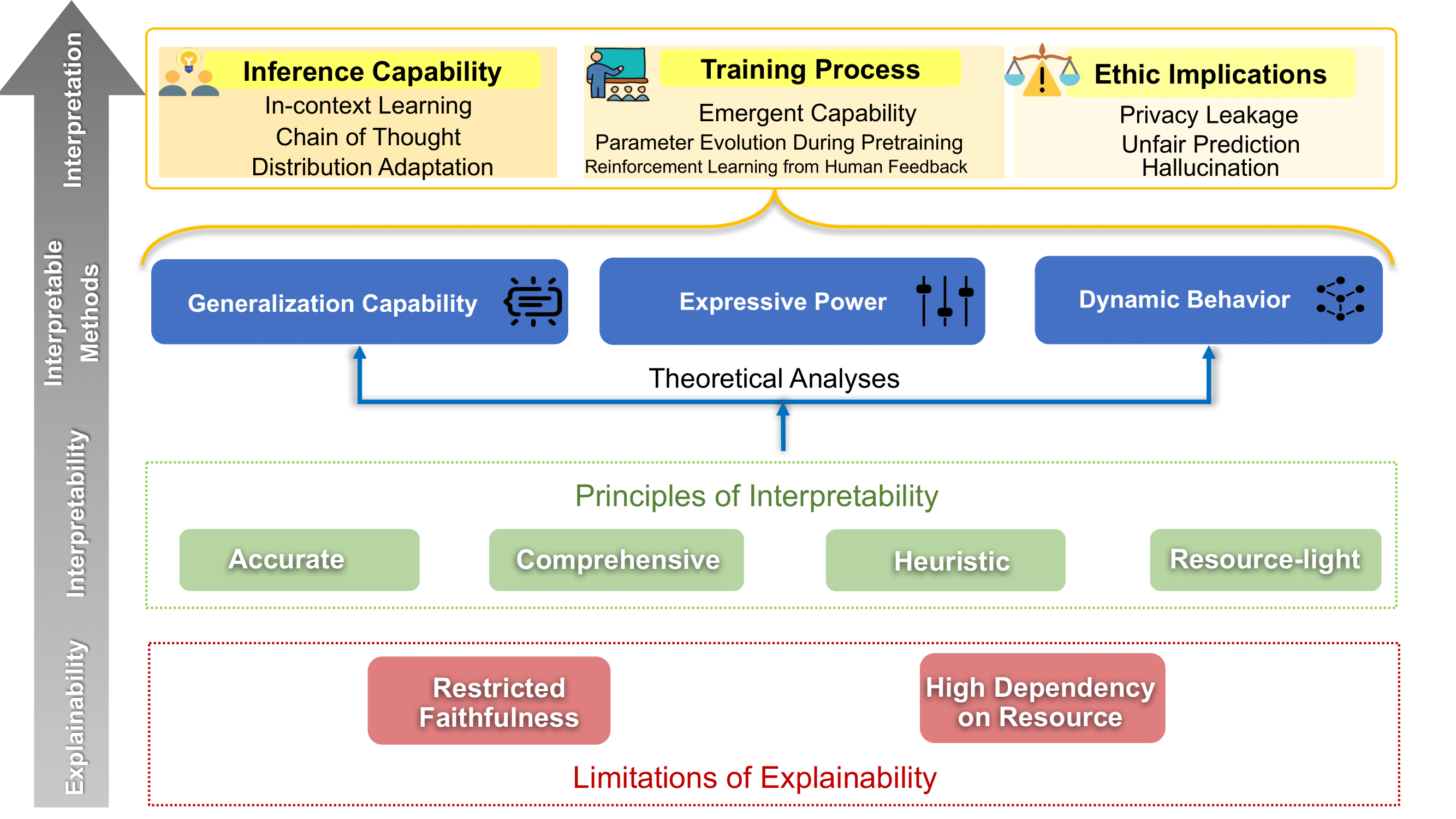}
    \caption{An overview of interpretability, interpretable method and interpretations tailored for FMs. }
    \label{fig:overview}
\end{figure}

\subsection{Objectives of This Survey}
From above all, there is no universally agreed-upon definition of interpretability within the machine learning community. The concept often varies based on the application domain and the intended audience \cite{Rudin_stopexplainability,carvalho2019machine}. This motivates us to define appropriate interpretability and develop effective interpretable methods to reveal the internal workings of FMs, as shown in Figure \ref{fig:overview}. In our perspective, the interpretability tailored for FMs refers to \emph{the ability to  reveal the underlying relationships between various related factors, such as model architecture, optimization, data and model performance, in an accurate, comprehensive, heuristic and resource-light way}.  To achieve this, we will develop interpretable methods tailored for FMs, grounded in rigorous learning theory. These methods will encompass analyses of generalization performance, expressivity, and dynamic behavior.  Generalization performance analysis, a form of uncertainty quantification, is an effective way to evaluate  FMs' ability to accurately predict or classify new, unseen data. Expressivity analysis is the process of evaluating a model's capacity to represent complex patterns and functions within the data. Dynamic behavior analysis involves  capturing the transformations and adaptations that occur within FMs as they learn from data, makes predictions, and undergoes training iterations. These interpretable methods align with the principles of interpretability as:
\begin{itemize}
\item  \emph{Accurate}. Compared to explainable methods based on post-hoc explanation, theoretical analysis, grounded in solid mathematical or statistical principles, has the ability to  predict the expected performance of algorithms under various conditions, ensuring accuracy  in their assessments. 
\item \emph{Comprehensive}. The insights drawn from theoretical constructs are typically more comprehensive, delving into the multidimensional influences on model performance. For instance, within the realm of model generalization analysis, we can discern the intricate ways in which factors like optimization algorithms and model architecture shape and impact the model's ability to generalize effectively.
\item \emph{Heuristic}. Theoretical analyses can provide heuristic insights by distilling complex information into practical rules, guidelines, or approximations that facilitate decision-making, problem-solving, and understanding of complex systems. These insights are derived from a deep understanding of the underlying principles but are presented in a simplified and practical form for application in real-world scenarios.
\item \emph{Resource-light}.  Theoretical analyses in machine learning are regarded as resource-light because they emphasize understanding fundamental concepts, algorithmic properties, and theoretical frameworks through mathematical abstractions. This approach stands in contrast to explainable methods that entail heavy computational demands for training additional models.
\end{itemize}



\begin{table}[htp!]
\centering\scriptsize
\begin{tabular}{c|ccc}
\hline
Methods & Generalization & Expressive Power & Dynamic Behavior \\
\hline
In-Context Learning & \checkmark & \checkmark & - \\
Chain-of-Thought & \checkmark & \checkmark & - \\
Adaption to Distribution Shifts & \checkmark & - & - \\
Emergent Capability & - & - & \checkmark \\
Parameter Evolution & - & - & \checkmark \\
RLHF & - & - & \checkmark \\
Privacy Leakage & \checkmark & - & \checkmark \\
Unfair Prediction & \checkmark & - & - \\
Hallucination & - & \checkmark & - \\
\hline
\end{tabular}
\caption{Various interpretable methods are employed to unveil distinct aspects of FMs. Generalization analysis is utilized to interpret In-Context Learning, Chain-of-Thought, Adaptation to Distribution Shift, Privacy Leakage, and Unfair Prediction. Expressive power analysis is applied to interpret In-Context Learning, Chain-of-Thought, and Hallucination. Dynamic behavior analysis is employed to interpret Emergent Capability, Parameter Evolution, RLHF, and Privacy Leakage.}
\label{tab:method}
\end{table}

The structure of this survey is as follows: Section \ref{sec1-1} shows the bottleneck of explainability in FMs. Section \ref{sec2} reviews key interpretable methods, including analyses of generalization performance, expressivity, and dynamic behavior. The remaining sections aim to interpret the characteristics of FMs via three interpretable methods (as shown in Table \ref{tab:method}). Section \ref{sec3} delves into the inference capabilities of FMs, exploring in-context learning, CoT reasoning, and distribution adaptation. Section \ref{sec4} provides detailed interpretations of FMs training dynamics. Section \ref{sec5} focuses on the ethical implications of FMs, including privacy preservation, fairness, and hallucinations. Finally, Section \ref{sec6} discusses future research directions based on the insights gained from the aforementioned interpretations.

\section{The Bottleneck of Explainability in Foundation Models}\label{sec1-1}
 Most explainable methods concentrates on providing users with post-hoc explanations on the specific decisions already made by the black-box models \cite{Rudin_stopexplainability,Zhao-XLLM-survey,luo2024understanding}. Despite their promising performance in explaining some black-box models, the explanations garnered from these methods for complex FMs are often not entirely reliable and may even be misleading.
\subsection{Restricted Faithfulness}
Faithfulness usually refers to how accurately the explainable methods reflect the true decision process of the black-box models \cite{herman2017promise}. Unfortunately, some studies indicate that these post-hoc explainable methods appear to be inherently unable to fully maintain faithfulness.  \cite{Rudin_stopexplainability} argue that if an explanation perfectly mirrored the model's computations, the explanation itself would suffice, rendering the original model unnecessary. This highlights a key limitation of explainable methods for complex models: any explanation technique applied to a black-box model is likely to be an inaccurate representation. Based on that fact, the  mainstream work focuses on  evaluating the faithfulness of post-hoc explainable methods from different perspectives, such as consistency and sufficiency \cite{dasgupta2022framework}, usefulness to humans \cite{jesus2021can,poursabzi2021manipulating} and robustness \cite{agarwal2022rethinking}. 

\begin{figure}
    \centering
    \includegraphics[width=1\linewidth]{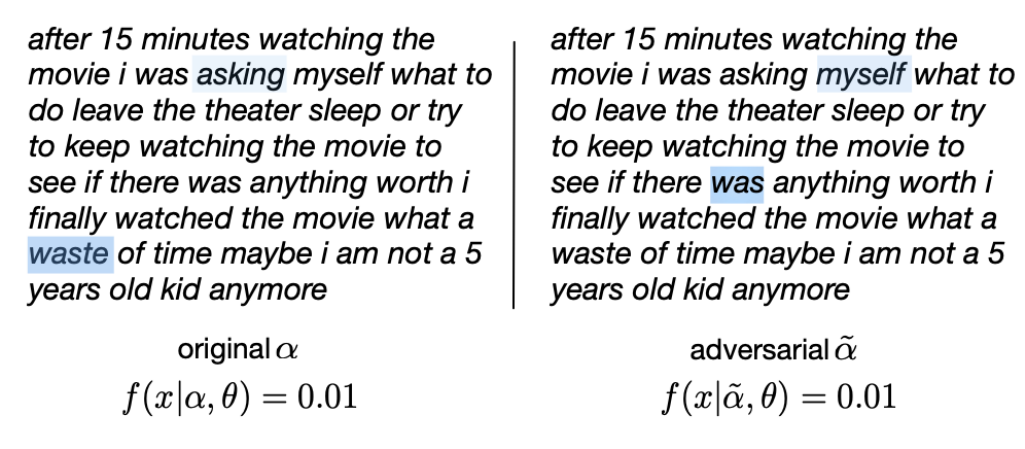}
    \caption{Heatmap displaying attention weights generated from a negative movie review. On the left, we present the model's actual attention pattern on, while on the right, we showcase a set of attention weights created adversarially. Despite their significant differences, both patterns result in the same prediction (0.01) effectively.}
    \label{fig:comprehensiveness}
\end{figure}

From our perspective, comprehensiveness appears as an important metric for evaluating the faithfulness of FMs. Comprehensiveness specifically denotes that the explanations provided should be closely linked to the FMs' performance. Taking an example to illustrate the limitation of explainable methods in terms of comprehensiveness. Due to the capability to capture the long-distance dependency relationship between features, self-attention mechanisms have been widely-used in various FMs. The self-attention visualized by some explainable methods such as BertViz \cite{vig-2019-multiscale}, is commonly viewed as a explanation for the working mechanism of FMs by presenting the relative importance of units.   However, we may drawn  misleading  relationship between theses attention weights and model outputs (as shown in Figure \ref{fig:comprehensiveness} shown in \cite{jain2019attention}). The left panel displays the initial attention distribution $\alpha$ across the words in a specific movie review using a conventional attentive BiLSTM setup for sentiment analysis. It might be tempting to infer from this that the token ``waste'' predominantly influences the model's `negative' classification ($\hat y=0.01$). However, an alternative attention distribution $\tilde{\alpha}$ (right panel) could be crafted to focus on entirely different tokens while still producing a nearly identical prediction (assuming all other parameters of $f$, $\theta$, remain constant). This means that the discovered attention is linked to the model's performance at a lower level.

 \subsection{Heavy Resource-dependency} 
 As models become more complex, the difficulty of explaining them using traditional post-hoc methods like SHAP \cite{lundberg2017unified} and Local Interpretable Model-agnostic Explanations (LIME) \cite{ribeiro2016should}. Despite the valuable explanations provided by these methods, they have limitations in terms of computational complexity. For example, for each testing sample, the weight parameters derived from local linear models are integrated into SHAP. Nevertheless, SHAP necessitates models to be constructed for every testing or validation sample, leading to a computationally expensive process when globally calculating SHAP values.

 \begin{table}[htp!]
\centering\scriptsize
\begin{tabular}{c|cc}
\hline
TreeSHAP Version & Time Complexity & Space Complexity \\
\hline
Original TreeSHAP & $O(MTLD^2)$ & $O(D^2+|N|)$ \\
Fast TreeSHAP v1 & $O(MTLD^2)$ & $O(D^2+|N|)$ \\
Fast TreeSHAP v2 (General case)  & $O(TL2^DD+MTLD)$ & $O(L2^D)$  \\
Fast TreeSHAP v2 (Balanced case)  & $O(TL^2+MTLD)$ & $O(L^2)$  \\
\hline
\end{tabular}
\caption{Summary of space complexity and computational complexities of TreeSHAP algorithms \cite{yang2021fast}}
\label{tab:shap}
\end{table}

Recently, there are two ways to mitigate the highly computational complexity with SHAP. Firstly,  to address the computational challenges,  a novel class of stochastic estimators is proposed by leveraging feature subset or permutation sampling \cite{vstrumbelj2014explaining,covert2020understanding}. While these estimators provide consistent results, their high computational complexity, often requiring many model evaluations, hinders their scalability and real-world deployment. Secondly, 
previous work has explored model-specific Shapley value approximations for trees  \cite{lundberg2020local}  and neural networks \cite{shrikumar2017learning,chen2018shapley,ancona2019explaining}, often leveraging their unique structures. While these methods can reduce computational cost, they may introduce bias in feature attribution and struggle to accurately handle features that were not seen during training \cite{aas2021explaining,janzing2020feature,covert2021explaining}. Additionally, these methods result in a limited performance in computational cost when explaining large models with huge amount data. Table \ref{tab:shap}, borrowed from \cite{yang2021fast}, presents the summary of computational complexities of TreeSHAP algorithms, where $D$ is the maximum depth of a tree, $|N|$ stands for the dimension of features, $T$ denotes the number of trees, $M$ represents the total samples size, and $L$ indicates the maximum number of leaves in a tree. From this table,  even the accelerated variant of TreeSHAP reveals a considerable computational overhead when elucidating complex models with substantial datasets.

\section{Interpretable Methods}\label{sec2}
\begin{figure}[htb]
    \centering
    \includegraphics[width=1\linewidth]{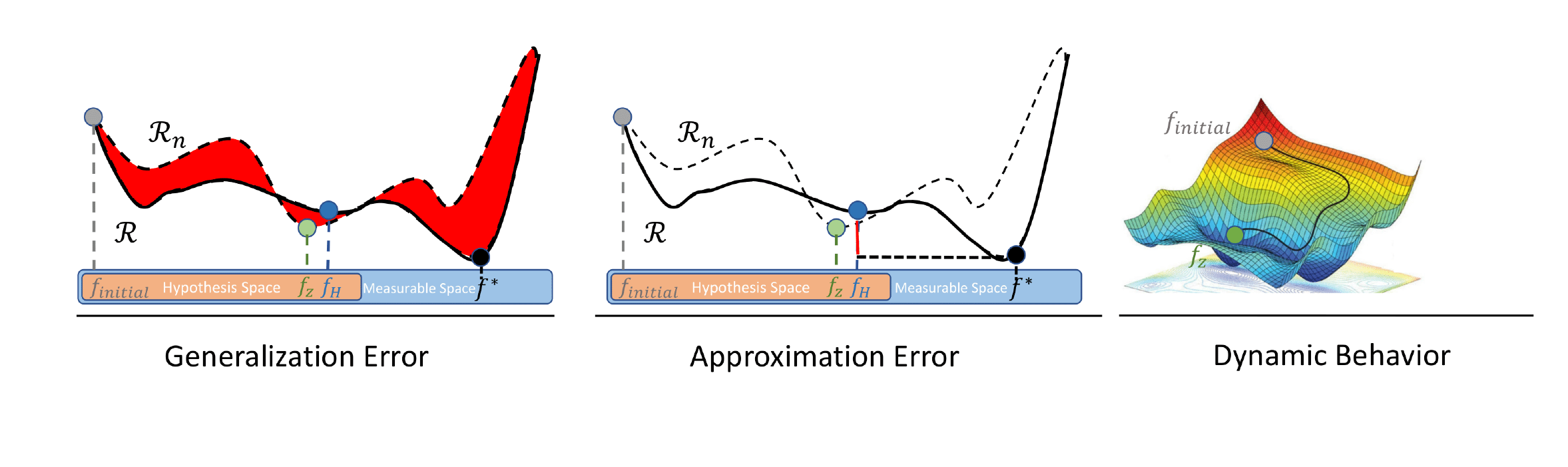}
    \caption{The interpretable methods: generalization analysis, approximation capability analysis, and dynamic behavior analysis.     In these figures, the empirical risk $\mathcal R_n$ (dotted line) and expected risk $\mathcal R$ (solid line) over a measurable function space are described, with $f_{initial}$, $f_\mathbf{z}$, $f_H$, and $f^*$ representing the initial model, learned model, optimal function over the hypothesis space, and optimal function over the measurable function space.  The generalization error (depicted by the red area and formally denoted as $|\mathcal R(f) - \mathcal{R}_n(f)|$) offers insights into a model's performance on unseen data. The approximation error (depicted by the red solid line and formally denoted as $|\mathcal R(f_H) - \mathcal{R}(f^*)|$), as a manifestation of expressive power,  investigates the discrepancy between the true underlying function and the optimal function within the hypothesis space. Dynamic behavior in this context refers to the trajectory of FMs' parameter changes from initial point $f_{initial}$ to converged solution $f_z$ using methods such as gradient descent or RLHF, which can provide valuable insights into the FMs' characteristics and potential issues. }
    \label{fig:interpretablemethod}
\end{figure}

In this section, we introduce three interpretable methods grounded in rigorous machine learning theory, namely generalization analysis, expressive power analysis, and dynamic behavior analysis (as shown in Figure \ref{fig:interpretablemethod})  \cite{suh2024survey,sun2019optimization,bartlett2021deep}. These methods provide a systematic theoretical framework for analyzing the complex behaviors of FMs spanning from training to inference phases. This includes interpreting phenomena like ICL learning and CoT reasoning, as well as addressing ethical implications such as privacy leakage, unfair prediction, and hallucinations. With the heuristic insights from these interpretations, we can further refine these powerful models. 

\subsection{Preliminary}\label{sec. preliminary}
Before the analysis of generalization, expressive power, and dynamic behavior, we first introduce the preliminary of the target of FMs and notions for the following analysis. 

FMs represent a class of machine learning techniques that utilize large-scale neural networks, characterized by multiple layers of neurons and nonlinear activation functions. These models are engineered to process and learn from diverse training datasets, extracting intricate patterns and features through a series of computational layers. The data they utilize can come from real-world records, human interactions, or even synthetic data generated by other models. By leveraging proper architectures and training algorithms, empirical results show that FMs can autonomously uncover and represent intricate relationships within most types of data. This enables them not only to make sophisticated predictions and informed decisions but also to generate high-quality synthetic data based on given prompts and the extensive knowledge they have accumulated. 

In real-world application, FMs serve the purpose of providing reasonable, meaningful, and helpful responses to user inputs. Mathematically, the goal of FMs is to approximate a mapping that aligns with what humans would consider an ideal output. This mapping treats $\mathcal{X}$ and $\mathcal{Y}$ as the input and output spaces, respectively. To achieve this, each FM is accessible to a set of input-output pairs
\begin{equation*}
 \mathcal{S}=\{(x,y):x\in\mathcal{X},y\in\mathcal{Y}\}. 
\end{equation*}
where $x$ is the input and $y$ is one of the reasonable output. For example, consider a language translation FM designed to translate English sentences into French. Here, $\mathcal{X}$ would be the space of all possible English sentences, and $\mathcal{Y}$ would be the space of all possible French sentences. The set $\mathcal{S}$ would consist of pairs of English sentences and their corresponding French translations, such as (``Hello, how are you?", ``Bonjour, comment \c{c}a va?"). 

To achieve better performance, the accessible dataset $\mathcal{S}$ is always divided into three subsets: training, validation, and testing sets. The training set is used to teach the FM by allowing it to learn the intricate patterns and relationships within the data. This subset forms the core of the learning process, enabling the model to adjust its parameters and improve its performance iteratively. The validation set, on the other hand, is used to evaluate the model's performance during the training process. It helps in selecting the best model architecture and hyperparameters by providing a separate dataset that the model has not seen during training. This ensures that the model generalizes well to new, unseen data and prevents overfitting to the training set. Finally, the testing set is used to evaluate the model's performance objectively. This subset is completely isolated from the training and validation processes, providing an unbiased assessment of how well the FM can handle real-world data. 

It is important to note that not all $(x,y)$ pairs are ideal; some pairs may be noisy or contain errors, reflecting real-world imperfections. Additionally, in some cases, the output $y$ may not be provided, corresponding to unsupervised learning scenarios. 

\begin{remark}
It’s crucial to acknowledge that we always assume samples in $\mathcal{S}$ is i.i.d, where each data point is assumed to be drawn from the same distribution $\mathcal{D}$ and is unrelated to other data points, may not hold in many real-world scenarios. 
For instance, in recommendation systems, user interactions and preferences are often influenced by various factors such as temporal dynamics, social influences, and contextual information, leading to data that is inherently non-i.i.d. 
Despite these complexities and the non-i.i.d nature of real-world data, FMs have shown remarkable robustness and effectiveness. They are capable of capturing intricate patterns and interactions within the data, thereby maintaining strong performance even when the i.i.d assumption is violated. 
However, for the purpose of this survey, we will primarily focus on the scenarios where the i.i.d assumption is maintained. 
\end{remark}

With the accessible dataset, FMs first map all $x\in\mathcal{X}$ to $R^{d_X}$ and $y\in\mathcal{Y}$ to $R^{d_Y}$. This transformation is crucial in various applications as it allows models to work with structured data representations. For example, in image classification tasks, $(x, y)$ represents an image-label pair, where the image is mapped to a vector of pixel values and the label is transformed into a one-hot encoded vector. In this survey, this process is not important thus we just let $\mathcal{X}=R^{d_X}$ and $\mathcal{Y}=R^{d_Y}$. 

Then, FMs are trained to learn a hypothesis, denoted by $h$, from the training data. Due to the specified architecture and numerical precision established by FMs, there are constraints on $h$. Mathematically, all such possible hypotheses constitute the hypothesis space, denoted by $\mathcal{H}$. For instance, when an FM is a neural network, $\mathcal{H}$ is denoted as
\begin{equation*}
\{h:h(x) = W_D \sigma_{D-1} (W_{D-1} \sigma_{D-2} (\ldots \sigma_1(W_1 x)))\}
\end{equation*}
where the depth of FMs, the $j$-th weight matrix and the $j$-th activation function are denoted by  $D$,  $W_j$  and $\sigma_j$. Here, the activation function $\sigma_j$ is assumed to be  continuous such that $\sigma_j(0) = 0$. Popular activation functions include the softmax, sigmoid, and tanh functions. 

To interpret FMs, theoretical research can be categorized into 3 parts: generalization analysis, expressive power analysis, and dynamic behavior analysis. Their purpose and motivation are as follows: 
\begin{enumerate}
    \item Generalization analysis: This analysis focuses on understanding how well foundation models generalize from the training data to unseen data. The goal is to theoretically explain the factors influencing the generalization performance, such as model complexity, data distribution, and the training dynamics. It provides insights into why FMs can perform well in diverse applications even with limited training examples in some domains.
    \item Expressive power analysis: This part examines the capacity of foundation models to represent complex functions or tasks. The motivation is to quantify the types of functions that the model can approximate, given its architecture. Understanding the expressive power helps in determining the potential and limitations of FMs in capturing the intrinsic structure of different data types, thus guiding architectural improvements. 
    \item Dynamic behavior analysis: This analysis investigates the training dynamics of FMs, including optimization trajectories and the evolution of model parameters during training. The purpose is to understand how the model’s behavior changes during training and how different training techniques, such as learning rate schedules or regularization, impact the final model. This perspective is crucial for improving training stability and efficiency, as well as for mitigating issues like catastrophic forgetting or mode collapse.
\end{enumerate}

\subsection{Generalization  Analysis}

Generalization error measures the performance of a machine learning model on unseen data.  Mathematically, for a hypothesis $h$ and a dataset $S = \left\{  {\left( {{x}_{1},{y}_{1}}\right) ,\ldots ,\left( {{x}_{n},{y}_{n}}\right) }\right\}$, the generalization error can be expressed as:
$$
\mathcal R(\mathcal{A}(S))-\mathcal{R}_n(\mathcal{A}(S)) = {\mathbb{E}}_{(x,y) \sim  \mathcal{D}}\lbrack \ell (\hat{h}(x),y)\rbrack  - \frac{1}{n}\mathop{\sum }\limits_{{i = 1}}^{n}\ell (\hat{h}({x}_{i}),{y}_{i})
$$
where $\mathcal{D}$ denotes the data distribution, $\ell$ is the loss function, and $\mathcal{A}: S \rightarrow h$ represents the learning algorithm. Lower generalization error bound indicates a model that better generalizes from the training data to new data, thereby enhancing its predictive performance on real-world tasks.

Usually, excess risk, a directly related concept to generalization error are used to quantify the difference between the expected loss of a model on the training data and the expected loss of the best possible model on the same data.  Mathematically, it can be expressed as:
$$
\mathcal R(\mathcal{A}(S))-\min_{h\in \mathcal{H}}\mathcal{R}(h) = {\mathbb{E}}_{(x,y) \sim  \mathcal{D}}\lbrack \ell (\hat{h}(x),y)\rbrack  - \min_{h\in \mathcal{H}}{\mathbb{E}}_{(x,y) \sim  \mathcal{D}}\lbrack \ell (h(x),y)\rbrack.
$$
 If the excess risk is small, it implies that the model is not overfitting to the training data and can generalize well to unseen data. Excess risk is a direct upper bound on generalization error. This means that if you can control the excess risk, you can also control the generalization error.

Usually,  generalization error depends on the complexity of the hypothesis space \cite{shalev2014understanding}. Following this approach, a series of works have examined the VC-dimension and Rademacher complexity to establish the upper bounds on the generalization error.

~\\
\textbf{VC-dimensition}.
The Vapnik-Chervonenkis dimension (VC-dimension, \cite{blumer1989learnability,cherkassky1999model,vapnik2006estimation}) 
is an expressive capacity measure for assessing the hypothesis complexity. The pertinent definitions are presented below.
~\\
\begin{definition}
For any non-negative integer $n$, the growth function of a hypothesis space $\mathcal H$ is defined as follows:
\begin{equation}
\Pi_{\mathcal H} (n) \coloneqq \max_{x_1, \ldots, x_n \in \mathcal X} | \{ (h(x_1), \ldots, h(x_n)): h \in \mathcal H \} |.
\end{equation}
If $\Pi_{\mathcal H} (n) = 2^n$, the dataset $\{x_1, \ldots, x_n\}$ is  shattered by $\mathcal H$. The VC-dimension, donated as $\text{VCdim}(\mathcal H)$, is the largest size $n$ of a shattered set. In other words, $\text{VCdim}(\mathcal H)$ is the largest value of $m$ such that $\Pi_H(n)=2^n$. If there does not exist a largest such $n$, we define $\text{VCdim}(\mathcal H)=\infty$.
\end{definition}
~\\
Elevated VC dimensions correlate with the model's increased capability to shatter a wider range of samples, enhancing expressive capacity. Through the VC dimension, one can establish a uniform generalization bound as follows \cite{mohri2018foundations}. Specifically, assume the hypothesis space $\mathcal H$ has VC-dimension $\text{VCdim}(\mathcal H)=d$. Then, for any $\delta > 0$, the following inequality holds 
\begin{equation*}
 \mathcal R(h)- \mathcal{R}_n (h) \le  \sqrt{\frac{2 d \log \frac{en}{d}}{n}} + \sqrt{\frac{\log \frac{1}{\delta}}{2n}}, ~~\forall h \in \mathcal H
\end{equation*}
with probability $1 - \delta$ over the choice of $S \sim \mathcal{X}^n$. 

~\\
\textbf{Rademacher complexity}.
Rademacher complexity captures the ability of a hypothesis space to fit random labels as a measure of expressive capacity \cite{mohri2018foundations,tu2020understanding}. An increased Rademacher complexity signifies higher hypothesis complexity and, correspondingly, an augmented capacity for model expressiveness. The common definition are
presented below.
\begin{definition}[Empirical Rademacher complexity]
Given a real-valued function class $\mathcal H$ and a dataset $S$, the empirical Rademacher complexity is defined as follows,
\begin{equation*}
\hat{\mathfrak R}(\mathcal H) = \mathbb E_{\boldsymbol{\sigma}}\left[ \sup_{h \in \mathcal H} \frac{1}{m} \sum_{i = 1}^n \sigma_i h(x_i) \right],
\end{equation*}
where $\boldsymbol{\sigma}=\{\sigma_1, \ldots, \sigma_n\}$ are i.i.d. Rademacher random variables with $$
\mathbb{P}\left\{\sigma_i=1\right\}=\mathbb{P}\left\{\sigma_i=-1\right\}=\frac{1}{2}.
$$
\end{definition}
One can establish a worst-case generalization error bound via the Rademacher complexity. Specifically, for a loss function with an upper bound of $B$, for any $\delta \in(0,1)$, with probability at least $1-\delta$, the following holds for all $h \in \mathcal{H}$,
$$
\mathcal{R}(h)- \mathcal{R}_n(h) \leq 2 B \hat{\mathfrak R}(\mathcal H)+3 B \sqrt{\log(2/\delta) / n} .
$$

Both \cite{Neyshabur_2017} and \cite{zhang2017understanding} indicate that deep learning models commonly display overparameterization in practical settings, possessing a vastly greater number of parameters than training samples. Under such overparameterized regimes, both the VC dimension and Rademacher complexity of these models tend to become excessively large. Thus, the practical utility of bounds derived from these complexity measures is relatively limited.

~\\
\textbf{Covering Numbers.} Next, we examine an alternative approach to assessing the complexity of sets, known as covering numbers \cite{shalev2014understanding}.

    \begin{definition}(Covering Numbers) Let \( A \subseteq \mathbb{R}^m \) denote a collection of vectors. We define that \( A \) is \( r \)-covered by a set \( A' \) with regard to the Euclidean metric if, for every \( \boldsymbol{a} \in A \), there exists \( \boldsymbol{a'} \in A' \) such that \( \|\boldsymbol{a} - \boldsymbol{a'}\| \leq r \). We denote by \( \mathcal{N}(r, A) \) the cardinality of the smallest set \( A' \) that \( r \)-covers \( A \).
\end{definition}
This defines the concept of one set being covered by another within a specified radius \( r \) in Euclidean space, where \( \mathcal{N}(r, A) \) denotes the minimum number of points required to cover \( A \) within that radius. Next, we provide an upper bound for the Rademacher complexity of \( A \), using the covering numbers \( \mathcal{N}(r, A) \) as the basis. This technique, referred to as \textit{Chaining}, is attributed to \cite{dudley2010universal}.
\begin{theorem}(From Covering to Rademacher Complexity via Chaining \cite{shalev2014understanding})\label{theorem_chaining} For any \( A \subseteq \mathbb{R}^m \), and vector \( \boldsymbol{a}, \boldsymbol{a'} \in \mathbb{R}^m \). Let \( c = \min_{\boldsymbol{a'}} \max_{\boldsymbol{a} \in A} \|\boldsymbol{a} - \boldsymbol{a'}\| \). Then, for any integer \( M > 0 \), the rademacher complexity of $A$ satisfies:
\[
\hat{\mathfrak R}(A) \leq \frac{c 2^{-M}}{\sqrt{m}} + \frac{6c}{m} \sum_{k=1}^{M} 2^{-k} \sqrt{\log(\mathcal{N}(c 2^{-k}, A))}.
\]
\end{theorem}
Chaining is a robust method in statistical learning theory that refines the bounds on the complexity of function classes by constructing progressively finer covers of the underlying space, thereby providing tighter control over the Rademacher complexity. The corresponding proof framework is outlined as follows:

~\\
\textbf{Proof Sketch for Chaining Arguments in Theorem \ref{theorem_chaining}}

\begin{itemize}
\item \textbf{Initial Setup} Assume that \( \boldsymbol{a'} = \boldsymbol{0} \), where \( \boldsymbol{a'} \) is a minimizer of the objective function as defined for the complexity measure \( c \). And define the set \( B_0 = \{0\} \), which serves as a \( c \)-cover of the set \( A \). Then, construct a sequence of sets \( B_1, B_2, \dots, B_M \), where each \( B_k \) represents a minimal \( (c2^{-k}) \)-cover of \( A \).
\item \textbf{Maximizer \( \boldsymbol{a^*} \)}. Define \( \boldsymbol{a^*} = \arg\max_{\boldsymbol{a} \in A} \langle \boldsymbol{\sigma}, \boldsymbol{a} \rangle \), which is chosen as a function of \( \boldsymbol{\sigma} \). If there are multiple maximizers, one can be chosen arbitrarily, and if a maximizer does not exist, select \( \boldsymbol{a^*} \) such that \( \langle \boldsymbol{\sigma}, \boldsymbol{a^*} \rangle \) is sufficiently close to the supremum.
\item \textbf{Nearest Neighbor and the Triangle Inequality}. For each \( k \), let \( \boldsymbol{b^{(k)}} \) denote the nearest neighbor of \( \boldsymbol{a^*} \) in the set \( B_k \). The vector \( \boldsymbol{b^{(k)}} \) is also a function of \( \boldsymbol{\sigma} \). Apply the triangle inequality:
\[
\|\boldsymbol{b^{(k)}} - \boldsymbol{b^{(k-1)}}\| \leq \|\boldsymbol{b^{(k)}} - \boldsymbol{a^*}\| + \|\boldsymbol{a^*} - \boldsymbol{b^{(k-1)}}\| = 3c \cdot 2^{-k}.
\]
For each \( k \), define the set 
$$
\hat{B}_k = \{ (\boldsymbol{a} - \boldsymbol{a'}) : \boldsymbol{a} \in B_k, \boldsymbol{a'} \in B_{k-1}, \|\boldsymbol{a} - \boldsymbol{a'}\| \leq 3c \cdot 2^{-k} \}.
$$
\item \textbf{Upper Bound on the Rademacher Complexity \(\hat{\mathfrak R}(A)\)}. The Rademacher complexity \( \hat{\mathfrak R}(A) \) can be written as:
\[
\hat{\mathfrak R}(A) = \frac{1}{m} \mathbb{E} \left[ \langle \boldsymbol{\sigma}, \boldsymbol{a^*} \rangle \right].
\]
This is further expanded as:
\[
\hat{\mathfrak R}(A) \leq \frac{1}{m} \mathbb{E} \|\boldsymbol{\sigma}\| \cdot \|\boldsymbol{a^*} - \boldsymbol{b^{(M)}}\| + \sum_{k=1}^{M} \frac{1}{m} \mathbb{E} \left[ \langle \boldsymbol{\sigma}, \boldsymbol{b^{(k)}} - \boldsymbol{b^{(k-1)}}\rangle \right].
\]
\item  \textbf{Application of Massart's Lemma and Covering Number Bound}. By Massart's lemma, we obtain the bound:
\[
\frac{1}{m} \mathbb{E} \sup_{\boldsymbol{a} \in B_k} \langle \boldsymbol{\sigma}, \boldsymbol{a} \rangle \leq 6c \cdot 2^{-k} \sqrt{\frac{\log(\mathcal{N}(c 2^{-k}, A))}{m}},
\]
where \( \mathcal{N}(c 2^{-k}, A) \) is the \( (c 2^{-k}) \)-covering number of the set \( A \).
\item \textbf{Final Result via Combination of Inequalities}. Combining the above inequalities, we arrive at the final upper bound for the Rademacher complexity:
\[
\hat{\mathfrak R}(A) \leq \frac{c 2^{-M}}{\sqrt{m}} + \frac{6c}{m} \sum_{k=1}^{M} 2^{-k} \sqrt{\log(\mathcal{N}(c 2^{-k}, A))}.
\]

\end{itemize}

~\\
\textbf{Algorithmic Stability.} Stability concepts address these limitations and offer more refined insights into the optimization-dependent generalization behavior. 
Early work by \cite{bousquet2002stability} introduced the notion of algorithmic stability and demonstrated how it could be used to derive generalization bounds. They showed that if an algorithm is uniformly stable, meaning the output hypothesis does not change significantly with small perturbations in the training data, then it is possible to establish tight generalization bounds. Their pioneering work laid the foundation for subsequent research in understanding the stability properties of various learning algorithms and their impact on generalization performance \cite{feldman2018generalization,feldman2019high,bousquet2020sharper,fu2022sharper}. In particular, uniform stability and error stability are key algorithmic properties that will be utilized in the analyses presented in this section.

\begin{definition}[Uniform stability \cite{bousquet2002stability}] Let $S$ and $S^{\prime}$ be any two training samples that differ by a single point.  A learning algorithm $\mathcal{A}$ is $\beta$-uniformly stable if 
\begin{eqnarray*}
		\sup_{z\in \mathcal{Z}}[|\ell(\mathcal{A}(S),z)-\ell(\mathcal{A}(S^{\prime}),z)|]\leq\beta.
	\end{eqnarray*}
\end{definition}
where $z=(x,y)$ and $\ell$ is the loss function.
\begin{definition}[Error stability \cite{bousquet2002stability}]\label{errorstability} A learning algorithm $\mathcal{A}$ has error stability $\beta$ with respect
to the loss function $l$ if the following holds:
$$\forall S\in\mathcal{Z}^n,\forall i\in\{1,\ldots,n\},|\mathbb{E}_z[\ell(\mathcal{A}(S),z)]-\mathbb{E}_z[\ell(\mathcal{A}(S^{\setminus i}),z)]|\leq\beta.$$
\end{definition}

Consequently, if algorithm $\mathcal{A}$ is applied to two closely similar training sets, the disparity in the losses associated with the resulting hypotheses should be constrained to no more than $\beta$.  \cite{bousquet2020sharper} have demonstrated that, utilizing moment bound and concentration inequality, generalization error bounds of stable learning algorithms can be derived as follows:

\begin{theorem}[Exponential generalization bound in terms of uniform stability]
\label{thm:stability}
Let algorithm $\mathcal{A}$ be $\beta$-uniformly stable, and let the loss function satisfy $l(h,z)\leq M$, $\forall h\in\mathcal{H}$ and $\forall z\in\mathcal{Z}$. Given a training sample $S=\{(x_i,y_i)\}_{i=1}^n$, for any $\delta\in(0,1)$, it holds that:
$$
\mathcal{R}(\mathcal{A}(S))- \mathcal{R}_n(\mathcal{A}(S)) \lesssim  \beta \log n \log \left(1/\delta\right)+M \sqrt{\log \left(1/\delta\right)/n}
$$
with probability at least $1-\delta$.
\end{theorem}
~\\
\textbf{Proof Sketch for Stability Bound in Theorem \ref{thm:stability}}

~\\
\emph{Step 1: Initial Setup and Assumptions}
\begin{itemize}
    \item \emph{Assume Uniform Stability and Define Risk Difference}: Let \( \mathcal{A} \) be a learning algorithm with uniform stability parameter \( \beta \), meaning:
   \[
   | \ell(\mathcal{A}_S(x), y) - \ell(\mathcal{A}_{S_i}(x), y) | \leq \beta,
   \]
   where \( S_i \) denotes the sample \( S \) with the \( i \)-th point replaced. Define the difference between the empirical risk \( \mathcal{R}_{n}(\mathcal{A}_S) \) and true risk \( \mathcal{R}(A_S) \) as
   \[
   n(\mathcal{R}(\mathcal{A}_S) - \mathcal{R}_{n}(\mathcal{A}_S)).
   \]
\end{itemize}
~\\
\emph{Step 2: Expressing Risk Difference in Terms of Stability-Bound Deviation Functions}
\begin{itemize}
    \item \emph{Decomposition via Stability-Bound Deviation Functions}: Define auxiliary functions \( g_i \) to express the stability-based deviation between empirical and true risks. Specifically, let
   \[
   g_i = \mathbb{E}_{(X', Y')} \left[ \mathbb{E}_{(X, Y)} \left( \ell(\mathcal{A}_{S_i}(X), Y) - \ell(\mathcal{A}_{S_i}(X_i), Y_i) \right) \right],
   \]
   where \( S_i \) is the modified sample as above. It is noteworthy that the function  $g_i$ also satisfies the conditions $\left|\mathbb{E}[g_{i}(Z)|Z_{i}]\right|\leq M$ and $\mathbb{E}[g_{i}(Z)|Z_{[n]\setminus\{i\}}]=0$. By uniform stability, the total risk difference can be approximated as:
   \[
   \left| n(\mathcal{R}(\mathcal{A}_S) - \mathcal{R}_{n}(\mathcal{A}_S)) - \sum_{i=1}^n g_i \right| \leq 2\beta n.
   \]
   Thus, the true risk difference closely follows the sum of \( g_i \) terms, up to a factor proportional to \( \beta n \).
\end{itemize}
~\\
\emph{Step 3: Establishing a Moment Bound for the Sum of \( g_i \) Terms}
This step involves deriving a moment bound for the sum \( \sum_{i=1}^n g_i \) to control the risk difference with high probability.

\begin{itemize}
    \item \emph{Step 3.1: Constructing a Sequence of Partitions for Conditional Expectations}. Assume without loss of generality that \( n = 2^k \). If \( n \) is not a power of two, we can add extra zero functions to double the terms, making \( n = 2^k \). Define a sequence of partitions \( B_0, B_1, \ldots, B_k \), where each \( B_l \) is a partition of \( [n] \) into \( 2^{k-l} \) subsets, each of size \( 2^l \). For example:
   \[
   B_0 = \{ \{1\}, \ldots, \{2^k\} \}, \ldots, B_k = \{ \{1, \ldots, 2^k\} \}.
   \]
   For each \( i \in [n] \) and each level \( l = 0, \ldots, k \), denote by \( B_l(i) \) the unique set in \( B_l \) that contains \( i \).
   \item  \emph{Step 3.2: Defining Conditional Expectations and Telescoping Terms}. For each \( i \in [n] \) and each level \( l = 0, \ldots, k \), let:
   \[
   g_i^l = g_i^l(Z_i, Z_{[n] \setminus B_l(i)}) = \mathbb{E}[g_i | Z_i, Z_{[n] \setminus B_l(i)}],
   \]
   which is the conditional expectation of \( g_i \) based on \( Z_i \) and all variables not in the same subset as \( Z_i \) in \( B_l \). Then, write:
   \[
   g_i - \mathbb{E}[g_i | Z_i] = \sum_{l=0}^{k-1} (g_i^l - g_i^{l+1}).
   \]
   To bound \( \sum_{i=1}^n g_i \), we use the triangle inequality:
   \[
   \left\| \sum_{i=1}^n g_i \right\|_p \leq \left\| \sum_{i=1}^n \mathbb{E}[g_i | Z_i] \right\|_p + \sum_{l=0}^{k-1} \left\| \sum_{i=1}^n (g_i^l - g_i^{l+1}) \right\|_p.
   \]
   The first term, \( \left\| \sum_{i=1}^n \mathbb{E}[g_i | Z_i] \right\|_p \), can be bounded through McDiarmid’s inequality as:
   \[
   \left\| \sum_{i=1}^n \mathbb{E}[g_i | Z_i] \right\|_p \leq 4 \sqrt{pn} M,
   \]
   since \( | \mathbb{E}[g_i | Z_i] | \leq M \) and \( \mathbb{E}[\mathbb{E}[g_i | Z_i]] = 0 \).
\item \emph{Step 3.3: Handling Conditional Sums with Bounded Differences.} For each partition \( B_l \), the terms \( g_i^l - g_i^{l+1} \) for \( i \in B_l \) are independent and zero-mean conditioned on \( Z_{[n] \setminus B_l} \). By the bounded differences property, we have:
   \[
   \| g_i^l - g_i^{l+1} \|_p \leq 2 \sqrt{p} 2^l \beta.
   \]
   Applying a concentration inequality (e.g., Marcinkiewicz-Zygmund \cite{ren2001best}), we find that:
   \[
   \left\| \sum_{i \in B_l} (g_i^l - g_i^{l+1}) \right\|_p \leq 6 \sqrt{2} p 2^l \beta.
   \]

Summing over all sets \( B_l \) in each partition and then over all levels \( l \) gives:
   \[
   \sum_{l=0}^{k-1} \left\| \sum_{i=1}^n (g_i^l - g_i^{l+1}) \right\|_p \leq 12 \sqrt{2} p n \beta \lceil \log_2 n \rceil.
   \]
Combining the bounds, we have:
   \[
   \left\| \sum_{i=1}^n g_i \right\|_p \leq 12 \sqrt{2} p n \beta \lceil \log_2 n \rceil + 4 M \sqrt{p n}.
   \]
\end{itemize}
~\\
\emph{Step 4: Conclusion and Final High-Probability Bound}
\begin{itemize}
    \item \emph{Using the Moment Bound to Derive a High-Probability Bound}: To turn this moment bound into a high-probability bound, we set \( p = \log \left( \frac{1}{\delta} \right) \). This choice yields:
   \[
   \left| \sum_{i=1}^n g_i \right| \leq O \left( n\gamma \log n \log \left( \frac{1}{\delta} \right) + L \sqrt{n \log \left( \frac{1}{\delta} \right)} \right).
   \]
   \item  \emph{Final Bound}: Combining this with the earlier decomposition, we conclude that, with probability at least \( 1 - \delta \), we have:
   \[
   |\mathcal{R}(\mathcal{A}_S) - \mathcal{R}_{n}(\mathcal{A}_S))| \leq O \left( \beta \log n \log \left( \frac{1}{\delta} \right) + L \sqrt{\log \left( \frac{1}{n}\frac{1}{\delta} \right)} \right).
   \]

This completes the proof framework for Theorem \ref{thm:stability}, providing a refined high-probability generalization bound for uniformly stable algorithms.
\end{itemize}

\textbf{PAC-Bayesian approach.} PAC-Bayesian theory was initially developed by McAllester \cite{mcallester1999pac} to explain Bayesian learning from the perspective of learning theory. Following the exposition in \cite{shalev2014understanding}, we assign a prior probability $P(h)$, or a probability density if
$\mathcal{H}$ is continuous, to each hypothesis $h\in\mathcal{H}$. By applying Bayesian reasoning, the learning process establishes a posterior distribution over $\mathcal{H}$, denoted as $Q.$ In a supervised learning scenario where $\mathcal{H}$ consists of functions mapping from $\mathcal{X}$ to $\mathcal{Y}$, with $\mathcal{Z}=\mathcal{X}\times\mathcal{Y}$. Then, $Q$ can be interpreted as
defining a randomized prediction rule. Specifically, for a unseen example $x$, a hypothesis $h\in\mathcal{H}$ is drawn according to $Q$, and the prediction $h(x)$ is then made. The loss associated with $Q$  is defined as:
$$\ell(Q,z)\stackrel{\mathrm{def}}{=}\underset{h\sim Q}{\mathbb{E}}[\ell(h,z)].$$
Then, both the generalization loss and the empirical training loss of
$Q$ can be expressed as follows:
$$\mathcal{R}(Q)\stackrel{\text{def}}{=}\underset{h\sim Q}{\mathbb{E}}\underset{z\sim \mathcal{D}}{\mathbb{E}}[\ell(h,z)]\quad\text{and}\quad \mathcal{R}_{n}(Q)\stackrel{\text{def}}{=}\underset{h\sim Q}{\mathbb{E}}\left[\frac{1}{n}\sum_{i=1}^n\ell(h,z_i)\right].$$

In accordance with \cite{shalev2014understanding}, we proceed to present the following generalization bound:
\begin{theorem}(PAC-Bayes Bounds \cite{shalev2014understanding})\label{thm: pac bayesian}
Let $P$ be a prior distribution over hypothesis space $\mathcal{H}$, and $\mathcal{D}$ be any distribution over an input-output pair space $\mathcal{Z}$.
For any $\delta\in(0,1)$ and  $Q$ over $\mathcal{H}$ (including those that depend on the training data $S=\{z_{1},\ldots,z_{n}\}$), we have:
$$\mathcal{R}(Q)\leq \mathcal{R}_{n}(Q)+\sqrt{\frac{KL(Q\|P)+\ln(n/\delta)}{2(n-1)}}$$
with a probability of at least $1-\delta$, where
$$KL(Q\|P)\stackrel{\text{def}}{=}\mathbb{E}_{h\sim Q}\left[\ln\frac{Q(h)}{P(h)}\right]$$
is the Kullback-Leibler (KL) divergence between the distributions $Q$ and $P.$
\end{theorem}
This theorem establishes a bound on the generalization error for the posterior $Q$, expressed in terms of the KL divergence between $Q$ and the prior $P$. These generalization bounds offer a promising way for interpreting the behavior of FMs associated with various optimization strategies, such as pretraining, fine-tuning and RLHF.

~\\
\textbf{Proof Sketch for PAC-Bayes Bounds in Theorem \ref{thm: pac bayesian}.}
~\\
\emph{Step 1: Establishing the Goal with Markov's Inequality}
\begin{itemize}
    \item \emph{Objective}: The goal is to bound \( L_D(Q) \), the true risk of \( Q \), using the empirical risk \( L_S(Q) \) plus an additional term that depends on the KL divergence between \( Q \) and a prior distribution \( P \) over \( H \).
   \item \emph{Applying Markov's Inequality}: For any function \( f(S) \), Markov's inequality provides that
   \[
   \mathbb{P}_S(f(S) \geq \epsilon) \leq \frac{\mathbb{E}_S[e^{f(S)}]}{e^{\epsilon}}.
   \]
   This inequality is used here to control the probability that \( f(S) \) exceeds a threshold \( \epsilon \), giving us a probabilistic upper bound.
\end{itemize}
~\\
\emph{Step 2: Defining the Risk Difference and Setting Up the Bounding Function}
\begin{itemize}
    \item \emph{Risk Difference \( \Delta(h) \)}: Define the risk difference for a hypothesis \( h \) as
   \[
   \Delta(h) = L_D(h) - L_S(h),
   \]
   where \( L_D(h) \) is the true risk and \( L_S(h) \) is the empirical risk. This difference quantifies the discrepancy between true and empirical risks, which we aim to bound.

  \item  \emph{Setting up \( f(S) \)}: Choose \( f(S) \) as the supremum over \( Q \) of a combination involving the mean squared risk difference \( \mathbb{E}_{h \sim Q} [\Delta(h)^2] \) and the KL divergence between \( Q \) and \( P \). Specifically, define
   \[
   f(S) = \sup_Q \left( 2(m - 1) \, \mathbb{E}_{h \sim Q} [\Delta(h)^2] - D(Q \| P) \right).
   \]
   This expression balances the risk difference term with the KL divergence, serving as a function we can bound probabilistically.
\end{itemize}
~\\
\emph{Step 3: Bounding \( \mathbb{E}_S[e^{f(S)}] \) by Using Jensen's Inequality and KL Divergence Properties}
\begin{itemize}
    \item \emph{Upper Bound on \( \mathbb{E}_S[e^{f(S)}] \)}: By analyzing \( \mathbb{E}_S[e^{f(S)}] \), the strategy is to express \( f(S) \) in terms of quantities independent of \( Q \). This simplifies the bound and makes it depend solely on the prior \( P \) rather than on the specific posterior \( Q \).
    \item \emph{Applying Jensen’s Inequality}: By Jensen's inequality and the convexity of the log function, we obtain an upper bound for \( \mathbb{E}_S[e^{f(S)}] \) in terms of expectations over \( P \). This step ensures that the KL divergence term contributes to controlling the growth of \( f(S) \).
\end{itemize}
~\\
\emph{Step 4: Using Hoeffding's Inequality to Bound the Expectation of Exponential Terms}
\begin{itemize}
    \item \emph{Applying Hoeffding’s Inequality}: For each hypothesis \( h \), apply Hoeffding’s inequality to the risk difference \( \Delta(h) \), ensuring that
   \[
   \mathbb{P}_S(\Delta(h) \geq \epsilon) \leq e^{-2m\epsilon^2}.
   \]
   This provides a concentration bound, which we use to control the tail behavior of the sum of risk differences over \( S \).
  \item  \emph{Combining Bounds}: Integrate the bound on \( \mathbb{E}_S[e^{f(S)}] \) with Hoeffding’s inequality to show that
   \[
   \mathbb{P}_S(f(S) \geq \epsilon) \leq \frac{m}{e^\epsilon}.
   \]
\end{itemize}
~\\
\emph{Step 5: Final Probability Bound and Rearrangement}
\begin{itemize}
    \item \emph{Setting \( \epsilon = \ln(m / \delta) \)}: Define \( \epsilon = \ln(m / \delta) \), so that with probability at least \( 1 - \delta \), we have
   \[
   2(m - 1) \, \mathbb{E}_{h \sim Q} [\Delta(h)^2] - D(Q \| P) \leq \ln(m / \delta).
   \]
    \item  \emph{Rearranging the Bound}: Using Jensen’s inequality again, we arrive at the final form:
   \[
   \left( \mathbb{E}_{h \sim Q} [\Delta(h)] \right)^2 \leq \mathbb{E}_{h \sim Q} [\Delta(h)^2] \leq \frac{\ln(m / \delta) + D(Q \| P)}{2(m - 1)}.
   \]
   Taking the square root on both sides gives:
   \[
   L_D(Q) \leq L_S(Q) + \sqrt{\frac{D(Q \| P) + \ln(m / \delta)}{2(m - 1)}}.
   \]
\end{itemize}
This final bound completes the proof, establishing that with high probability, the true risk \( L_D(Q) \) is bounded by the empirical risk \( L_S(Q) \) plus a term involving the KL divergence and sample size, capturing the generalization behavior under the PAC-Bayes framework.

\subsection{Expressive Power Analysis}
Expressive Power traditionally refers to the ability of a neural network to represent a wide and diverse range of functions. However, with the advent of FMs, the concept of expressive power has been further expanded. For instance, the use of prompts such as chain of thought can also influence a model's expressive capabilities. This concept is essential for evaluating the potential of FMs to capture complex structures and relationships within data. 

Understanding the expressive power of FMs is particularly important for interpreting both their capabilities and limitations. As FMs  are increasingly utilized in various applications, analyzing their expressive power helps us determine the range of tasks they can handle. This insight is crucial for identifying scenarios where these models are likely to excel or encounter difficulties, thereby guiding the development of more robust and versatile architectures. Moreover, a thorough examination of expressive power ensures that a model’s performance aligns with expectations, not only in traditional applications but also in novel and unforeseen scenarios.


Mathematically, expressive power can be defined in terms of a model's ability to approximate functions within a certain class. Let $\mathcal{H}$ denote the class of functions that a neural network can represent, parameterized by its architecture (e.g., depth, width, and activation functions). Consider a target function $f : {\mathbb{R}}^{d} \rightarrow  \mathbb{R}$ within a broader function space $\mathcal{F}$ , such as the space of continuous functions $C\left( X\right)$ over a compact domain $X \subseteq  {\mathbb{R}}^{d}$ . The hypothesis space $\mathcal{H}$ is said to have sufficient expressive power with respect to $\mathcal{F}$ if $\forall f \in  \mathcal{F}$ and  $\forall \epsilon  > 0$ , there exists a function $h \in  \mathcal{H}$ such that:
$$
\mathop{\sup }\limits_{{x \in  X}}\left| {f\left( x\right)  - h\left( x\right) }\right|  \leq  \epsilon 
$$
This condition signifies that the model $\mathcal{H}$ can approximate any function $f$ from the space $\mathcal{F}$ with arbitrary precision, given sufficient capacity. Therefore, expressive power serves as a critical metric for understanding the theoretical capabilities of neural networks, particularly in terms of their potential to  adapt, and perform complex tasks across various domains. 

 The commonly-used theoretical tools for analyzing the expressive power of models are as follows: 
\begin{itemize}
    \item{\textbf{Universal Approximation Theorem:}} The Universal Approximation Theorem states that, with a single hidden layer and finite number of neurons, a feedforward neural network  can approximate any continuous function on a compact input space to arbitrary accuracy, given a sufficiently large number of neurons. Specifically,  for a neural network with a single hidden layer containing $m$ neurons, the output $f(x)$ can be represented as 
$$
f(x)=\sum_{i=1}^mw_i^{(2)}\sigma(w_i^{(1)}x+b_i^{(1)})+b^{(2)},
$$
where $w^{(1)}_i$ is the weights connecting input to the 
$i$-th neuron in the hidden layer, $b_i^{(1)}$ is bias term of the $i$-th neuron in the hidden layer, and $\sigma$ represents a nonlinear activation function. Thus, given a continuous function \( g: \mathbb{R}^n \rightarrow \mathbb{R} \) and a compact set \( K \subset \mathbb{R}^n \), for any \( \epsilon > 0 \) and \( \delta > 0 \), there exists a neural network with a single hidden layer and a nonlinear activation function \( \sigma \) such that \[ |f(x) - g(x)| < \epsilon \] for all \( x \in K \), where \( f(x) \) is the output of the neural network.

\item \textbf{Expressivity in Higher Complexity Classes:} Expressivity in Higher Complexity Classes examines the ability of neural networks, particularly transformers using Chain-of-Thought prompting, to solve problems within higher computational complexity classes, such as $NC^1$ (log-space parallel computation) and $AC^0$ (constant-depth Boolean circuits). CoT enhances a network's ability to decompose and solve complex tasks, thereby increasing its expressiveness:
$$
NC^1=\{\text{problems solvable in log-space parallel computation}\}
$$
\end{itemize}
These theoretical tools play a critical role in deepening our understanding and evaluation of the expressive power of neural networks and transformer-based models.

\subsection{Dynamic Behavior Analysis}
Training dynamic behavior analysis is the study of how machine learning algorithms evolve over time during the training process, with a focus on understanding how parameter updates influence the convergence behavior, speed, and final quality of the solution. This analysis is crucial for optimizing the performance and reliability of machine learning models, particularly in complex scenarios involving large-scale data and deep architectures.

Following the work in \cite{sun2019optimization}, dynamic behavior analysis can be systematically divided into three key steps. First, it ensures that the algorithm begins running effectively and converges to a reasonable solution, such as a stationary point, by examining conditions that prevent issues like oscillation or divergence. Next, it focuses on the speed of convergence, aiming to accelerate the process and reduce the computational resources and time required for training. Techniques like adaptive learning rates or momentum may be used to enhance efficiency. Finally, the analysis ensures that the algorithm converges to a solution with a low objective value, ideally achieving a global minimum rather than settling for a local one. This step is crucial for securing a high-quality final solution that generalizes well to unseen data. 
Additionally, several fundamental tools used for the theoretical analysis of training dynamics are as follows:
\begin{itemize}
    \item \textbf{Neural Tangent Kernel (NTK):} The NTK is a theoretical framework that models the training dynamics of deep neural networks in the infinite-width limit. NTK approximates the training process by treating the network as a linear model with a fixed kernel, defined by
$$
\Theta\left(x, x^{\prime}\right)=\lim _{n \rightarrow \infty} \nabla_\theta f_\theta(x) \cdot \nabla_\theta f_\theta\left(x^{\prime}\right)
$$
where $f_\theta(x)$ is the network output, $\theta$ represents the network parameters, and $x$ and $x^{\prime}$ are input samples. The NTK captures the evolution of the network's output during gradient descent and is instrumental in understanding the expressive capabilities of neural networks.
\item \textbf{(Stochastic) Differential Equation}: Gradient flow represents the continuous-time analog of gradient descent, described by the differential equation:
$$
\frac{{d\theta }\left( t\right) }{dt} =  - {\nabla }_{\theta }R\left( {\theta \left( t\right) }\right) 
$$
This mathematical approach allows for a smooth analysis of the training dynamics, modeling how parameters evolve continuously over time and providing deeper insights into the convergence behavior of the algorithm.
\end{itemize}

Next, let us recall that our objective is to minimize the risk $\mathcal{R}(f)$ through optimization techniques. However, since the data distribution $\mathcal{D}$ is unknown, we cannot directly optimize this risk function. Instead, we approximate it by minimizing the empirical risk $\mathcal{R}_n(f)$ using methods such as gradient descent and stochastic gradient descent (SGD).

Gradient descent is an iterative optimization method where, at each step, we refine the solution by moving in the direction opposite to the gradient of the function at the current point. In contrast, SGD addresses this limitation by allowing the optimization process to proceed along a randomly selected direction, provided the expected direction aligns with the negative gradient.

Below, we will first introduce the fundamentals of the gradient descent algorithm and discuss the associated convergence analysis. The gradient of a differentiable function \( f : \mathbb{R}^d \to \mathbb{R} \) at a point \( \boldsymbol{\theta} \), represented as \( \nabla f(\boldsymbol{\theta}) \), is a vector consisting of \( f \)'s partial derivatives with respect to each component of \( \boldsymbol{\theta} \):
\[
\nabla f(\boldsymbol{\theta}) = \left( \frac{\partial f(\boldsymbol{\theta})}{\partial \theta_1}, \ldots, \frac{\partial f(\boldsymbol{\theta})}{\partial \theta_d} \right).
\]
Gradient descent proceeds iteratively, beginning from an initial point \( \boldsymbol{\theta} \), often set to zero (e.g., \( \boldsymbol{\theta}^{(1)} = 0 \)). In each iteration, the algorithm updates the current point by moving in the direction opposite to the gradient. Specifically, the update rule is:
\[
\boldsymbol{\theta}^{(t+1)} = \boldsymbol{\theta}^{(t)} - \eta \nabla f(\boldsymbol{\theta}^{(t)}),
\]
where \( \eta > 0 \) is the step size. Intuitively, because the gradient indicates the direction of the steepest ascent for \( f \) at \( \boldsymbol{\theta}^{(t)} \), taking a small step in the opposite direction reduces \( f \)'s value. After \( T \) iterations, the algorithm produces an output that approximates the minimum. We explore the convergence rate of gradient descent by using a convex and Lipschitz continuous function as an example.
\begin{theorem}(Convergence rate of GD for convex-Lipschitz functions \cite{shalev2014understanding})\label{thm: convergence GD}
Consider a convex function \( f \) that is \( \rho \)-Lipschitz, and let \( \boldsymbol{\theta}^\star \in \arg \min_{\{\boldsymbol{\theta} : \|\boldsymbol{\theta}\| \leq B\}} f(\boldsymbol{\theta}) \) denote an optimal solution within the constraint \( \|\boldsymbol{\theta}\| \leq B \). When we apply gradient descent to \( f \) for \( T \) iterations with a step size \( \eta = \sqrt{\frac{B^2}{\rho^2 T}} \), the resulting average vector \( \bar{\boldsymbol{\theta}}=\frac{1}{T}\sum_{t=1}^T\boldsymbol{\theta^{(t)}} \) satisfies the inequality
\[
f(\bar{\boldsymbol{\theta}}) - f(\boldsymbol{\theta}^\star) \leq \frac{B \rho}{\sqrt{T}}.
\]
Moreover, to ensure that \( f(\bar{\boldsymbol{\theta}}) - f(\boldsymbol{\theta}^\star) \leq \epsilon \) for any \( \epsilon > 0 \), it is sufficient to perform gradient descent for a number of iterations satisfying
\[
T \geq \frac{B^2 \rho^2}{\epsilon^2}.
\]
\end{theorem}
~\\
\textbf{Proof Sketch for Convergence Rate of Gradient Descent for Convex-Lipschitz Functions in Theorem \ref{thm: convergence GD}}

This proof framework establishes an upper bound on the convergence rate of gradient descent when applied to a convex function \( f \) that is also \( \rho \)-Lipschitz. Given a constraint on the norm of the optimal solution \( \|\boldsymbol{\theta}^\star\| \leq B \), the objective is to show that after \( T \) iterations with an appropriately chosen step size \( \eta \), the average iterate achieves a suboptimality gap that decreases at the rate \( O \left( \frac{1}{\sqrt{T}} \right) \).

~\\
\emph{Step 1: Definition of Suboptimality and Jensen's Inequality Application.}
\begin{itemize}
    \item \emph{Suboptimality Definition:} We aim to bound the suboptimality of the averaged iterate $\bar{\boldsymbol{\theta}}=\frac1T\sum_{t=1}^T\boldsymbol{\theta}^{(t)}$ with respect to the optimal solution $\boldsymbol{\theta}^\star$,i.e.,
$$
f(\bar{\boldsymbol{\theta}})-f(\boldsymbol{\theta}^\star).
$$
\item \emph{Jensen's Inequality:} Since $f$ is convex, we can apply Jensen's inequality to show:
$$
f(\bar{\boldsymbol{\theta}})-f(\boldsymbol{\theta}^\star)=f\left(\frac1T\sum_{t=1}^T\boldsymbol{\theta}^{(t)}\right)-f(\boldsymbol{\theta}^\star)\leq\frac1T\sum_{t=1}^T\left(f(\boldsymbol{\theta}^{(t)})-f(\boldsymbol{\theta}^\star)\right).
$$
This reduces the problem to bounding the average of the suboptimalities $f(\boldsymbol{\theta}^{(t)})-f(\boldsymbol{\theta}^{\star})$ over the iterations.
\end{itemize}
~\\
\emph{Step 2: Utilizing Convexity to Relate Suboptimality and Gradients
.}
\begin{itemize}
    \item \emph{Convexity Inequality:} By convexity of \( f \), for each \( t \),
\[
f(\boldsymbol{\theta}^{(t)}) - f(\boldsymbol{\theta}^\star) \leq \langle \boldsymbol{\theta}^{(t)} - \boldsymbol{\theta}^\star, \nabla f(\boldsymbol{\theta}^{(t)}) \rangle.
\]
This allows us to bound the suboptimality at each iteration in terms of the inner product of the difference \( \boldsymbol{\theta}^{(t)} - \boldsymbol{\theta}^\star \) with the gradient \( \nabla f(\boldsymbol{\theta}^{(t)}) \).

\item \emph{Reduction to Summing Gradient Inner Products:} Combining the above results, we find that
\[
f(\bar{\boldsymbol{\theta}}) - f(\boldsymbol{\theta}^\star) \leq \frac{1}{T} \sum_{t=1}^T \langle \boldsymbol{\theta}^{(t)} - \boldsymbol{\theta}^\star, \nabla f(\boldsymbol{\theta}^{(t)}) \rangle.
\]
Define \( \boldsymbol{v}_t = \nabla f(\boldsymbol{\theta}^{(t)}) \) for convenience.
\end{itemize}
~\\
\emph{Step 3: Expanding Inner Products via the Update Rule.}
\begin{itemize}
    \item \emph{Inner Product Expansion:} Observe that
\begin{align}
&\langle \boldsymbol{\theta}^{(t)} - \boldsymbol{\theta}^\star, \boldsymbol{v}_t \rangle = \frac{1}{\eta} \langle \boldsymbol{\theta}^{(t)} - \boldsymbol{\theta}^\star, \eta \boldsymbol{v}_t \rangle \notag\\
&= \frac{1}{2 \eta} \left( -\|\boldsymbol{\theta}^{(t)} - \boldsymbol{\theta}^\star - \eta \boldsymbol{v}_t\|^2 + \|\boldsymbol{\theta}^{(t)} - \boldsymbol{\theta}^\star\|^2 + \eta^2 \|\boldsymbol{v}_t\|^2 \right).\notag
\end{align}
Using the definition of the update rule \( \boldsymbol{\theta}^{(t+1)} = \boldsymbol{\theta}^{(t)} - \eta \boldsymbol{v}_t \), we can rewrite this as:
\[
\langle \boldsymbol{\theta}^{(t)} - \boldsymbol{\theta}^\star, \boldsymbol{v}_t \rangle = \frac{1}{2 \eta} \left( -\|\boldsymbol{\theta}^{(t+1)} - \boldsymbol{\theta}^\star\|^2 + \|\boldsymbol{\theta}^{(t)} - \boldsymbol{\theta}^\star\|^2 \right) + \frac{\eta}{2} \|\boldsymbol{v}_t\|^2.
\]
\end{itemize}
~\\
\emph{Step 4: Summing Over Iterations and Telescoping}
\begin{itemize}
    \item \emph{Telescoping Sum}: Summing this inequality over \( t = 1, \ldots, T \), we get:
   \begin{align}
  & \sum_{t=1}^T \langle \boldsymbol{\theta}^{(t)} - \boldsymbol{\theta}^\star, \boldsymbol{v}_t \rangle \notag\\
  &= \frac{1}{2 \eta} \sum_{t=1}^T \left( -\|\boldsymbol{\theta}^{(t+1)} - \boldsymbol{\theta}^\star\|^2 + \|\boldsymbol{\theta}^{(t)} - \boldsymbol{\theta}^\star\|^2 \right) + \frac{\eta}{2} \sum_{t=1}^T \|\boldsymbol{v}_t\|^2.
   \end{align}
   The first term on the right side is a telescoping sum that simplifies to:
   \[
   \|\boldsymbol{\theta}^{(1)} - \boldsymbol{\theta}^\star\|^2 - \|\boldsymbol{\theta}^{(T+1)} - \boldsymbol{\theta}^\star\|^2.
   \]
  \item  \emph{Simplified Bound}: Substituting back, we get:
   \[
   \sum_{t=1}^T \langle \boldsymbol{\theta}^{(t)} - \boldsymbol{\theta}^\star, \boldsymbol{v}_t \rangle \leq \frac{1}{2 \eta} \|\boldsymbol{\theta}^{(1)} - \boldsymbol{\theta}^\star\|^2 + \frac{\eta}{2} \sum_{t=1}^T \|\boldsymbol{v}_t\|^2.
   \]
   Assuming \( \boldsymbol{\theta}^{(1)} = 0 \), we further simplify to:
   \[
   \sum_{t=1}^T \langle \boldsymbol{\theta}^{(t)} - \boldsymbol{\theta}^\star, \boldsymbol{v}_t \rangle\leq \frac{1}{2 \eta} \|\boldsymbol{\theta}^\star\|^2 + \frac{\eta}{2} \sum_{t=1}^T \|\boldsymbol{v}_t\|^2.
   \]
\end{itemize}
~\\
\emph{Step 5: Choosing Step Size and Final Convergence Rate}
\begin{itemize}
    \item \emph{Step Size Selection}: Set \( \eta = \sqrt{\frac{B^2}{\rho^2 T}} \) so that each gradient \( \|\boldsymbol{v}_t\| \leq \rho \) results in
   \[
   \frac{1}{T} \sum_{t=1}^T \langle \boldsymbol{\theta}^{(t)} - \boldsymbol{\theta}^\star, \boldsymbol{v}_t \rangle \leq \frac{B \rho}{\sqrt{T}}.
   \]
\item \emph{Final Bound}: This choice of step size ensures that after \( T \) iterations,
   \[
   f(\bar{\boldsymbol{\theta}}) - f(\boldsymbol{\theta}^\star) \leq \frac{B \rho}{\sqrt{T}}.
   \]
   To achieve an error \( \epsilon \), it is sufficient to perform \( T \geq \frac{B^2 \rho^2}{\epsilon^2} \) iterations.

\end{itemize}
In SGD, the update direction does not need to precisely align with the true gradient. Instead, we permit the update direction to be a random vector, requiring only that its expected value aligns with the gradient direction at each iteration. More generally, this expectation must be consistent with the subgradient of the function at the current point in the optimization process. The update for \( \boldsymbol{\theta}^{(t+1)} \) is:
\[
\boldsymbol{\theta}^{(t+1)} = \boldsymbol{\theta}^{(t)} - \eta \boldsymbol{v}_t,
\]
where \( \eta \) is the step size, and \( \boldsymbol{v}_t \) is chosen randomly from a distribution such that \( \mathbb{E}[\boldsymbol{v}_t \mid \boldsymbol{\theta}^{(t)}] \in \partial f(\boldsymbol{\theta}^{(t)}) \), ensuring that, on average, the update direction remains within the subdifferential of \( f \) at \( \boldsymbol{\theta}^{(t)} \). Given that the expected value of \( \boldsymbol{v}_t \) corresponds to the subgradient of \( f \) at \( \mathbf{w}^{(t)} \), it remains possible to establish a bound on the expected outcome of stochastic gradient descent, similar to that in Theorem \ref{thm: convergence GD}. This result is rigorously formalized in the following theorem.
\begin{theorem}(Convergence rate of SGD for convex-Lipschitz functions \cite{shalev2014understanding})\label{thm: convergence of SGD}
Let \( B, \rho > 0 \), and let \( f \) be a convex and $\rho$-Lipschitz function with an optimal solution \( \boldsymbol{\theta}^\star \in \arg \min_{\|\boldsymbol{\theta}\| \leq B} f(\boldsymbol{\theta}) \). Suppose that we run SGD for \( T \) iterations, using a step size \( \eta = \sqrt{\frac{B^2}{\rho^2 T}} \), and assume that \( \|\boldsymbol{v}_t\| \leq \rho \) holds with probability 1 for each iteration \( t \). Then we have
\[
\mathbb{E}[f(\bar{\boldsymbol{\theta}})] - f(\boldsymbol{\theta}^\star) \leq \frac{B \rho}{\sqrt{T}}.
\]
Thus, to ensure \( \mathbb{E}[f(\bar{\boldsymbol{\theta}})] - f(\boldsymbol{\theta}^\star) \leq \epsilon \) for any \( \epsilon > 0 \), it is sufficient to run the SGD algorithm for a number of iterations satisfying
\[
T \geq \frac{B^2 \rho^2}{\epsilon^2}.
\]
\end{theorem}
~\\
\textbf{Proof Sketch for Convergence Rate of SGD for Convex-Lipschitz Functions in Theorem \ref{thm: convergence of SGD}.}
~\\
\emph{Step 1: Taking Expectation of the Suboptimality Bound}
\begin{itemize}
    \item \emph{Expectation of Suboptimality}: The goal is to bound the expected suboptimality of the averaged iterate \( \bar{\boldsymbol{\theta}} = \frac{1}{T} \sum_{t=1}^T \boldsymbol{\theta}^{(t)} \) with respect to the optimal solution \( \boldsymbol{\theta}^\star \), i.e.,
   \[
   \mathbb{E}[f(\bar{\boldsymbol{\theta}})] - f(\boldsymbol{\theta}^\star).
   \]
   \item  \emph{Applying Proof framework of Theorem \ref{thm: convergence GD}}: For any sequence \( \boldsymbol{v}_1, \boldsymbol{v}_2, \ldots, \boldsymbol{v}_T \), we can obtain a bound on the inner product of the gradient with the difference \( \boldsymbol{\theta}^{(t)} - \boldsymbol{\theta}^\star \). By taking expectations, we find:
   \[
   \mathbb{E}_{\boldsymbol{v}_{1:T}} \left[ \frac{1}{T} \sum_{t=1}^T \langle \boldsymbol{\theta}^{(t)} - \boldsymbol{\theta}^\star, \boldsymbol{v}_t \rangle \right] \leq \frac{B \rho}{\sqrt{T}}. 
   \]
\end{itemize}
~\\
\emph{Step 2: Establishing a Comparison Between Expected Suboptimality and Gradient Inner Products.}
\begin{itemize}
    \item \emph{Expected Suboptimality Inequality}: It remains to show that
   \[
   \mathbb{E}_{\boldsymbol{v}_{1:T}} \left[ \frac{1}{T} \sum_{t=1}^T \left( f(\boldsymbol{\theta}^{(t)}) - f(\boldsymbol{\theta}^\star) \right) \right] \leq \mathbb{E}_{\boldsymbol{v}_{1:T}} \left[ \frac{1}{T} \sum_{t=1}^T \langle \boldsymbol{\theta}^{(t)} - \boldsymbol{\theta}^\star, \boldsymbol{v}_t \rangle \right].
   \]
\item  \emph{Using Linearity of Expectation}: By the linearity of expectation, we rewrite the right-hand side as:
   \[
   \mathbb{E}_{\boldsymbol{v}_{1:T}} \left[ \frac{1}{T} \sum_{t=1}^T \langle \boldsymbol{\theta}^{(t)} - \boldsymbol{\theta}^\star, \boldsymbol{v}_t \rangle \right] = \frac{1}{T} \sum_{t=1}^T \mathbb{E}_{\boldsymbol{v}_{1:T}} \left[ \langle \boldsymbol{\theta}^{(t)} - \boldsymbol{\theta}^\star, \boldsymbol{v}_t \rangle \right].
   \]
\end{itemize}
~\\
\emph{Step 3: Applying the Law of Total Expectation.}
\begin{itemize}
    \item \emph{Law of Total Expectation}: For two random variables \( \alpha \) and \( \beta \), and a function \( g \), the law of total expectation states \( \mathbb{E}_\alpha [g(\alpha)] = \mathbb{E}_\beta \mathbb{E}_\alpha [g(\alpha) | \beta] \). Applying this with \( \alpha = \boldsymbol{v}_{1:t} \) and \( \beta = \boldsymbol{v}_{1:t-1} \), we find:
   \[
   \mathbb{E}_{\boldsymbol{v}_{1:T}} \left[ \langle \boldsymbol{\theta}^{(t)} - \boldsymbol{\theta}^\star, \boldsymbol{v}_t \rangle \right] = \mathbb{E}_{\boldsymbol{v}_{1:t-1}} \left\langle \boldsymbol{\theta}^{(t)} - \boldsymbol{\theta}^\star, \mathbb{E}_{\boldsymbol{v}_t} \left[ \boldsymbol{v}_t \mid \boldsymbol{v}_{1:t-1} \right] \right\rangle.
   \]
Since \( \boldsymbol{\theta}^{(t)} \) depends only on \( \boldsymbol{v}_{1:t-1} \), once we condition on \( \boldsymbol{v}_{1:t-1} \), the vector \( \boldsymbol{\theta}^{(t)} \) becomes deterministic. 
\end{itemize}
~\\
\emph{Step 4: Applying the SGD Condition on Expected Gradients.}
\begin{itemize}
    \item \emph{SGD Condition on Gradients:} The property of SGD ensures that \( \mathbb{E}_{\boldsymbol{v}_t}[\boldsymbol{v}_t \mid \boldsymbol{\theta}^{(t)}] \in \partial f(\boldsymbol{\theta}^{(t)}) \), meaning \( \mathbb{E}_{\boldsymbol{v}_t}[\boldsymbol{v}_t \mid \boldsymbol{v}_{1:t-1}] \in \partial f(\boldsymbol{\theta}^{(t)}) \).
   
 \item \emph{Convexity and Expected Suboptimality:} Since \( f \) is convex, we have:
\[
\langle \boldsymbol{\theta}^{(t)} - \boldsymbol{\theta}^\star, \mathbb{E}_{\boldsymbol{v}_t} \left[ \boldsymbol{v}_t \mid \boldsymbol{v}_{1:t-1} \right] \rangle \geq f(\boldsymbol{\theta}^{(t)}) - f(\boldsymbol{\theta}^\star).
\]
Taking the expectation with respect to \( \boldsymbol{v}_{1:t-1} \), we conclude that:
\[
\mathbb{E}_{\boldsymbol{v}_{1:t}} \left[ \langle \boldsymbol{\theta}^{(t)} - \boldsymbol{\theta}^\star, \boldsymbol{v}_t \rangle \right] \geq \mathbb{E}_{\boldsymbol{v}_{1:t-1}} \left[ f(\boldsymbol{\theta}^{(t)}) - f(\boldsymbol{\theta}^\star) \right].
\]

\end{itemize}
~\\
\emph{Step 5: Concluding the Proof with Expected Suboptimality Bound}.
\begin{itemize}
    \item \emph{Summing Over All Iterations:} Summing over \( t \), dividing by \( T \), and applying the linearity of expectation yields:
   \[
   \mathbb{E}_{\boldsymbol{v}_{1:T}} \left[ f(\boldsymbol{\theta}^{(t)}) - f(\boldsymbol{\theta}^\star) \right] \leq \mathbb{E}_{\boldsymbol{v}_{1:T}} \left[ \frac{B \rho}{\sqrt{T}} \right].
   \]
\end{itemize}
The proof framework is complete.

\section{Interpreting Inference Capabilities of Foundation Models}\label{sec3}
In this section, we aim to leverage the interpretable methods discussed in Section \ref{sec2}  to interpret the fundamental causes driving impressive inference capability of FMs, including ICL, CoT, and adaptability to distribution shifts.

\subsection{In-Context Learning}\label{section_ICL}

As models and their training datasets have grown in complexity, FMs have demonstrated remarkable capabilities, notably in ICL \cite{brown2020language}. ICL allows models to learn from a few demonstrations provided in the input context, enabling them to handle complex tasks without parameter updates, as seen in mathematical reasoning problems \cite{wei2022chain}. Unlike supervised learning, ICL leverages pretrained FMs for prediction without requiring a dedicated training phase. Formally, ICL is defined as follows.

\begin{definition} In a formal framework, the process takes a query input $x$ and a set of candidate answers $Y=\left\{y_1, \ldots, y_n\right\}$, where $Y$ may represent class labels or a collection of free-text expressions. A pre-trained model $f_M$ generates the predicted answer by conditioning on a demonstration set $C_n$ alongside the query input $x$. The demonstration set \( C_n \) encompasses  \( n \) demonstration examples \( C_n = \{ (x_1, y_1), . . . , (x_n, y_n)\} \), where each \( (x_k, y_k), k=1,\cdots, n, \) represents an in-context example aligned with the task at hand.  The answer \( \hat{y} \) is modeled by a pre-trained function \( f_M \), defined as:
\[ 
\hat{y}= f_M(C_n, x;\theta),
\]
where $\theta$ is the parameter of the pre-trained model. The function \( f_M \) outputs the current answer given the demonstration and the query input. 
\end{definition}
Based upon the above definition, the generalization of FMs under ICL can be measured by the upper bound on population risk. In the most common practical setting, where predictions are made at every position, the population risk is defined as follows:
$$
\mathcal{R}(f_M)=\frac{1}{n}\left[\sum_{k=1}^{n}\mathbb{E}_{C_{k},(x_{k+1},y_{k+1})} \ell\big(f_M(C_k, x_{k+1};\theta),y_{k+1}\big)\right],
$$
Moreover, for the scenario where predictions are made after observing a fixed length of demonstration examples, the population risk is defined as follows:
$$
\mathcal{R}_{fixed}(f_M)=\mathbb{E}_{C_{n},(x_{n+1},y_{n+1})} \ell\big(f_M(C_n, x_{n+1};\theta),y_{n+1}\big),
$$
where $\ell$ is the loss function and $C_k=\{(x_1,y_1),\cdots,(x_k,y_k)\}$. Despite the impressive performance of FMs with ICL prompts, significant confusions persist from a layman, user or even researcher perspective. These challenges primarily revolve around the sensitivity of ICL to minor prompt variations and the difficulty of constructing effective in-context examples to achieve desired outputs.

\begin{figure}[htbp]
    \centering




\includegraphics[width=\textwidth]{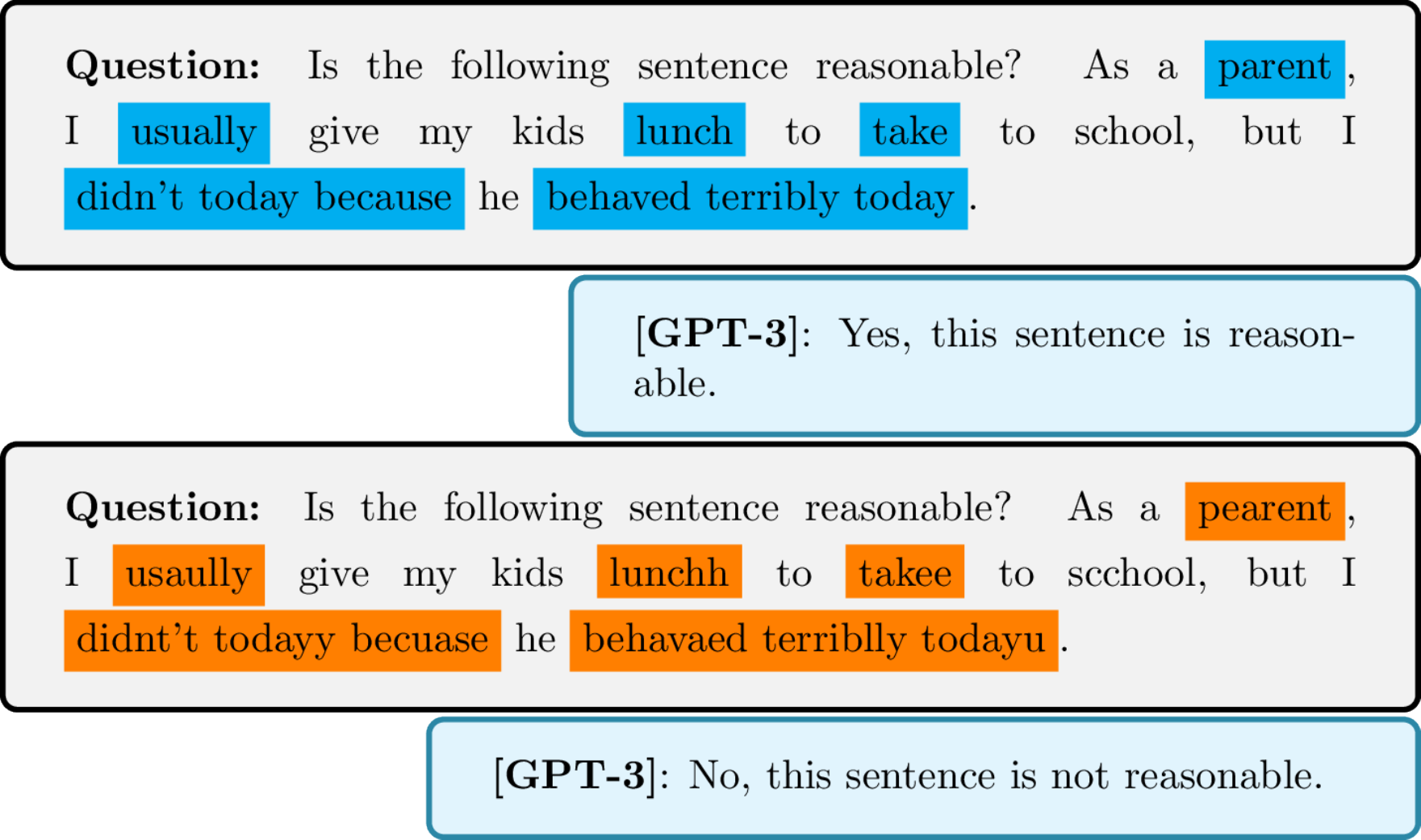}
\caption{Different Responses by GPT-3 to Identical Questions with Typos \cite{liu2023trustworthy}.}\label{fig_ICL_1}
\end{figure}

\textbf{Sensitivity to Prompt Variations.} A critical challenge  is understanding why small differences in the input can produce significantly different outputs. For instance, a prompt containing typographical errors or slight changes in structure might not only confuse the model but can also result in drastically inaccurate responses. This phenomenon reflects the underlying mechanics of FMs, which heavily depend on precise token sequences for accurate interpretation. As shown in Fig. \ref{fig_ICL_1}, a simple prompt with minor errors, such as incorrect spelling, can cause the model to misinterpret the entire context, thereby affecting the final output.

\begin{figure}[htbp]
    \centering



\includegraphics[width=\textwidth]{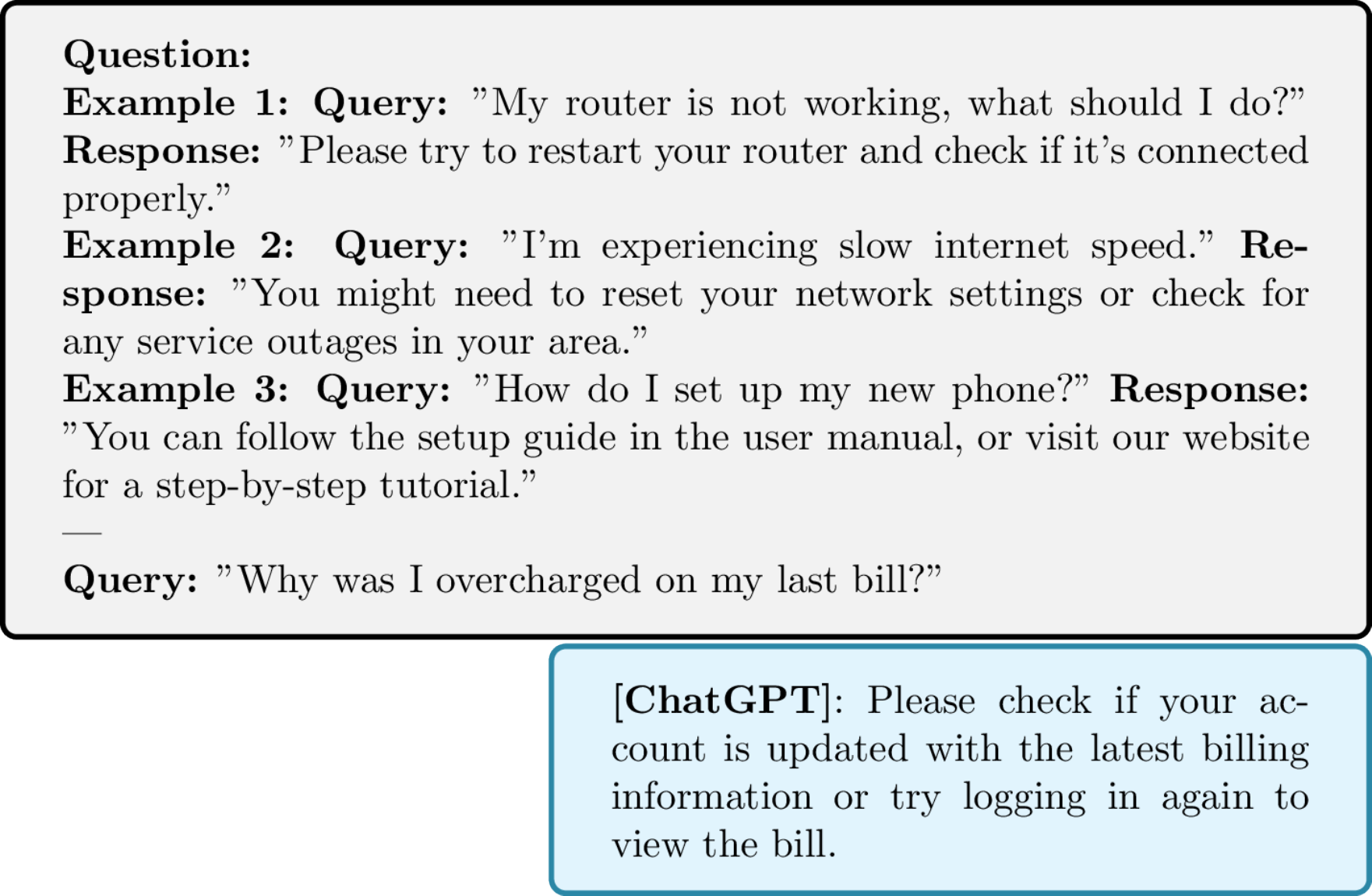}
\caption{Mismatch Between Context Examples and Query Leading to Misinterpretations and Erroneous Outputs}\label{fig_ICL2}
\end{figure}

\textbf{Constructing Effective In-Context Examples.} Another challenge lies in constructing effective in-context examples that align with the user's expectations. Users often struggle with determining the appropriate length and specificity of examples to include in the prompt. The relevance and level of detail in these examples are critical factors that directly impact the quality and accuracy of the model's responses. For instance, as illustrated in Fig. \ref{fig_ICL2}, when a FM is used to automatically generate customer service responses, providing examples that primarily focus on technical support may lead to errors when the query is about billing issues. In such cases, the model might incorrectly categorize the billing query as a technical problem, resulting in an irrelevant or inaccurate response. Therefore, users must carefully craft examples that provide enough detail to guide the model accurately and ensure that these examples are directly aligned with the nature of the query.

To address these concerns,  the analyses of generalization and expressivity can be utilized to offer insights into  optimizing the responses of  FMs by adjusting the structure and content of input examples within ICL.

\subsubsection{Interpreting Through Generalization Analysis}




Recently, various theoretical perspectives on the generalization of FMs under ICL have been explored \cite{liu2024towards,lintransformers,li2023transformers,zhang2023and,li2024nonlinear,wies2024learnability,wu2023many,bai2024transformers,jeoninformation}. Since FMs rely on the context provided during inference to perform tasks, these theoretical frameworks offer valuable insights into how models generalize to novel tasks without requiring additional training. They highlight the pivotal importance of model architecture, pretraining methodologies, and input context structure in driving this generalization. In this section, we review and interpret relevant studies on FMs from the perspective of generalization error, focusing on several key approaches. One line of work explores the dynamics of gradient flow over the loss function. Another major strand of research applies statistical learning principles, including algorithmic stability, covering numbers, and PAC-Bayes theories. These methods offer different avenues for understanding the fundamental mechanisms behind in-context learning and model generalization.

In this line of research examining gradient flow behavior over the loss function, the primary focus remains on linear problem settings. To rigorously analyze gradient flow dynamics, studies like \cite{wu2023many} often simplify the transformer architecture to a single-layer attention mechanism with linear activation functions. Specifically, \cite{wu2023many} assumes that the parameters of a linear task are drawn from a Gaussian prior, $N(0, \psi^2 \mathbf{I}_d)$, and the prompt $C_n = {(x_1, y_1), \dots, (x_n, y_n)}$ is an independent copy of the testing sample $(x, y)$, where $x \sim N(0, \mathbf{H})$, $\mathbf{H} \succeq 0$, and $y \sim N(\beta^T x, \sigma^2)$. Furthermore, they consider a pretraining dataset where each of the $T$ independent regression tasks is associated with $n+1$ in-context prompt examples. The generalization is defined as 
$$
\mathcal{R}_{fixed}(f_M) = \mathbb{E}[(y - f_M(C_n, x;\theta))^2],
$$ 
establishing a sample complexity bound for the pretraining of a linear parameterized single-layer attention model within the framework of linear regression under a Gaussian prior. The theorem is stated as follows:

\begin{theorem}(Task complexity for pertaining \cite{wu2023many}).\label{theo_icl_3.2} Let 
$n \geq 0$ represent the quantity of in-context examples utilized to generate the dataset. Let $\theta_T$ denote the parameters output by SGD on the pretraining dataset with a geometrically decaying step size. Assume that the initialization $\theta_0$ commutes with $\mathbf{H}$ and $\gamma_0 \leq 1 /(c \operatorname{tr}(\mathbf{H}) \operatorname{tr}(\tilde{\mathbf{H}}_n))$, where $c>1$ is an absolute constant and $\tilde{\mathbf{H}}_n$ is defined as 
$$
\psi^{2}\mathbf{H}((\mathrm{tr}(\mathbf{H})/n+\sigma^{2}/n\psi^{2})\mathbf{I}+(n+1)/n\mathbf{H}).
$$
Then we have
\begin{align}
&\mathbb{E}\left[ \mathcal{R}_{fixed}(f_M(C_n,x;\theta_T)-\operatorname*{min}_{\theta}\mathcal{R}_{fixed}(f_M( C_n,x;\theta))\right]\nonumber\\
&\lesssim\left\langle\mathbf{H}\tilde{\mathbf{H}}_n, \left(\prod_{t=1}^T\left(\mathbf{I}-\gamma_t\mathbf{H}\tilde{\mathbf{H}}_n\right)\theta_n^*\right)^2\right\rangle+\left(\psi^2\mathbf{tr}(\mathbf{H})+\sigma^2\right)\frac{D_{\text{eff}}}{T_{\text{eff}}}. \label{icl-task}
\end{align}
where the effective number of tasks and effective dimension are given by
$$
T_{\text {eff }}:=\frac{T}{\log (T)}, \quad D_{\text {eff }}:=\sum_i \sum_j \min \left\{1, T_{\text {eff }}^2 \gamma_0^2 \lambda_i^2 \tilde{\lambda}_j^2\right\},
$$
respectively. Here, $\left(\lambda_i\right)_{i \geq 1}$ and $(\tilde{\lambda}_j)_{j \geq 1}$ represent the eigenvalues of $\mathbf{H}$ and $\tilde{\mathbf{H}}_n$, which satisfy the following conditions:
$$
\tilde{\lambda}_j:=\psi^2 \lambda_j\left(\frac{\operatorname{tr}(\mathbf{H})+\sigma^2 / \psi^2}{n}+\frac{n+1}{n} \lambda_j\right), \quad j \geq 1.
$$
\end{theorem}
The theorem above demonstrates that SGD-based pretraining can efficiently recover the optimal parameter $\theta_n^*$ given a sufficiently large $T$. Referring to the first term in Equation \ref{icl-task}, the initial component represents the error associated with applying gradient descent directly to the population ICL risk $\mathcal{R}_{fixed}(f_M)$, which decays exponentially. However, given the finite number of pretraining tasks, directly minimizing the population ICL risk is not possible, and the second term in Equation \ref{icl-task} accounts for the uncertainty (variance) arising from  pretraining on a finite number of $T$ distinct tasks. As the effective dimension of the model shrinks compared to the number of tasks, the variance also decreases. Importantly, the initial step size $\gamma_0$ balances the two terms: increasing the initial step size reduces the first term while amplifying the second, whereas decreasing it has the opposite effect. Their theoretical analysis indicates that the attention model can be efficiently pre-trained using a number of linear regression tasks independent of the dimension. Furthermore, they show that the pre-trained attention model performs as a Bayes-optimal predictor when the context length during inference closely matches that used in pretraining.

Additionally, \cite{liu2024towards} revisits the training of transformers for linear regression, focusing on a bi-objective task that involves predicting both the mean and variance of the conditional distribution.   They investigate scenarios where transformers operate under the constraint of a finite context window capacity $S$ and explore the implications of making predictions at every position. This leads to the derivation of a generalization bound expressed as
$$\mathcal{R}(f_M(C,x;\hat{\theta}))-\min_\theta\mathcal{R}(f_M(C,x;\theta))\lesssim \sqrt{\min \{ S, |C|\} /\left( {n|C|}\right) }, $$
 where $|C|$ denotes the length of the demonstration examples. Their analysis utilizes the structure of the context window to formulate a Markov chain over the prompt sequence and derive an upper bound for its mixing time.
 
Another significant line of research employs statistical learning theory techniques, including covering numbers, PAC-Bayes arguments, and algorithm stability \cite{bai2024transformers, lintransformers, zhang2023and, li2023transformers}. Specifically, \cite{bai2024transformers} and \cite{lintransformers} utilize chaining arguments with covering numbers to derive generalization bounds. In \cite{bai2024transformers}, the authors employ standard uniform concentration analysis via covering numbers to establish an excess risk guarantee. Specifically, their technique adopts the chaining-based proof framework detailed in Theorem \ref{theorem_chaining} of Section \ref{sec. preliminary}. They investigate key architectural parameters, including embedding dimensions $D$, the number of layers $L$, the number of attention heads $M$, feedforward width $D'$, and the norm of parameters $\|\theta\|\leq B$. Furthermore, they explore the use of $T$ prompts where each has $|C_{n+1}|=n+1$ in-context examples  for pretraining, leading to the derivation of the following excess risk bound:

\begin{theorem}(Excess risk of Pretraining \cite{bai2024transformers})\label{theo_icl_3.3}. With probability at least $1 - \delta$ over the pretraining instances, the ERM solution $\hat{\theta}$ to satisfies
\begin{align}
&\mathcal{R}_{fixed}(f_M(C_n,x;\hat{\theta}))-\operatorname*{min}_{\theta}\mathcal{R}_{fixed}(f_M(C_n,x;\theta)) \notag \\
&~~~~~~~~~~\lesssim \sqrt{\frac{L^2\left(M D^2+D D^{\prime}\right) \log B}{T}}.\notag
\end{align}
Furthermore, regarding sparse linear models such as LASSO, under certain conditions,
it can be demonstrated that with probability at least $1-\delta$ over the training instances, the solution $\hat{\theta}$ of the ERM problem, with input $x\in\mathbb{R}^d$, possessing $L=\mathcal{O}(1+d/n)$ layers, $M=2$ attention heads, $D^\prime=2d$, and $\sigma$ represents the
standard deviation of the labels $y$ in the Gaussian prior of the in-context data distribution,
ultimately achieves a smaller excess ICL risk, i.e.,
$$
\mathcal{R}_{fixed}(f_M(C_n,x;\hat{\theta}))\lesssim \sigma^2+\left(\sqrt{\frac{d^4}{T}}+\frac{s\log d}{n}\right).
$$
\end{theorem}
It is important to highlight that the theoretical result in Theorem \ref{theo_icl_3.3} of research \cite{bai2024transformers} deviates from the standard transformer architecture by substituting the softmax activation in the attention layers with a normalized ReLU function. This differs from the findings in Theorem
\ref{theo_icl_3.2} of research \cite{wu2023many}, which adheres to the standard architecture. Furthermore, \cite{lintransformers} also employs the standard transformer architecture with softmax activation, focusing on a more specific context involving sequential decision-making problems. It utilizes covering arguments, as demonstrated within the proof framework of Theorem \ref{theorem_chaining}, to derive a generalization upper bound of $\mathcal{O}(1/\sqrt{n})$ for the average regret.

Additionally, \cite{zhang2023and} investigate the pretraining error and establish a connection between the pretraining error and the ICL regret. Their technique primarily builds upon the PAC-Bayes proof framework we outlined for Theorem \ref{thm: pac bayesian}. Their findings demonstrate that the total variation distance between the trained model and a reference model is bounded as follows:
\begin{theorem}(Total Variation Distance \cite{zhang2023and}) Let $\mathbb{P}_\theta$ represent the probability distribution induced by the transformer with parameter $\theta$. Additionally, the model $\mathbb{P}_{\hat{\theta}}$ is pretrained by the algorithm:
$$
\hat{\theta}=\underset{\theta \in \Theta}{\operatorname{argmin}}-\frac{1}{n} \sum_{t=1}^{n-1} \log \mathbb{P}_\theta\left(\boldsymbol{x}_{t+1}^n \mid C_t\right).
$$
Furthermore, they consider the realizable setting, where ground truth probability distribution $\mathbb{P}(\cdot \mid C_n)$ and $\mathbb{P}_{\theta^*}(\cdot \mid C_n)$ are consistent for some $\theta^* \in \Theta$. Then, with probability at least $1-\delta$, the following inequality holds:
\begin{align}
\operatorname{TV}\left(\mathbb{P}(\cdot \mid C_n), \mathbb{P}_{\hat{\theta}}(\cdot \mid C_n)\right) \lesssim \frac{1}{n^{1/2}}\log (1+n)+\frac{1}{n^{1/4}}\log (1/\delta),
\end{align}
where $\operatorname{TV}$ denotes the total variation (TV) distance.
\end{theorem}
This result is achieved through a concentration argument applied to Markov chains, providing a quantifiable measure of model deviation during pretraining. They also interpret a variant of the attention mechanism as embedding Bayesian Model Averaging (BMA) within its architecture, enabling the transformer to perform ICL through prompting.

Further, we expand on the use of stability analysis to interpret the generalization  of FMs during ICL prompting. Recently, there has been growing interest in formalizing ICL as an algorithm learning problem, with the transformer model implicitly forming a hypothesis function during inference. This perspective has inspired the development of generalization bounds for ICL by associating the excess risk with the stability of the algorithm executed by the transformer \cite{li2023transformers, liu2024towards, deoraoptimization}.

Several studies have investigated how the architecture of transformers, particularly the attention mechanism, adheres to stability conditions \cite{liu2024towards,deoraoptimization}. These works characterize when and how transformers can be provably stable, offering insights into their generalization. For instance, it has been shown that under certain conditions, the attention mechanism in transformers can ensure that the model's output remains stable even when small changes are introduced in the training data \cite{li2023transformers}. This stability, in turn, translates into better generalization performance, making transformers a powerful tool in various applications requiring robust predictions from complex and high-dimensional data.

Next, we demonstrate that, during ICL prompting, a multilayer transformer satisfies the stability condition under mild assumptions, as indicated by the findings in \cite{li2023transformers}. This property is critical for maintaining robust generalization performance, as it implies that the model's predictions remain stable and reliable despite minor perturbations in the training dataset.

\begin{theorem}[Stability of Multilayer Transformer (\cite{li2023transformers})] 
The formal representation of the $n$-th query and $n-1$ in-context examples in the $n$-th prompt is ${\boldsymbol{x}}_{\text{prompt }}^{\left( n\right) }=( C_{(n-1)},{x}_{n})$, where \( C_{(n-1)}=  \left( s(x_1, y_1), . . . , s(x_{n-1}, y_{n-1})\right)\). Let \( C_{(n-1)} \) and \( C_{(n-1)'} \) be two prompts that differ only at the inputs \( (x_j, y_j)\) and \( (x_j^{\prime}, y_j^{\prime})\) where \( j < n \). Assume that inputs and labels are contained inside the unit Euclidean ball in \(\mathbb{R}^d\). These prompts are shaped into matrices \(\boldsymbol{X}_{\text{prompt}}\) and \(\boldsymbol{X}_{\text{prompt}}^{\prime} \in \mathbb{R}^{(2n-1) \times d}\), respectively. Consider the following definition of \( TF(\cdot) \) as a \( D \)-layer transformer: Starting with \(\boldsymbol{X}_{(0)} := \boldsymbol{X}_{\text{prompt}}\), the \( i \)-th layer applies MLPs and self-attention, with the softmax function being applied to each row:
\begin{align}
    \boldsymbol{X}_{(i)} = \text{ParallelMLPs}\left(\operatorname{ATTN}\left(\boldsymbol{X}_{(i-1)}\right)\right),\notag
\end{align}
where
\begin{align}
    \operatorname{ATTN}(\boldsymbol{X}) := \operatorname{softmax}\left(\boldsymbol{X} \boldsymbol{W}_i \boldsymbol{X}^{\top}\right) \boldsymbol{X} \boldsymbol{V}_i. \notag
\end{align}
Assume the transformer \( TF \) is normalized such that \(\|\boldsymbol{V}\| \leq 1\) and \(\|\boldsymbol{W}\| \leq \Gamma / 2\), and the MLPs adhere to \(\operatorname{MLP}(\boldsymbol{x}) = \operatorname{ReLU}(\boldsymbol{M} \boldsymbol{x})\) with \(\|\boldsymbol{M}\| \leq 1\). Let \( TF \) output the last token of the final layer \(\boldsymbol{X}_{(D)}\) corresponding to the query \( x_n \). Then,
\[
\left|TF\left({\boldsymbol{x}}_{\text{prompt }}^{\left( n\right) }\right) - TF\left({\boldsymbol{x}}_{\text{prompt }}^{\prime\left( n\right) }\right)\right| \leq \frac{2}{2n-1}\left((1+\Gamma) e^{\Gamma}\right)^D .
\]
Thus, assuming the loss function \(\ell(y, \cdot)\) is \( L \)-Lipschitz, the algorithm induced by \( TF(\cdot) \) exhibits error stability with \(\beta = \frac{2L\left((1+\Gamma) e^{\Gamma}\right)^D}{n}\).

\end{theorem}

Additionally, we need to provide further remarks regarding this theorem \cite{li2023transformers}. While the reliance on depth is growing exponentially, this constraint is less significant for most transformer architectures, as they typically do not have extreme depths. For instance, from 12 to 48 layers are present in various versions of GPT-2 and BERT \cite{devlin2018bert}. In this theorem, the upper bound on $\Gamma$ ensures that no single token can exert significant influence over another. The crucial technical aspect of this result is demonstrating the stability of the self-attention layer, which is the core component of a transformer. This stability is essential for maintaining the integrity and performance of the model, as it ensures that the influence of any individual token is limited, thereby preserving the overall structure and coherence of the model's output.

Assume the model is trained on $T$ tasks, each including a data sequence with $n$ examples. Each sequence is input to the model in an auto-regressive manner during training. By conceptualizing ICL as an algorithmic learning problem, \cite{li2023transformers} derive a multitask generalization rate of $1/\sqrt{nT}$ for i.i.d. data. To establish the appropriate reliance on sequence length (the $\sqrt{n}$ factor), they address temporal dependencies by correlating generalization with algorithmic stability. Their approach synthesizes the stability bound proof framework, as rigorously detailed in Theorem \ref{thm:stability}, with the covering number analysis presented in Theorem \ref{theorem_chaining}. By combining these complementary methodologies, they establish a solid theoretical foundation that addresses both stability and capacity, enabling a detailed analysis of the model's generalization behavior. Next, we introduce the excess risk bound, which quantifies the performance gap between the learned model and the optimal model. This measure accounts for both generalization error and optimization error, providing a comprehensive evaluation of the model's effectiveness.

\begin{theorem}[Excess Risk Bound in \cite{li2023transformers}] For all $T$ tasks, let's assume that the set of algorithmic hypotheses $\mathcal{A}$ is $\frac{K}{n}$-stable as defined in \ref{errorstability}, and that the loss function $\ell(y, \cdot)$ is $L$-Lipschitz continuous, taking values in the interval $[0,1]$. Allow $\mathcal{A}$ to stand for the empirical solution that was acquired. Afterwards, the following bound is satisfied by the excess multi-task learning (MTL) test risk with a probability of at least $1-2\delta$:
\begin{equation}
    \mathcal{R}(\mathcal{A})-\mathop{\min}_{\mathcal{A}}  \mathcal{R}(\mathcal{A})\leq \inf_{\epsilon > 0} \left\{ 4L\epsilon + 2(1 + K \log n) \sqrt{\frac{\log (\mathcal{N}(\mathcal{A}, \rho, \epsilon) / \delta)}{cnT}} \right\}. \notag\label{MTL excess}
\end{equation}
\end{theorem}
The term $\mathcal{N}(\mathcal{A}, \rho, \epsilon)$ denotes the $\epsilon$-covering number of the hypothesis set $\mathcal{A}$, utilized to regulate model complexity. In this context, $\rho$ represents the prediction disparity between the two algorithms on the most adverse data sequence, governed by the Lipschitz constant of the transformer design. The bound in \ref{MTL excess} attains a rate of $1/\sqrt{nT}$ by encompassing the algorithm space with a resolution of $\varepsilon$. For Lipschitz architectures possessing $\operatorname{dim}(\mathcal{A})$ trainable weights, it follows that $\log \mathcal{N}(\mathcal{A}, \rho, \varepsilon) \sim \operatorname{dim}(\mathcal{H}) \log(1/\varepsilon)$. Consequently, the excess risk is constrained by $\sqrt{\frac{\operatorname{dim}(\mathcal{A})}{nT}}$ up to logarithmic factors, and will diminish as $n$ and $T$ go toward infinity.

It is worth noting that \cite{li2023transformers} addresses the multi-layer multi-head self-attention model, whereas \cite{deoraoptimization} derives convergence and generalization guarantees for gradient-descent training of a single-layer multi-head self-attention model through stability analysis, given a suitable realizability condition on the data.

Further contributions have explored how ICL can be framed as a process of searching for the optimal algorithm during the training phase, while research on the inference phase has demonstrated that generalization error tends to decrease as the length of the prompt increases. Although these studies provide valuable insights into interpreting the performance of FMs through the lens of ICL’s generalization abilities, they often remain limited to simplified models and tasks. There is still a need for more in-depth research, particularly in understanding the impact of various components within complete transformer architectures and extending these analyses to more complex tasks. This deeper exploration could lead to a more comprehensive understanding of how foundation models achieve their impressive generalization performance across a wider range of scenarios.

\subsubsection{Interpreting Through Expressivity Analysis }

The utilization of prompts plays a crucial role in enhancing the expressive capabilities of FMs, particularly through techniques such as ICL prompting.

Recent research has provided significant insights into the ICL capabilities of transformers, establishing a link between gradient descent (GD) in classification and regression tasks and the feedforward operations within transformer layers \cite{dong2022survey}. This relationship has been rigorously formalized by \cite{akyurek2022learning}, \cite{von2023transformers}, and \cite{dai2022can}, who have demonstrated through constructive proofs that an attention layer possesses sufficient expressiveness to perform a single GD step. These studies aim to interpret ICL by proposing that, during the training of transformers on auto-regressive tasks, the forward pass effectively implements in-context learning via gradient-based optimization of an implicit auto-regressive inner loss, derived from the in-context data.

Specifically, research in \cite{von2023transformers} presents a construction where learning occurs by simultaneously updating all tokens, including the test token, through a linear self-attention layer. In this framework, the token produced in response to a query (test) token is transformed from its initial state ${W}_{0}{x}_{\text{test }}$, where ${W}_{0}$ is the initial weight matrix, to the post-learning prediction $\widehat{y} = \left( {{W}_{0} + {\Delta W}}\right) {x}_{\text{test }}$ after one gradient descent step. Let $P$, $V$, and $K$ be the projection, value, and key matrices, respectively. The theorem is as follows:

\begin{theorem}(Linking Gradient Descent and Self-Attention \cite{von2023transformers})
Given a single-head linear attention layer and a set of in-context example ${e}_{j} = \left( {{x}_{j},{y}_{j}}\right)$ for $j = 1,\ldots,N$, it is possible to construct key, query, and value matrices ${W}_{K},{W}_{Q},{W}_{V}$, as well as a projection matrix $P$, such that a transformer step applied to each in-context example ${e}_{j}$ is equivalent to the gradient-induced update 
$$
{e}_{j} \leftarrow  \left( {{x}_{j},{y}_{j}}\right)  + \left( {0, - {\Delta W}{x}_{j}}\right)  = \left( {{x}_{j},{y}_{j}}\right)  +{PV}{K}^{T}{q}_{j},
$$
resulting in ${e}_{j} = \left( {{x}_{j},{y}_{j} - \Delta {y}_{j}}\right)$. This dynamic is similarly applicable to the query token $\left( {{x}_{N + 1},{y}_{N + 1}}\right)$ .
\end{theorem}

Drawing from these observations, \cite{giannou2023looped} have shown that transformers are capable of approximating a wide range of computational tasks, from programmable computers to general machine learning algorithms. Additionally, \cite{li2023closeness} explored the parallels between single-layer self-attention mechanisms and GD in the realm of softmax regression, identifying conditions under which these similarities hold.

\cite{ahn2024transformers} and \cite{zhang2024trained} conducted further investigation into the implementation of ICL for linear regression, specifically through the utilization of single-layer linear attention models. They demonstrated that a global minimizer of the population ICL loss can be interpreted as a single gradient descent step with a matrix step size. \cite{mahankali2023one} provided a similar result specifically for ICL in isotropic linear regression. It is noteworthy that \cite{zhang2024trained} also addressed the optimization of attention models; however, their findings necessitate an infinite number of pretraining tasks. Similarly, \cite{bai2024transformers} established task complexity bounds for pretraining based on uniform convergence and demonstrated that pretrained transformers are capable of performing algorithm selection. Furthermore, a simplified linear parameterization was utilized in \cite{wu2023many}, which provided more refined task complexity boundaries for the purpose of pretraining a single-layer linear attention model.

Notably, exploring the connection between gradient descent and ICL is highly valuable for advancing the theoretical understanding of ICL. This is because it allows us to leverage the extensive tools from traditional learning theory, such as the convergence proof frameworks for gradient descent and stochastic gradient descent, as outlined in Theorems \ref{thm: convergence GD} and \ref{thm: convergence of SGD}, respectively.

Furthermore, there remain areas in existing research that require refinement. For example, many current studies attempt to model each gradient descent step using a single attention layer within transformer architectures. This approach raises questions regarding its suitability, particularly given that transformers are typically not extremely deep. For instance, variants of models like GPT-2 and BERT usually consist of 12 to 48 layers, which constrains the number of gradient descent steps that can be effectively modeled. This limitation suggests that further investigation is needed to refine these models and better align them with the theoretical underpinnings of gradient descent in the context of ICL.

\subsubsection{Concluding Remarks}

In this section, we delve into the theoretical underpinnings of FMs to address the challenges users encounter when employing ICL prompts. By understanding the inner workings of FMs, we aim to provide theoretical explanations and insights that can guide practitioners and inspire future research directions.

\emph{Theoretical Interpretations of Prompt Sensitivity:} One key challenge users encounter is understanding why seemingly similar prompts can lead to vastly different outputs, as shown in Figure \ref{fig_ICL_1}. This issue can be interpreted through the stability-based generalization bounds established by \cite{li2023transformers}. From a theoretical perspective, \cite{li2023transformers} quantified the influence of one token on another and demonstrated that the stability of ICL predictions improves with an increased number of in-context examples, an appropriate (non-extreme) depth of Transformer layers, and training on complete sequences. These factors help enhance the model's resistance to input perturbations. Furthermore, exploring the impact of introducing noisy data during training on ICL stability could be an interesting direction for future research, offering additional ways to reinforce model stability.

\emph{Constructing Effective In-Context Examples:} Another major challenge for users is how to construct appropriate in-context examples as prompts, as shown in Figure \ref{fig_ICL2}. Numerous theoretical studies have provided insights into this challenge.  \cite{li2023transformers,li2024nonlinear} suggests that longer prompts generally lead to lower generalization errors and improved performance because they offer richer contextual information that aids the model in better understanding the task. Additionally, \cite{wu2023many} found that model performance improves when the context length during inference closely matches that of pretraining. Furthermore, when in-context examples share similar patterns or structures with the query, the model's performance can be significantly enhanced. Specifically, the amount of influence that various parameters have on the performance of ICL generalization was quantified by \cite{li2024nonlinear}. These factors include the quantity of relevant features and the fraction of context instances that include the same relevant pattern as the new query.

\subsection{Chain-of-Thought Reasoning}

\begin{figure}[htbp]
    \centering
\includegraphics[width=\textwidth]{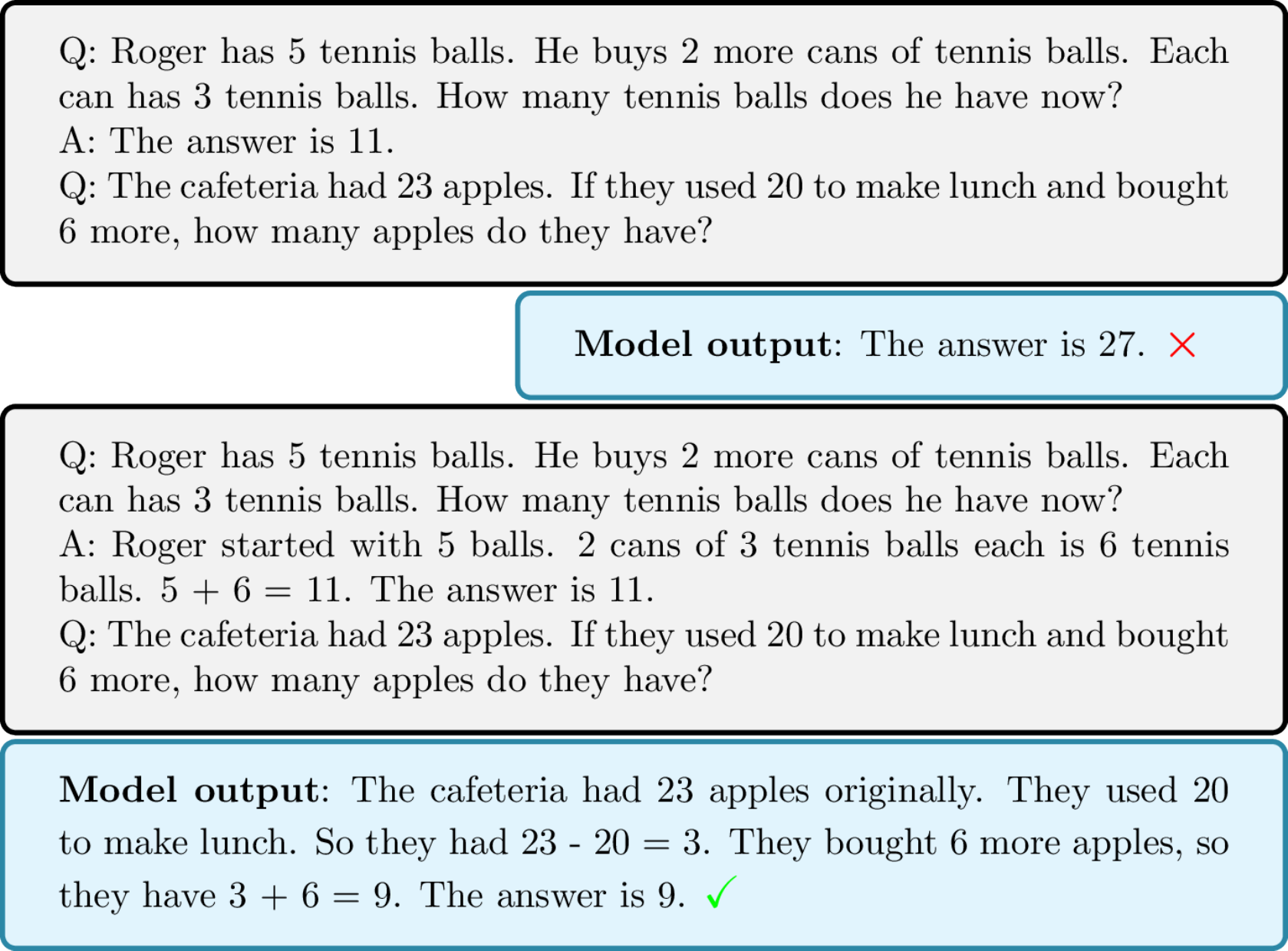}
\caption{Impact of CoT Prompting on Solving Complex Arithmetic Problems \cite{wei2022chain}}\label{figure_cot2}
\end{figure}

When users employ FMs for reasoning tasks involving CoT prompting, several challenges and uncertainties arise. One important question is whether CoT can effectively improve model performance in complex reasoning tasks. While CoT prompting has been shown to enable FMs to handle intricate arithmetic problems, as illustrated in Figure \ref{figure_cot2}, there is still uncertainty about the generalizability of this effect. practitioners are particularly interested in understanding why CoT improves performance. This raises questions about the underlying mechanisms that allow CoT to facilitate more effective reasoning in FMs.

\begin{figure}[htbp]
\includegraphics[width=\textwidth]{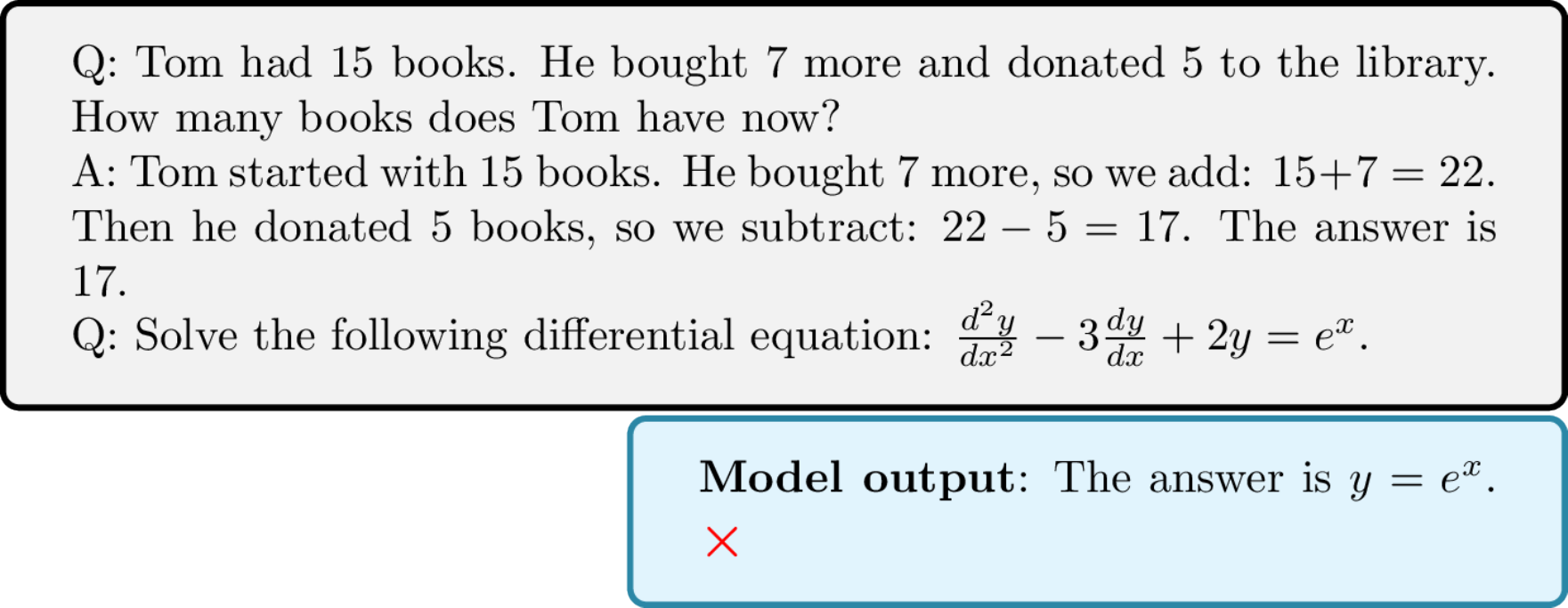}
\caption{Chain-of-Thought Failure Due to Sample-Query Mismatch.}\label{figure_COT1}
\end{figure}

Another challenge involves determining the optimal prompt structure to leverage CoT for accurate results. Users are often uncertain about how many steps or how detailed the prompts need to be for the model to fully utilize CoT and produce correct inferences. This uncertainty makes it difficult for users to craft effective prompts that guide the model towards the desired outcome.

Additionally, users face challenges when they cannot provide precise reasoning samples. In such scenarios, they wonder whether CoT remains effective. This uncertainty about the robustness of CoT, particularly when ideal samples are not available, poses a significant challenge for users attempting to harness the full potential of FMs for complex reasoning tasks. As illustrated in the Figure \ref{figure_COT1}, when the reasoning samples involve only simple arithmetic, but the query requires solving a complex differential equation, CoT reasoning in FMs fails.

\subsubsection{Interpreting Through Generalization Analysis}
The CoT method \cite{wei2022chain} enhances a model's reasoning capabilities by guiding it through a structured, step-by-step problem-solving process. This approach is particularly effective for tasks that require multi-step reasoning, such as logical puzzles and complex question-answering scenarios. By prompting the model to generate intermediate steps, CoT facilitates more accurate conclusions through a sequence of structured thought processes.

Numerous studies \cite{li2024dissecting,feng2024towards,lichain} investigate the training of Transformer models to develop CoT capabilities for evaluating novel data and tasks. In this context, we articulate
the supervised learning framework proposed in \cite{li2024nonlinear}, which utilizes pairs of prompts and labels. For each prompt-output pair associated with a task $f=f_K\circ\cdots\circ f_2\circ f_1$, we construct a prompt $\mathbf{P}$ that integrates the query input $z_{k-1}$ by incorporating $n_{tr}$ reasoning examples alongside the initial $k-1$ steps of the reasoning query. The formulation of the prompt $\mathbf{P}$ for the query input $z_{k-1}$ is expressed as follows:
$$
\mathbf{P} = \left( {{\mathbf{E}}_{1},{\mathbf{E}}_{2},\cdots ,{\mathbf{E}}_{{l}_{tr}},{\mathbf{Q}}_{k}}\right)  \in  {\mathbb{R}}^{2{d} \times  \left( {{n}_{tr}K + k}\right) },
$$
and
$$
{\mathbf{E}}_{i} = \left( \begin{matrix} {\mathbf{x}}_{i} & {\mathbf{y}}_{i,1} & \cdots & {\mathbf{y}}_{i,K - 1} \\  {\mathbf{y}}_{i,1} & {\mathbf{y}}_{i,2} & \cdots & {\mathbf{y}}_{i,K} \end{matrix}\right) ,{\mathbf{Q}}_{k} = \left( \begin{matrix} {\mathbf{z}}_{0} & {\mathbf{z}}_{1} & \cdots & {\mathbf{z}}_{k - 2} & {\mathbf{z}}_{k - 1} \\  {\mathbf{z}}_{1} & {\mathbf{z}}_{2} & \cdots & {\mathbf{z}}_{k - 1} & \mathbf{0} \end{matrix}\right),
$$
where ${\mathbf{E}}_{i}$ is the $i$-th context example, and ${\mathbf{Q}}_{k}$ is the first $k$ steps of the reasoning query for any $k$ in $\left\lbrack  K\right\rbrack$ . We have ${\mathbf{y}}_{i,k} = {f}_{k}\left( {\mathbf{y}}_{i,k - 1}\right)$ and ${\mathbf{z}}_{k} = {f}_{k}\left( {\mathbf{z}}_{k - 1}\right)$ for $i \in  \left\lbrack  {n}_{tr}\right\rbrack  ,k \in  \left\lbrack  K\right\rbrack$ with a notation ${\mathbf{y}}_{i,0} \mathrel{\text{:=}} {\mathbf{x}}_{i}$.

From this setting, it can be observed that CoT emphasizes demonstrating intermediate reasoning steps in the problem-solving process, while referring to the definition from the section \ref{section_ICL}, ICL focuses on learning task mappings directly by leveraging several input-output pairs provided in the context. In certain scenarios, ICL can be regarded as a simplified reasoning process without intermediate steps, whereas CoT represents an enhanced version with multi-step reasoning \cite{li2024nonlinear}.


To interpret the theoretical underpinnings of how transformers can be trained to acquire CoT capabilities, \cite{li2024nonlinear} offers a comprehensive theoretical analysis focused on transformers with nonlinear attention mechanisms. Their study delves into the mechanisms by which these models can generalize CoT reasoning to previously unseen tasks, especially when the model is exposed to examples of the new task during input. Specifically, they consider a learning model that is a single-head, one-layer attention-only transformer. The training problem is framed to enhance reasoning capabilities, addressing the empirical risk minimization defined as
$$
\min _{\theta} R_n(\theta):=\frac{1}{n} \sum_{i=1}^n \ell\left(\mathbf{P}^i, \mathbf{z}^i; \boldsymbol{\theta}\right),
$$
where $n$ represents the number of prompt-label pairs $\left\{\mathbf{P}^i, \mathbf{z}^i\right\}_{i=1}^n$ and $\boldsymbol{\theta}$ denotes model weights. The loss function utilized is the squared loss, specifically given by 
$$
\ell\left(\mathbf{P}^i, \mathbf{z}^i; \boldsymbol{\theta}\right)=\frac{1}{2}\left\|\mathbf{z}^i-F\left(\mathbf{P}^i; \boldsymbol{\theta}\right)\right\|^2,
$$
where $F\left(\mathbf{P}^i; \boldsymbol{\theta}\right)$ denotes the output of the transformer model.
Furthermore, following the works  \cite{li2024dissecting,feng2024towards,lichain,li2024nonlinear}, the CoT generalization error for a testing query $\mathbf{x}_{\text {query }}$, the testing data distribution $\mathcal{D}^{\prime}$, and the labels $\left\{\mathbf{z}_k\right\}_{k=1}^K$ on a $K$-step testing task $f \in \mathcal{T}^{\prime}$ is defined as
$$
R_{C o T, \mathbf{x}_{\text {query }} \sim \mathcal{D}^{\prime}, f \in \mathcal{T}^{\prime}}^f(\boldsymbol{\theta})=\mathbb{E}_{\mathbf{x}_{\text {query }} \sim \mathcal{D}^{\prime}}\left[\frac{1}{K} \sum_{k=1}^K 1\left[\mathbf{z}_k \neq \mathbf{v}_k\right]\right].
$$
where $\mathbf{v}_1, \ldots, \mathbf{v}_K$ are the model's final outputs presented as the CoT results for the query. They further quantify the requisite number of training samples and iterations needed to instill CoT capabilities in a model, shedding light on the computational demands of achieving such generalization. Specifically, they describe the model's convergence and testing performance during the training phase using sample complexity analysis, as detailed in the following theorem:

\begin{theorem}(Training Sample Complexity and Generalization Guarantees for CoT \cite{li2024nonlinear}) Consider $M$ training-relevant (TRR) patterns ${\mathbf{\mu }}_{1},{\mathbf{\mu }}_{2},\cdots ,{\mathbf{\mu }}_{M}$, which form an orthonormal set $\mathcal{M} = {\left\{  {\mathbf{\mu }}_{i}\right\}  }_{i = 1}^{M}$. The fraction of context examples whose inputs share the same TRR patterns as the query is denoted by $\alpha$. For any $\epsilon  > 0$, when (i) the number of context examples in every training sample is
$$
{n} \geq  \Omega \left( {\alpha }^{-1}\right),  
$$
(ii) the number of SGD iterations satisfies
$$
T \geq  \Omega \left( {{\eta }^{-1}{\alpha }^{-1}{K}^{3}\log \frac{K}{\epsilon } + {\eta }^{-1}{MK}\left( {{\alpha }^{-1} + {\epsilon }^{-1}}\right) }\right) , 
$$
and (iii) the training tasks and samples are selected such that every TRR pattern is equally likely in each training batch with batch size $B \geq  \Omega \left( {\max \left\{  {{\epsilon }^{-2},M}\right\}   \cdot  \log M}\right)$, the step size $\eta  < 1$ and $N = {BT}$ samples, then with a high probability, the returned model guarantees
\begin{align}
{\mathbb{E}}_{{\mathbf{x}}_{\text{query }} \in  \mathcal{M}}\left\lbrack  {\ell \left( { \mathbf{P},\mathbf{z}; \boldsymbol{\theta}}\right) }\right\rbrack   \leq  \mathcal{O}\left( \epsilon \right), \label{cot_train_gene}
\end{align}
where $\theta$ denotes the set of all model weights. Furthermore, as long as (iv) testing-relevant (TSR) patterns ${\mathbf{\mu }}_{1}',{\mathbf{\mu }}_{2}',\cdots ,{\mathbf{\mu }}_{M'}'$ satisfy
$$
\mu_j^{\prime} \in \operatorname{span}\left(\mu_1, \mu_2, \cdots, \mu_M\right),
$$
for $j \in\left[M^{\prime}\right]$, and (v) the number of testing examples for every test task $f \in \mathcal{T}^{\prime}$ is
\begin{align}
n_{t s}^f \geq \Omega\left(\left(\alpha^{\prime} \tau^f \rho^f\right)^{-2} \log M\right).\label{cot_genrali}
\end{align}
where $\alpha^{\prime}$ represents the fraction of context examples whose inputs exhibit the same TSR
patterns as the query. The term $\rho^{f}$ denotes the primacy of the step-wise transition matrices,
while $\tau^{f}$ refers to the min-max trajectory transition probability. Finally, we have 
\begin{align}
R_{C o T, \mathbf{x}_{\text {query }} \sim \mathcal{D}^{\prime}, f \in \mathcal{T}^{\prime}}^f(\boldsymbol{\theta})=0. 
\end{align}
\end{theorem}


This theorem elucidates the necessary conditions and properties that facilitate CoT generalization. It establishes that, to train a model capable of guaranteed CoT performance, the number of context examples per training sample, the total number of training samples, and the number of training iterations must all exhibit a linear dependence on ${\alpha }^{-1}$. This finding supports the intuition that CoT effectiveness is enhanced when a greater number of context examples closely align with the query. Moreover, the attention mechanisms in the trained model are observed to concentrate on testing context examples that display similar input TSR patterns relative to the testing query during each reasoning step. This focus is a pivotal characteristic that underpins the efficacy of CoT generalization.

Additionally, to achieve a zero CoT error rate on test tasks utilizing the learned model, Equation \ref{cot_genrali} indicates that the requisite number of context examples associated with task $f$ in the testing prompt is inversely proportional to $(\alpha^{\prime}\tau^{f}\rho^{f})^{-2}.$ The minimum probability of the most probable $K$-step reasoning trajectory over
the initial TSR pattern is quantified by $\tau^{f}$, whereas $\alpha^\prime$ denotes the fraction of context examples whose inputs share the same TSR patterns as the query in this context. A larger constant $\rho^{f}$ indicates a higher reasoning accuracy at each step. This result formally encapsulates the notion that both the quantity of accurate context examples and their similarity to the query significantly enhance CoT accuracy.

Furthermore, \cite{li2024dissecting} demonstrate, the success of CoT can be attributed to its ability to decompose the ICL task into two phases: data filtering and single-step composition. This approach significantly reduces the sample complexity and facilitates the learning of complex functions.


\subsubsection{Interpreting Through Expressivity Analysis}

CoT prompting further amplifies expressive capabilities by guiding the model through a structured reasoning process. Following this observation, research has shifted towards comparing the task complexity between reasoning with CoT and that without CoT \cite{li2024dissecting,feng2024towards,lichain,merrill2023expressive}. Under the ICL framework, \cite{li2024dissecting} demonstrates the existence of a transformer capable of learning a multi-layer perceptron, interpreting CoT as a process of first filtering important tokens and then making predictions through ICL. They establish the required number of context examples for achieving the desired prediction with the constructed Transformer. 
Additionally, studies \cite{feng2024towards,lichain,merrill2023expressive} reveal that transformers utilizing CoT are more expressive than those without CoT. 

In detail, \cite{feng2024towards} consider tasks on evaluating arithmetic expressions and solving linear equations. They demonstrate that the complexity of the problems is lower bounded by $NC^1$ (Nick's Class 1), indicating that no log-precision autoregressive Transformer with limited depth and $poly(n)$ hidden dimension could solve the non-CoT arithmetic problem or non-CoT equation problem of size $n$ if $NC^1 \neq TC^0$ (Threshold Circuits of Constant Depth 0). Conversely, they construct autoregressive Transformers with finite depth and $poly(n)$ hidden dimension that solve these two problems in CoT reasoning. 
In the context of iterative rounding, \cite{lichain} expands the analysis to constant-bit precision and establishes $AC^0$ (Alternating Circuit of Depth 0) as a more stringent upper bound on the expressiveness of constant-depth transformers with constant-bit precision. 
On the other hand, \cite{merrill2023expressive} considers the impact of the chain length in the CoT process. They prove that logarithmic intermediate steps enhance the upper bound of computational power to $L$ (log-space), increasing the intermediate steps to linear size enhances the upper bound within $NC^1$, and polynomial sizes have upper bound $P$ (Polynomial Time). 

By breaking down complex tasks into intermediate steps, CoT allows the model to articulate its thought process clearly and logically. This structured prompting not only improves the accuracy of the model's responses but also enables it to handle intricate, multi-step problems with greater finesse.


\subsubsection{Concluding Remarks}
In this section, we delve into the theoretical underpinnings of FMs to address the challenges users encounter when employing CoT prompts. We aim to provide theoretical explanations and insights that can guide practitioners and inspire future research directions about CoT.

\emph{CoT-driven  Expressive Capabilities Enhancement:} A body of research has demonstrated that CoT significantly enhances the expressive capabilities of FMs. For example, \cite{feng2024towards} examined the expressiveness of FMs utilizing CoT in addressing fundamental mathematical and decision-making challenges. The researchers initially presented impossibility results, demonstrating that bounded-depth transformers cannot yield accurate solutions for fundamental arithmetic tasks unless the model size increases super-polynomially with input length. Conversely, they developed an argument demonstrating that constant-sized autoregressive transformers can successfully satisfy both requirements by producing CoT derivations in a standard mathematical language format. This effectively demonstrates that CoT improves the reasoning capabilities of FMs, particularly in mathematical tasks. Additionally, other studies, such as \cite{merrill2023expressive} and \cite{lichain}, have shown that CoT-based intermediate generation extends the computational power of FMs in a fundamental way. These findings reinforce the idea that CoT prompting enhances model performance by allowing FMs to break down complex reasoning processes into manageable steps, thereby enabling more accurate outcomes.

\emph{CoT Prompt Construction:}  How can we construct prompts that effectively leverage CoT to achieve accurate results?  \cite{li2024nonlinear} demonstrated that, to guarantee CoT functionality, the number of context examples per training sample, the total number of training samples, and the number of training iterations must all scale linearly with the inverse of the proportion of context examples whose inputs share the same TRR patterns as the query. This aligns with the intuitive understanding that CoT performance improves when more context examples are similar to the query. When users are unable to provide precise reasoning samples, they may question the effectiveness of CoT, as shown in Figure \ref{figure_COT1}. \cite{li2024nonlinear} addressed this concern by demonstrating that, when reasoning examples contain noise and inaccuracies, achieving zero CoT error on certain tasks requires a number of context examples that is proportional to the inverse square of the product of several factors. These factors include the fraction of accurate context examples and the similarity between the context examples and the query. This result formalizes the intuition that more accurate and relevant context examples, even in the presence of noise, significantly enhance CoT accuracy.


\subsection{Adaptation to Distribution Shift}

From a user’s perspective, FMs face significant challenges when dealing with distribution shifts. A common issue arises when users are unable to provide prompts that accurately align with the query. FMs often depend on prompts that mirror the data they were trained on, so when prompts deviate from this, the model’s performance can suffer, leading to inaccurate or irrelevant outputs. Below, we outline the primary challenges users encounter.

\begin{figure}[htbp]
    \centering




\includegraphics[width=\textwidth]{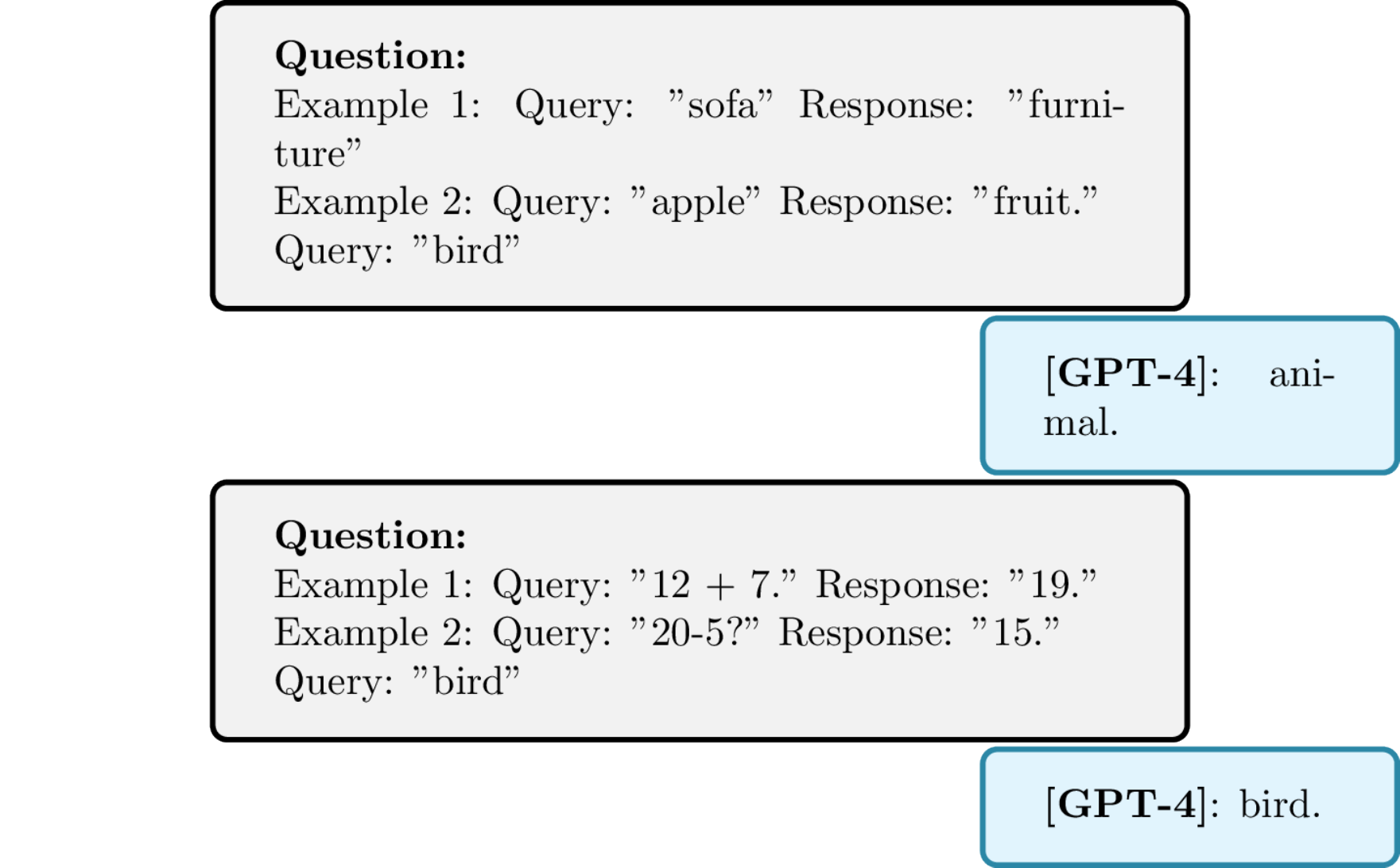}
\caption{Example of In-Context Example and Query Mismatch Impacting Model Performance}\label{figure_DIS1}
\end{figure}

\textbf{Handling Inaccurate Prompts.} A frequent issue arises when users are unable to provide prompts that accurately align with the query. FMs often depend on prompts that reflect the structure and data they were trained on. When prompts deviate from these familiar patterns, the model’s performance can suffer, leading to inaccurate or irrelevant outputs. As shown in Figure \ref{figure_DIS1}, in an extreme case where the in-context example involves arithmetic, but the query is about birds, the foundation model struggles to provide a reasonable response when there is a complete mismatch between the in-context example and the query.

\begin{figure}[htbp]
    \centering


\includegraphics[width=\textwidth]{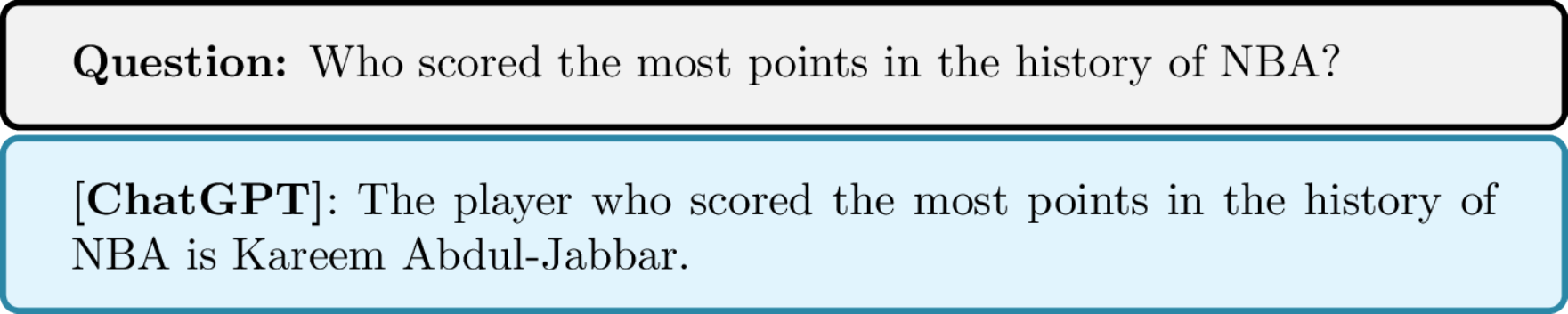}
\caption{FM Providing Outdated Information \cite{liu2023trustworthy}.}\label{figure_dis2}
\end{figure}

\textbf{Limitations in Providing Real-Time or Updated Information.} Another key challenge arises when users require the most current information, which FMs are often unable to provide. Since these models are typically trained on static datasets that are periodically updated, they often lack access to the most recent data or real-time information. For instance, as illustrated in Figure \ref{figure_dis2}, questions such as "Who scored the most points in NBA history?" or "Who is the richest person in the world?" often require up-to-date answers. However, due to the lag in data updates, the model might provide outdated or incorrect information, failing to meet the user’s need for timely responses. This limitation highlights the difficulty users face when relying on foundational models for information that evolves over time.

\textbf{Struggles with Uncommon Tasks.} FMs also face difficulties when confronted with uncommon or novel tasks. Users may question whether the model can handle tasks it wasn’t explicitly trained for. When asked to solve rare or unique problems, the model’s performance may degrade, raising concerns about its ability to generalize to new or unfamiliar domains.

Next, we will explore how generalization analysis can be used to interpret FM' performance under different distribution shifts, addressing the challenges and concerns users face in these scenarios.

\subsubsection{Interpreting Through Generalization Analysis}
Theoretical analysis of FMs under distribution shifts across various prompts can help interpret their remarkable out-of-distribution generalization capabilities. Although deep learning theory typically assumes that training and test distributions are identical, this assumption often does not hold in real-world scenarios, leading to models that may struggle to generalize effectively \cite{zhou2022domain}. Due to the remarkable ability of attention-based neural networks, such as transformers, to exhibit ICL, numerous studies have theoretically investigated the behavior of trained transformers under various distribution shifts \cite{zhang2024trained,lintransformers,li2023transformers,li2024nonlinear,wies2024learnability,huangcontext}. These studies delve into various distribution shifts that were initially examined experimentally for transformer models by \cite{garg2022can}.

Building on the methodologies outlined in \cite{garg2022can,zhang2024trained}, we focus on the design of training prompts structured as 
$$
\left( {{x}_{1},h\left( {x}_{1}\right) ,\ldots ,{x}_{N},h\left( {x}_{N}\right) ,{x}_{\text{query }}}\right).
$$
In this framework, we assume that ${x}_{i},{x}_{\text{query }}\overset{\text{ i.i.d. }}{ \sim }{\mathcal{D}}_{x}^{\text{train }}$ and $h \sim {\mathcal{D}}_{\mathcal{H}}^{\text{train }}$. For test prompts, we assume ${x}_{i}\overset{\text{ i.i.d. }}{ \sim }{\mathcal{D}}_{x}^{\text{test }}$, ${x}_{\text{query }} \sim {\mathcal{D}}_{\text{query }}^{\text{test }}$, and $h \sim {\mathcal{D}}_{\mathcal{H}}^{\text{test }}$. 
Next, we will explore the generalization ability of transformers under the ICL setting from three perspectives: task shifts \cite{li2023transformers,zhang2024trained,wies2024learnability,huangcontext}, query shifts \cite{zhang2024trained}, and covariate shifts \cite{zhang2024trained, li2024nonlinear}.

~\\
\textbf{Task Shifts:} Numerous studies have attempted to investigate the impact of task shift on generalization theoretically. \cite{zhang2024trained} delve into the dynamics of ICL in transformers trained via gradient flow on linear regression problems, with a single linear self-attention layer. They demonstrate that task shifts are effectively tolerated by the trained transformer. Their primary theoretical results indicate that the trained transformer competes favorably with the best linear model, provided the prompt length during training and testing is sufficiently large. Consequently, even under task shift conditions, the trained transformer generates predictions with errors comparable to the best linear model fitting the test prompt. This behavior aligns with observations of trained transformers by \cite{garg2022can}.

Additionally, \cite{wies2024learnability} adapt the PAC-based framework to analyze in-context learnability and leverage it to derive finite sample complexity results. They utilize their framework to demonstrate that, under modest assumptions, when the pretraining distribution consists of a combination of latent tasks, these tasks can be effectively acquired by ICL. This holds even when the input significantly diverges from the pretraining distribution. In their demonstration, they show that the ratio of the prompt probabilities, which is dictated by the ground-truth components, converges to zero at an exponential rate. This rate depends on the quantity of in-context examples as well as the minimum Kullback–Leibler divergence between the ground-truth component and other mixture components, indicating the distribution drift.

In contrast to the aforementioned studies focusing on task shift, \cite{li2023transformers} and \cite{huangcontext} explore transfer learning to evaluate the performance of ICL on new tasks. Specifically, they investigate the testing performance on unseen tasks that do not appear in the training samples but belong to the same distribution as the training tasks. Specifically, \cite{li2023transformers} theoretically establish that the transfer risk decays as $1/poly(T)$, where $T$ is the number of multi-task learning tasks. This is due to the fact that extra samples or sequences are usually not able to compensate for the distribution shift caused by the unseen tasks.

~\\
\textbf{Query Shifts:} 
In addressing the issue of query shift, wherein the distribution of queries and prompts in the test set is inconsistent, \cite{zhang2024trained} demonstrate that broad variations in the query distribution can be effectively managed. However, markedly different outcomes are likely when the test sequence length is limited and the query input is dependent on the training dataset. To be more specific, the prediction will be zero if the query input is orthogonal to the subspace that is spanned by the $x_i$'s. This phenomenon was prominently noticed in transformer models, as recorded by \cite{garg2022can}. This suggests that while certain architectures can generalize across diverse distributions, their performance can be severely constrained by the alignment of the query example with the training data's subspace, emphasizing the criticality of sequence length and data dependency in model evaluation.

~\\
\textbf{Covariate Shifts:} 
To address the issue of covariate shift, where the training and testing data distributions do not match, \cite{zhang2024trained} studied out-of-domain generalization within the context of linear regression problems with Gaussian inputs. They demonstrated that while gradient flow effectively identifies a global minimum in this scenario, the trained transformer remains vulnerable to even mild covariate shifts. Their theoretical findings indicate that, unlike task and query shifts, covariate shifts cannot be fully accommodated by the transformer. This limitation of the trained transformer with linear self-attention was also noted in the transformer architectures analyzed by \cite{garg2022can}. These observations suggest that despite the similarities in predictions between the transformer and ordinary least squares in some contexts, the underlying algorithms differ fundamentally, as ordinary least squares is robust to feature scaling by a constant.

Conversely, \cite{li2024nonlinear} explores classification problems under a data model distinct from that of \cite{zhang2024trained}. They offer a guarantee for out-of-domain generalization concerning a particular distribution type. They provide a sample complexity bound for a data model that includes both relevant patterns, which dictate the labels, and irrelevant patterns, which do not influence the labels. They demonstrate that advantageous generalization conditions encompass: (1) out-of-domain-relevant (ODR) patterns as linear combinations of in-domain-relevant (IDR) patterns, with the sum of coefficients equal to or exceeding one, and each out-of-domain-irrelevant (ODI) pattern situated within the subspace formed by in-domain-irrelevant (IDI) patterns; and (2) the testing prompt being adequately lengthy to include context inputs related to ODR patterns. Under these conditions, the transformer could effectively generalize within context, even amongst distribution shifts between the training and testing datasets.


\subsubsection{Concluding Remarks}
In this section, we explore theoretical frameworks that help interpret FMs and address challenges posed by distribution shifts. These challenges often occur when FMs struggle to handle prompts that deviate from the training distribution. 

\emph{The prompt-query alignment:} One critical challenge arises when users are unable to provide prompts that precisely align with the query, as illustrated in Figure \ref{figure_DIS1}, raising the question of how FMs handle query shifts—situations where the distribution of queries differs from that of prompts in the test set. Theoretical studies \cite{zhang2024trained} suggest that if the test sequence length is limited and query examples are dependent on the training data, FMs can still generalize effectively. However, when the query example is orthogonal to the subspace spanned by the training data, the model's prediction accuracy drops significantly. This highlights the importance of prompt-query alignment for optimal model performance.

\emph{The  diversity of tasks:} Another significant challenge lies in task shift, where users present uncommon tasks that FMs may not have been explicitly trained on. This scenario corresponds to the theoretical task shift problem. Research \cite{li2023transformers,zhang2024trained} shows that if sufficiently long prompts are provided during both training and testing, FMs can generalize well even under task shift conditions. Enhancing the diversity of tasks during the pretraining phase further strengthens the model's ability to handle task shifts. This is because greater task diversity during training exposes the model to a broader range of problem types, thereby enhancing its adaptability to new tasks.

While FMs show strong generalization under specific conditions,   FMs often struggle to provide real-time or updated information due to outdated training data, as illustrated in Figure \ref{figure_dis2}. Addressing distribution shifts will require developing strategies for detecting shifts and updating models efficiently. For example, policy compliance issues, such as handling evolving terms that may become offensive over time, can benefit from such advancements.

\subsubsection{Future Directions for Theoretical Foundations in Prompt Engineering for FMs}
In the preceding section, we highlighted the pivotal role of prompt engineering in expanding the reasoning capabilities of FMs. Foundational advancements, such as ICL and CoT prompting, allow FMs to engage in complex reasoning by guiding them through structured, sequential logic.

Building on these core techniques, an array of prompt engineering strategies has emerged, extending FMs' capacity to handle increasingly intricate tasks \cite{sahoo2024systematic}. For example, CoT prompting, introduced by \cite{wei2022chain}, enables FMs to decompose multi-step problems into logical sub-steps, effectively simulating human problem-solving approaches. More recent methods, such as Automatic Chain of Thought (Auto-CoT) prompting \cite{zhangautomatic}, automate the creation of reasoning chains, enabling more streamlined and systematic prompt generation through structured cues like “Let’s think step-by-step,” which supports systematic reasoning in complex problem-solving scenarios.

Further innovations include self-consistency, which enhances inference reliability by generating diverse reasoning paths and selecting the most coherent answer, a technique particularly suited to tasks where multiple solution paths are valid. Expanding on this, frameworks such as Tree of Thoughts (ToT) \cite{yao2024tree} and Graph of Thoughts (GoT) \cite{yao2023beyond} introduce tree-based and graph-based reasoning structures, allowing FMs to explore complex problem spaces with flexibility, backtracking, and exploration—abilities essential for addressing highly complex tasks.

Despite these practical advancements, a comprehensive theoretical foundation explaining the inner workings of prompt engineering remains an open frontier. Future research should prioritize developing formal frameworks that elucidate the mechanisms by which prompt engineering influences FM behavior and inference. Initial theoretical studies, such as those examining retrieval-augmented generation (RAG) \cite{xu2023search,shi2023replug,asai2023self,ram2023context}, have begun to quantify the trade-offs between the benefits of external information and the risks introduced by noise. By modeling RAG as a fusion of the FM’s internal knowledge distribution with the distribution of retrieved texts, these studies  offer a theoretical framework for understanding the balance between beneficial knowledge and the risk of misinformation, based on distributional differences in token prediction. While promising, these findings represent only an initial step. Comprehensive theoretical frameworks that can explain a broader range of prompting strategies are still urgently needed.

Establishing deeper theoretical insights into prompt engineering would not only clarify how prompts shape FM reasoning processes but also support the development of more interpretable, reliable FMs for complex real-world applications. Such insights would enable us to harness the full potential of FMs with greater predictability and control, paving the way for robust applications across diverse domains.

\section{Interpreting Training Process in Foundation Models}\label{sec4}
In this section, we aim to leverage the interpretable methods discussed in Section \ref{sec2}  to interpret the  training dynamics of FMs, including emergent capabilities, parameter evolution during pretraining and  reinforcement learning from human feedback.
\subsection{Emergent Capabilities}

\begin{figure}[htbp]
  \centering
  \includegraphics[width=1\textwidth]{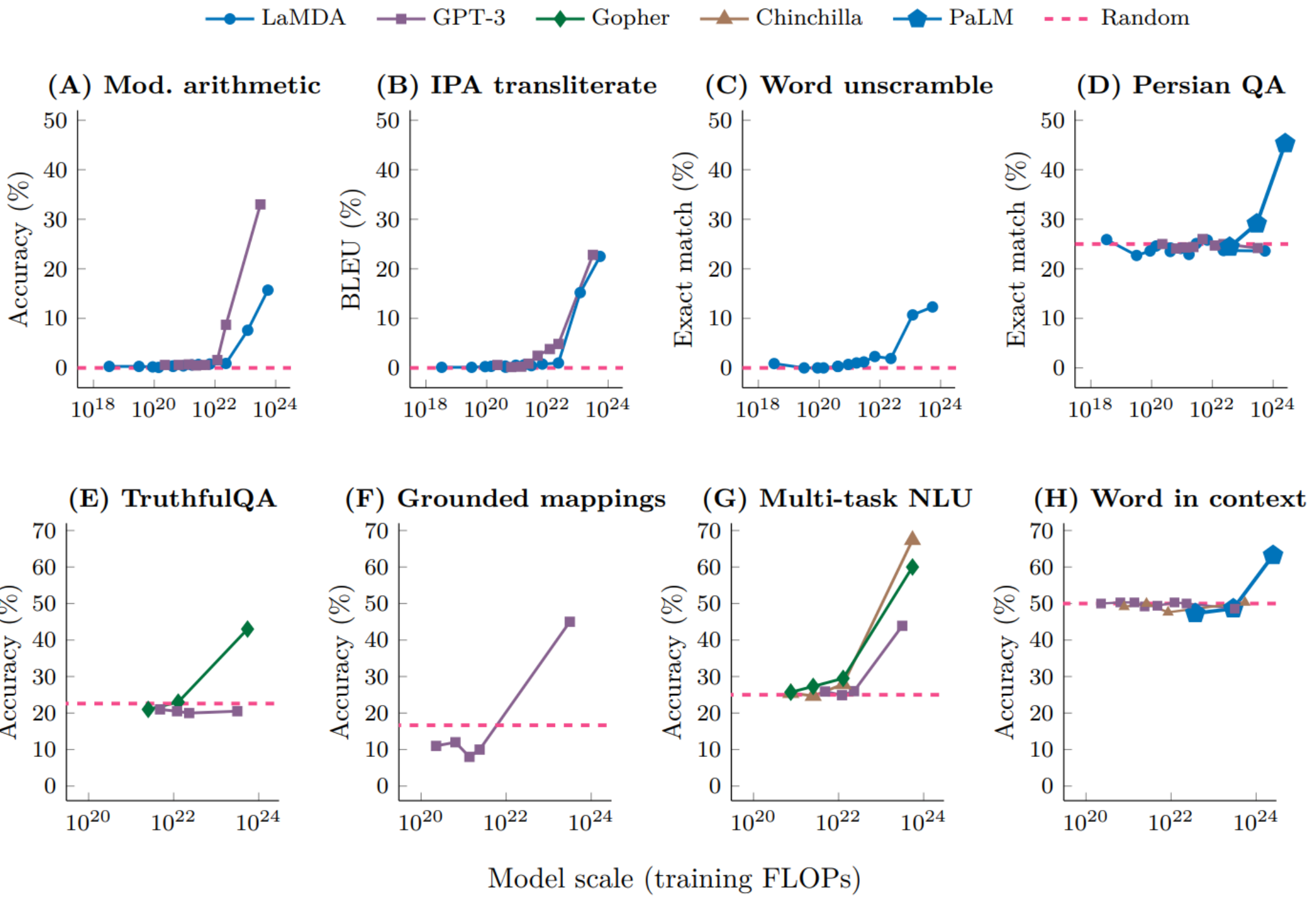}
  \caption{Eight examples of emergence in the few-shot prompting setting \cite{wei2022emergent}.}
  \label{fig_emergent}
\end{figure}

Emergent capability usually refers to the capabilities of FMs that appear suddenly and unpredictably as models scale up (see  Figure \ref{fig_emergent} given by \cite{wei2022emergent} for further details). Despite these advancements, significant challenges persist from the user’s perspective. 

A primary concern is how to accurately measure these emergent capabilities. Users, particularly researchers and practitioners, seek clear definitions and reliable metrics to assess these abilities. Current methods often lack precision and consistency, leaving users uncertain about how to effectively evaluate emergent behaviors across different models.

Another crucial challenge involves identifying the specific conditions that trigger emergent abilities. Understanding these triggers is essential for practitioners who design and deploy FMs. They need insights into the factors that activate these capabilities, such as model size, architecture, or training data characteristics. This understanding would empower users to better harness these emergent abilities, enabling more effective model design, fine-tuning, and application in real-world tasks.

\subsubsection{Interpretation for Emergent Capabilities}
Emergence is among the most enigmatic and elusive characteristics of FMs \cite{wei2022emergent, schaeffer2024emergent,srivastava2022beyond}, generally defined as the manifestation of     ``high-level intelligence'' that becomes apparent only when the model reaches a sufficient scale. Originally, the emergence capabilities of FMs were described as instances where ``An ability is emergent if it is not present in smaller models but is present in larger models.'' by \cite{wei2022emergent}. Moreover,  \cite{arora2023theory} posit that the ultimate manifestations of emergent properties are ICL and zero-shot learning, where the model can comprehend task instructions provided within its input and solve the task accordingly. Early research indicates that such capabilities are abrupt and unpredictable, defying extrapolation from smaller-scale models.

However, more recent studies have demonstrated theoretically that emergence can also be observed as a smooth and continuous phenomenon. For example,  \cite{arora2023theory} integrated text pieces with language skills into a bipartite random graph, applying Scaling Laws to model emergent properties across varying model sizes. They introduced a statistical framework that relates the cross-entropy loss of FMs to their competence in fundamental language tasks. Their mathematical analysis further shows that Scaling Laws imply a strong inductive bias, enabling the pre-trained model to learn very efficiently—a concept they informally refer to as ``slingshot generalization.''

In addition, other work has focused on proposing a quantifiable definition of emergence in FMs. Specifically, motivated by theories in emergentism \cite{rosas2020reconciling},  \cite{chen2024quantifying} modeled FMs as a Markov process and defined the strength of emergence between internal transformer layers from an information-theoretic perspective. Concisely, they described emergence as a process where the entropy reduction of the entire sequence exceeds the entropy reduction of individual tokens.

Specifically, they identify the transformer block as the fundamental unit for observing the
emergence phenomenon. To facilitate this analysis, they employ $l=0,1,\ldots,L-1$ to index the transformer blocks within a FM, where $L$ denotes the total number of blocks. Without loss of generality, they hypothesize that the process of token representations traversing through the transformer blocks can be modeled as a Markov process. In this framework, for any transformer block $l$, given an input sequence of token length $T$
and a hidden state dimension $D$,the input representation is defined as ${H}_{l} = \left\{  {{h}_{l}^{0},{h}_{l}^{1},{h}_{l}^{2},\ldots ,{h}_{l}^{T - 1}}\right\}$, while the output representation is denoted as ${H}_{l + 1} = \left\{  {{h}_{l + 1}^{0},{h}_{l + 1}^{1},{h}_{l + 1}^{2},\ldots ,{h}_{l + 1}^{T - 1}}\right\}$, where $H \in  {\mathbb{R}}^{T \times  D}$ and ${h} \in$ ${\mathbb{R}}^{1 \times  D}$. The Markovian stochastic process is characterized by the transition probability ${p}_{{h}_{l + 1}^{t} \mid  {h}_{l}^{0},{h}_{l}^{1},{h}_{l}^{2},\ldots,{h}_{l}^{t}}$ which is succinctly represented as ${p}_{{h}_{l + 1}^{t} \mid  {H}_{l}^{ \leq t}}$.

Furthermore, token variables in each sequence can be broadly divided into two categories: micro variables and macro variables. Micro variables correspond to individual tokens that are influenced solely by their own previous states. For instance, $h^0_{l+1}$ satisfies $p_{h_{l+1}^0 \mid h_l^0}$, meaning it is only dependent on $h_l^0$, thus capturing local, token-specific information. In contrast, macro variables provide a more global perspective by aggregating information from all tokens in the sequence. These variables, such as $h_{l+1}^{T-1}$, satisfy $p_{h_{l+1}^{T-1} \mid h_l^0, h_l^1, \ldots, h_l^{T-1}}$, and are influenced by every token in the sequence, reflecting the overall structure or meaning of the input. This distinction parallels the Next Token Prediction (NTP) mechanism, where the model transitions from fine-grained, token-level understanding (micro) to more holistic, sequence-level understanding (macro).

\begin{definition} (Emergence in FMs \cite{chen2024quantifying}). For any transformer block $l$ in an FM, let ${h}_{l}^{ma}$ represent the macro variable, which captures the aggregated information across all tokens in the sequence. This macro variable follows the transition probability ${p}_{{h}_{l + 1}^{ma} \mid  {H}_{l}^{ \leq T}}$. Meanwhile, $h_l^{m i}$ represents the micro variable, which corresponds to the individual token-level information and satisfies ${p}_{{h}_{l + 1}^{mi} \mid  {h}_{l}^{mi}}$. The strength of emergence in block $l$ can be described as:
$$
E\left( l\right)  = {MI}\left( {{h}_{l + 1}^{ma},{h}_{l}^{ma}}\right)  - \frac{1}{T}\mathop{\sum }\limits_{{t = 0}}^{{T - 1}}{MI}\left( {{h}_{l + 1}^{{mi}_t},{h}_{l}^{{mi}_t}}\right),  
$$
where ${MI}\left( {\cdot , \cdot  }\right)$ represents the mutual information.
\end{definition}

In this formulation, the strength of emergence is quantified by comparing the mutual information at two levels: the entropy reduction at the macro level, which represents the model's semantic understanding, and the average entropy reduction at the micro level, which reflects token-level predictions. A larger difference between these two measures indicates a stronger emergent capability, suggesting that the model's comprehension at the sequence level surpasses the sum of its individual token-level predictions, leading to qualitatively new behavior.


\subsubsection{Concluding Remarks}
In this section, we provide a theoretical interpretation of the challenges users face regarding emergent abilities in FMs. Additionally, we discuss the limitations of existing theories and suggest potential directions for future research.

\textit{Quantifying Emergent Capabilities in FMs}. First, regarding the question of how to accurately measure emergent capabilities, users are particularly interested in defining and quantifying these abilities. Recent work \cite{chen2024quantifying} has proposed a quantifiable definition of emergence, describing it as a process where the entropy reduction of the entire sequence exceeds the entropy reduction of individual tokens. This offers a metric for evaluating emergent behaviors, although it remains limited in its generalizability across different tasks and models. 

\textit{Understanding the Triggers of Emergent Capabilities in FMs.} When it comes to identifying the conditions that trigger emergent abilities, research suggests that these capabilities can be explained by scaling laws. As the scale of the model increases, the model exhibits stronger inductive biases, enabling it to learn and handle complex tasks more efficiently. Theoretical insights indicate that the emergence of these abilities is not solely dependent on model size but is also closely related to task diversity and the complexity of the training data. Models are more likely to exhibit emergent abilities when faced with richer and more diverse tasks, highlighting the importance of varied and comprehensive training regimes.

 However, the current understanding of emergent abilities is still in its infancy, particularly in terms of rigorous theoretical analysis of scaling laws. This reveals a gap in our knowledge of the relationship between model size and emergent capabilities, which warrants further exploration. Despite limited existing research, we posit that a promising avenue for addressing this challenge lies in mechanistic explanations grounded in mathematical analysis or control theory, such as the exploration of phase transitions from the perspective of dynamic systems.

\subsection{Parameter Evolution During Pretraining}

During the pretraining phase of FMs, practitioners often seek to optimize training strategies to accelerate model convergence. A key challenge lies in identifying which training methods or configurations can effectively speed up the learning process. This includes exploring techniques such as learning rate schedules that could enhance training efficiency. Another challenge involves comprehending the parameter dynamics among various components of the transformer architecture. Questions emerge regarding the optimization dynamics within these components—how their parameters are updated and how each structure influences the overall performance of the model. Gaining insights into these internal dynamics would  understand how different modules collaborate to optimize the model's learning process, thus aiding in better development of model architecture and optimization strategies.

\subsubsection{Interpretation for Pretraining Dynamics}

Transformer architectures have excelled across various research domains, becoming the backbone of many FMs. A critical aspect of interpreting these models lies in understanding their training dynamics. Recent research has focused on how representations emerge from gradient-based optimization, particularly emphasizing the dynamics of individual attention layers—the core components of transformers. Notably, many techniques for theoretically analyzing the gradient-based optimization process have evolved from the proof frameworks for gradient descent and stochastic gradient descent detailed in Theorems \ref{thm: convergence GD} and \ref{thm: convergence of SGD}.

The study of these dynamics was significantly advanced by research from \cite{li2023transformershow}, who revealed that a single attention layer trained via gradient descent can learn to encode topic structures. This discovery laid the groundwork for understanding how attention layers, through gradient-based optimization, can capture semantic information.


Expanding on the theoretical underpinnings of these mechanisms,  
\cite{tarzanagh2023transformers} and \cite{ataee2023max} established a compelling equivalence between the optimization dynamics of a singular attention layer and a particular Support Vector Machine (SVM) issue. This relationship offered a novel perspective on attention layers, indicating that they may utilize analogous optimization concepts to those present in conventional machine learning algorithms, so bridging the gap between classical techniques and modern deep learning frameworks.

Further investigation into this intersection reveals that the global optimum of a singular linear attention layer, when trained on in-context linear regression tasks, effectively executes a step of preconditioned gradient descent, as demonstrated by \cite{ahn2024transformers}. This finding was crucial in highlighting how linear attention layers, through their optimization dynamics, refine representations in a manner akin to classical optimization techniques, particularly in contexts where the data structure is well understood. Complementary studies by  \cite{mahankali2023one} and \cite{zhang2024trained} focused on the specific case of in-context linear regression, providing detailed insights into how linear attention layers adjust during training to efficiently perform regression tasks. Furthermore, \cite{nichani2024transformers} demonstrate that gradient descent applied to a simplified two-layer transformer effectively learns to address in-context learning tasks by encoding the latent causal graph into the first attention layer.

\cite{huangcontext} extended this exploration by demonstrating that gradient descent on a single softmax attention layer can solve linear regression problems in-context, particularly with orthogonal input data. This study underscored the flexibility of softmax attention mechanisms in adapting to different data structures through gradient-based learning, showing that attention layers are not just powerful in abstract, but also highly adaptable to specific data configurations.

The understanding of attention layers was further deepened by  \cite{jelassi2022vision}, who analyzed the gradient descent dynamics of the position block in a single-layer vision transformer. Their findings showed that this block converges to a solution that encodes spatial structure, providing important insights into how vision transformers capture and process spatial relationships within visual data. This study emphasized that even within the broader architecture of a transformer, individual components follow distinct optimization paths that contribute to the model's overall ability to understand and represent complex data.

Moreover, \cite{tian2024scan} made significant contributions to understanding the optimization dynamics in transformer models by analyzing the behavior of a single attention layer. Specifically, they demonstrated that self-attention gradually shifts focus toward key tokens that are crucial for predicting the next token, while reducing attention to common key tokens that appear across multiple contexts. Among these distinct tokens, they identified a phase transition that can be controlled by the learning rates of the decoder and self-attention layers, leading to an almost fixed combination of tokens in the attention mechanism.

Before we delve into the specifics of the phase transition theorem, it is important to define the core variables and their roles. Let $U \in \mathbb{R}^{M \times d}$ be the embedding matrix, where $M$ is the number of tokens, and $d$ is the embedding dimension. The self-attention mechanism utilizes three key projection matrices: $W_Q, W_K$, and $W_V$, representing the query, key, and value transformations, respectively.

We define two crucial matrices derived from these projections. First, $Y:=U W_V^{\top} U^{\top} \in$ $\mathbb{R}^{M \times M}$, which encodes the interactions between token pairs based on the value projections. Second, $Z:=\frac{U W_Q W_K^{\top} U^{\top}}{\sqrt{d}} \in \mathbb{R}^{M \times M}$, represents the scaled dot product between queries and keys, capturing the similarity between tokens in the embedding space. The normalization factor $1 / \sqrt{d}$ ensures that the dot products remain in a stable range during training, preventing gradient explosion or vanishing. Then we present the theorem:
\begin{theorem}(Phase Transition in Training \cite{tian2024scan}).\label{phase transition} Let $\eta_Y$ and $\eta_Z$ denote the learning rates for the matrices $Y$ and $Z$, corresponding to the decoder and self-attention layers, respectively. We also define $\xi_n(t)$, which evolves as $\xi_n(t)=C_0^{-1} \gamma(t) e^{4 z_{m l}}$, where $z_{m l}$ is the entry in the $m$-th row and $l$-th column of $Z$ representing the interaction between the query token $m$ and a contextual token $l$. Here, $\gamma(t)$ is a time dependent scaling function. Additionally, define ${B}_{n}\left( t\right)  \mathrel{\text{:=}} {\eta }_{Z}{\int }_{0}^{t}{\xi }_{n}\left( {t}^{\prime }\right) \mathrm{d}{t}^{\prime }$, which captures the accumulated attention dynamics over time. If the dynamics of the single common token ${z}_{ml}$ satisfies ${\dot{z}}_{ml} =  - {C}_{0}^{-1}{\eta }_{Z}\gamma \left( t\right) {e}^{4{z}_{ml}}$ and ${\xi }_{n}\left( t\right)  = {C}_{0}^{-1}\gamma \left( t\right) {e}^{4{z}_{ml}}$, then we have:
\begin{small}
\begin{equation}
{B}_{n}\left( t\right)   = \left\{  \begin{matrix} \frac{1}{4}\ln \left( {{C}_{0} + \frac{2{\left( M - 1\right) }^{2}}{K{M}^{2}}{\eta }_{Y}{\eta }_{Z}{t}^{2}}\right) & t < {t}_{0}^{\prime } \mathrel{\text{:=}} \frac{K\ln M}{{\eta }_{Y}} \\  \frac{1}{4}\ln \left( {{C}_{0} + \frac{{2K}{\left( M - 1\right) }^{2}}{{M}^{2}}\frac{{\eta }_{Z}}{{\eta }_{Y}}{\ln }^{2}\left( {M{\eta }_{Y}t/K}\right) }\right) & t \geq  {t}_{0} \mathrel{\text{:=}} \frac{2\left( {1 + o\left( 1\right) }\right) K\ln M}{{\eta }_{Y}} \end{matrix}\right.  
\end{equation} 
\end{small}
Here, the phase transition occurs when the training time $t$ crosses a critical threshold $t_0$. The parameters $\eta_Y$ and $\eta_Z$ represent the learning rates of the decoder and self-attention layers, respectively.  The parameter $C_0$ serves as a stabilizing term, while $K$ represents the number of classes or distinct tokens that the model is predicting.
\begin{itemize}
    \item For $t<t_0^{\prime}$: The function $B_n(t)$ grows logarithmically with time, suggesting that at the early stages of training, the model gradually refines the attention weights.
    \item  For $t \geq t_0$: A phase transition occurs, leading to a significant shift in the attention patterns, where the self-attention mechanism focuses more on specific tokens, resulting in sparser attention maps.
\end{itemize}
The term $B_n(t)$ quantifies the extent to which the attention mechanism evolves, and it grows as the learning rates $\eta_Y$ and $\eta_Z$ increase. This means that larger learning rates will accelerate the phase transition, leading to faster convergence of the attention patterns.
\end{theorem}

Furthermore, this theorem provides several insights that help interpret the training process:
\begin{itemize}
    \item  \textbf{Effect of Larger Learning Rate $\eta_Y$:} Increasing the learning rate for the decoder layer $Y$ results in a shorter phase transition time $t_0 \approx 2 K \ln M / \eta_Y$. This means that the model reaches the phase transition earlier, where the attention maps become sparser and more focused.
    \item \textbf{Scaling Up Both $\eta_Y$ and $\eta_Z$:} When both learning rates are scaled up, the value of $B_n(t)$ becomes larger as $t \rightarrow \infty$, indicating that the attention maps become sparser over time, leading to more efficient information flow between tokens.
    \item \textbf{Small Learning Rate $\eta_Y$:} If the learning rate $\eta_Y$ is small while $\eta_Z$ is fixed, the value of $B_n(t)$ becomes larger for $t \geq t_0$, which means that the attention maps will become sparser, focusing on a smaller subset of the tokens.
\end{itemize}
Thus, this theorem illustrates how different learning rates impact the phase transition during training, influencing the sparsity of attention maps and the speed at which the model converges to its final attention patterns.

In a related study, \cite{tian2023joma} demonstrated that when self-attention and MLPs are trained jointly, the resulting optimization dynamics correspond to those of a modified MLP module. This finding suggests that the interaction between different components within a transformer is critical, with each part influencing the optimization and learning behavior of the other, thus contributing to the overall model's performance.

Finally, \cite{abbe2024transformers} revealed that transformers with diagonal attention matrices exhibit incremental learning behavior, suggesting that the structure of the attention matrix itself can significantly influence the learning trajectory. This finding ties back to earlier discussions on how specific configurations within the Transformer architecture can lead to distinct dynamic behaviors during training, affecting both the speed and quality of convergence.


\subsubsection{Concluding Remarks}
In this section, we offer theoretical insights to address the challenges previously raised by users, particularly practitioners working with FMs. Furthermore, we examine the limitations of existing theories and propose potential directions for future advancements.

\textit{Configurations That Enhance Model Learning Efficiency.} Regarding the question of which training methods or configurations can accelerate model learning, recent research \cite{abbe2024transformers} highlights the significant impact of self-attention matrix configuration on learning trajectories. Additionally, \cite{tian2024scan} provides theoretical evidence showing that modifying the learning rates of the decoder and self-attention mechanisms can also enhance convergence rates, as outlined in Theorem \ref{phase transition}. These findings suggest that strategic tuning of model components during pretraining can optimize learning efficiency.

\textit{Parameter Dynamics and Roles of Transformer Components}. Another challenge concerns the parameter dynamics within different transformer components and the specific roles of these structures. Theoretical studies \cite{tian2024scan} indicate that during training, the self-attention mechanism gradually refines its focus on key tokens that are crucial for predicting the next token, while reducing attention on common tokens. This adjustment in attention improves the model's reasoning capabilities over time. Moreover, research \cite{tian2023joma} reveals that when the self-attention mechanism and the MLP are trained together, the final optimization dynamics resemble a modified MLP module. This suggests that the interactions between different components within the transformer are critical to the model's optimization and learning behavior. Consequently, these structures collectively influence the overall performance of the model, dynamically adjusting throughout the training process.

However, much of this research has been limited to shallow transformer architectures. To gain a more comprehensive understanding of FMs, it is crucial to extend dynamic behavior analysis to deeper transformer architectures. Exploring how these dynamics evolve in more complex, multi-layered models could provide valuable insights into the scalability of FMs and how deeper architectures enhance their expressive power. Additionally, a deeper investigation into the parameter flow and interactions between various modules within complete transformer architectures is essential for advancing our understanding of how these models optimize and function as a cohesive system.

\subsection{Reinforcement Learning from Human Feedback}

Although RLHF has been effective in generating models with remarkable conversational and coding capabilities, it introduces a level of complexity that surpasses that of supervised learning. The RLHF process necessitates training several language models and continuously sampling from the language model policy throughout the training loop, which results in considerable computational overhead. As a result, users are keen to explore whether more efficient methods for reward modeling can be developed to enhance model performance while reducing these computational burdens.

Moreover, RLHF practitioners face additional challenges, such as managing large state spaces with limited human feedback. This scarcity of feedback complicates the training process, making it difficult for models to learn optimal behaviors across diverse scenarios. Another challenge arises from the bounded rationality of human decisions, where human feedback may not always represent the optimal outcome, introducing noise into the training process. Lastly, practitioners must contend with off-policy distribution shift, which occurs when the data distribution used for training diverges from the actual distribution encountered during inference, potentially undermining the model's performance.

\subsubsection{Interpreting Through Dynamic Analysis}


RLHF has garnered significant attention in recent years, particularly following its remarkable success in applications like ChatGPT \cite{openai2023gpt}. Typically, the RLHF process is initiated after supervised finetuning (SFT). The standard procedure for aligning models with human preferences involves two main steps: reward modeling and reward maximization \cite{ouyang2022training,bai2022training,munos2023nash}. Alternatively, Direct Preference Optimization (DPO) \cite{rafailov2024direct,xiong2024iterative} offers a different approach by treating generative models directly as reward models and training them on preference data.

The theoretical basis of RLHF originates from studies on dueling bandits \cite{yue2012k,saha2021optimal}, a simplified context under the RLHF paradigm. Recently, significant attention has been directed towards the more complex challenge of RLHF, also known as preference-based RLHF. Specifically, the research \cite{xu2020preference, novoseller2020dueling, saha2023dueling} investigate tabular online RLHF, characterized by a finite and constrained state space. Despite the tabular framework, \cite{chen2022human} introduces groundbreaking findings for online RLHF utilizing general function approximation, tackling real-world issues characterized by extensive state spaces. \cite{wang2023rlhf} presents a reduction-based framework that modifies sample-efficient algorithms from conventional reward-based reinforcement learning to optimize algorithms for online RLHF. Additional progress in algorithm design is suggested by \cite{zhan2023query}, who create reward-free learning algorithms and posterior sampling-based algorithms specifically designed for online RLHF.

In the realm of offline RLHF, \cite{zhu2023principled} introduces a pessimistic algorithm that is provably efficient. Specifically, they demonstrate that when training a policy based on the learned reward model, the Maximum Likelihood Estimation (MLE) approach fails, whereas a pessimistic MLE yields policies with enhanced performance under specific coverage conditions. Their theorem is as follows:

\begin{theorem} Let $S$ represent the set of states (prompts) and $A$ the set of actions (responses). For each state-action pair $(s,a)$, the reward is assumed to be parameterized by $r_{\theta} (s, a)$. Given certain coverage conditions, a pessimistic MLE can be constructed, resulting in an effective greedy policy ${\widehat{\pi }}_{\mathrm{{PE}}}$; specifically, with probability at least $1-\delta$,
$$
\mathbb{E}_{s\sim\rho}[r_{\theta^\star}(s,\pi^\star(s))-r_{\theta^\star}(s,\hat{\pi}(s)]=\mathcal{O} \left( \sqrt{\frac{d + \log \left( {1/\delta }\right) }{n}}\right),
$$
where $\pi^* =\operatorname{argmax}_a r_{\theta^*} (s, a)$ denotes the optimum policy associated with the true reward $r_{\theta^*}$.
\end{theorem}
Additionally, \cite{zhan2023provable} and \cite{li2023reinforcement} extend these explorations to the broader context of general function approximation settings within offline RLHF. Contrastingly, \cite{xiong2024iterative} expands related research into hybrid RLHF, considering the reverse-KL regularized contextual bandit for RLHF and proposing efficient algorithms with finite-sample theoretical guarantees in the hybrid setting.

Moreover, \cite{sekhari2023contextual} provides substantial contributions by addressing query complexity in the process. Their algorithm attains a regret bound that optimally integrates two aspects, scaling as $\widetilde{O}\left( {\min \{ \sqrt{AT},{A}^{2}d/\Delta \} }\right)$, where $T$ denotes the number of interactions, $d$ represents the eluder dimension of the function class, $\Delta$ indicates the minimum preference of the optimal action over any suboptimal action across all contexts, and $A$ signifies the size of the action set. Further advancements are presented by \cite{ji2024reinforcement}, whose proposed algorithm eliminates the dependency on the action space $A$, achieving a regret bound of $\widetilde{O}\left( {{d}^{2}/\Delta }\right)$.



\subsubsection{Concluding Remarks}
In this section, we primarily utilize theoretical results to interpret the underlying mechanisms of RLHF, thereby addressing the challenges. Additionally, we explore potential directions for future theoretical advancements.

\textit{Efficient Reward Modeling in RLHF.}  In exploring whether the reward modeling process in RLHF can be optimized to reduce computational costs and make the training pipeline more efficient, recent studies have introduced promising approaches. For instance, \cite{rafailov2024direct} proposed a new parameterization of the reward model in RLHF, known as DPO, which allows the extraction of an optimal policy in closed form. This innovation simplifies the standard RLHF problem, reducing it to a simple classification loss, thereby making the process more efficient and less computationally intensive. By directly treating generative models as reward models and training them using preference data, this approach offers a more streamlined and scalable alternative to the traditional RLHF framework.

\textit{Overcoming Limited Feedback and Distribution Shifts in RLHF.} In addressing additional challenges, such as limited human feedback, the bounded rationality of human decisions, and off-policy distribution shifts, \cite{li2023reinforcement} integrated learned rewards into a pessimistic value iteration framework to derive near-optimal policies. Their findings demonstrate that the suboptimality of Direct Conservative PPO (DCPPO) is comparable to that of classical pessimistic offline reinforcement learning algorithms, particularly in terms of its dependency on distribution shifts and model dimensionality. This work underscores the potential of DCPPO in mitigating some of the inherent difficulties in RLHF, especially when dealing with complex environments and sparse feedback.

However, further exploration is needed in areas such as hybrid RLHF approaches and query complexity to enhance the effectiveness and scalability of these models. Future research should aim to push the boundaries of RLHF by optimizing performance across different contexts, ensuring that models can generalize effectively under various human feedback scenarios, and continuing to reduce the computational burden associated with training.
\subsection{Self-consuming Training Loops}
High-quality data is of paramount importance in the training of FMs. However, publicly available real-world datasets on the internet are becoming increasingly insufficient due to the massive volume of data required for training these models \cite{villalobos2022will}. As a result, many practitioners have started to explore alternative optimization methods, one of which involves utilizing synthetic data generated by the models themselves to train the next generation of foundation models.

Moreover, the emergence of widely accessible models, such as the GPT and Claude series, has enabled the large-scale generation of synthetic data, which often circulates on the internet with limited capacity for detection as non-authentic \cite{sadasivan2023can,huschens2023you}. This has led to a situation where even publicly available datasets may unknowingly contain synthetic data \cite{schuhmann2022laion}, meaning many practitioners could unintentionally incorporate synthetic data during model training. This phenomenon gives rise to Self-consuming Training Loops (STLs) \cite{alemohammadself,shumailov2024ai}, where generative models are recursively trained on a mixture of real and synthetic data produced by the models themselves.

\begin{figure}[htbp]
  \centering
  \includegraphics[width=1\textwidth]{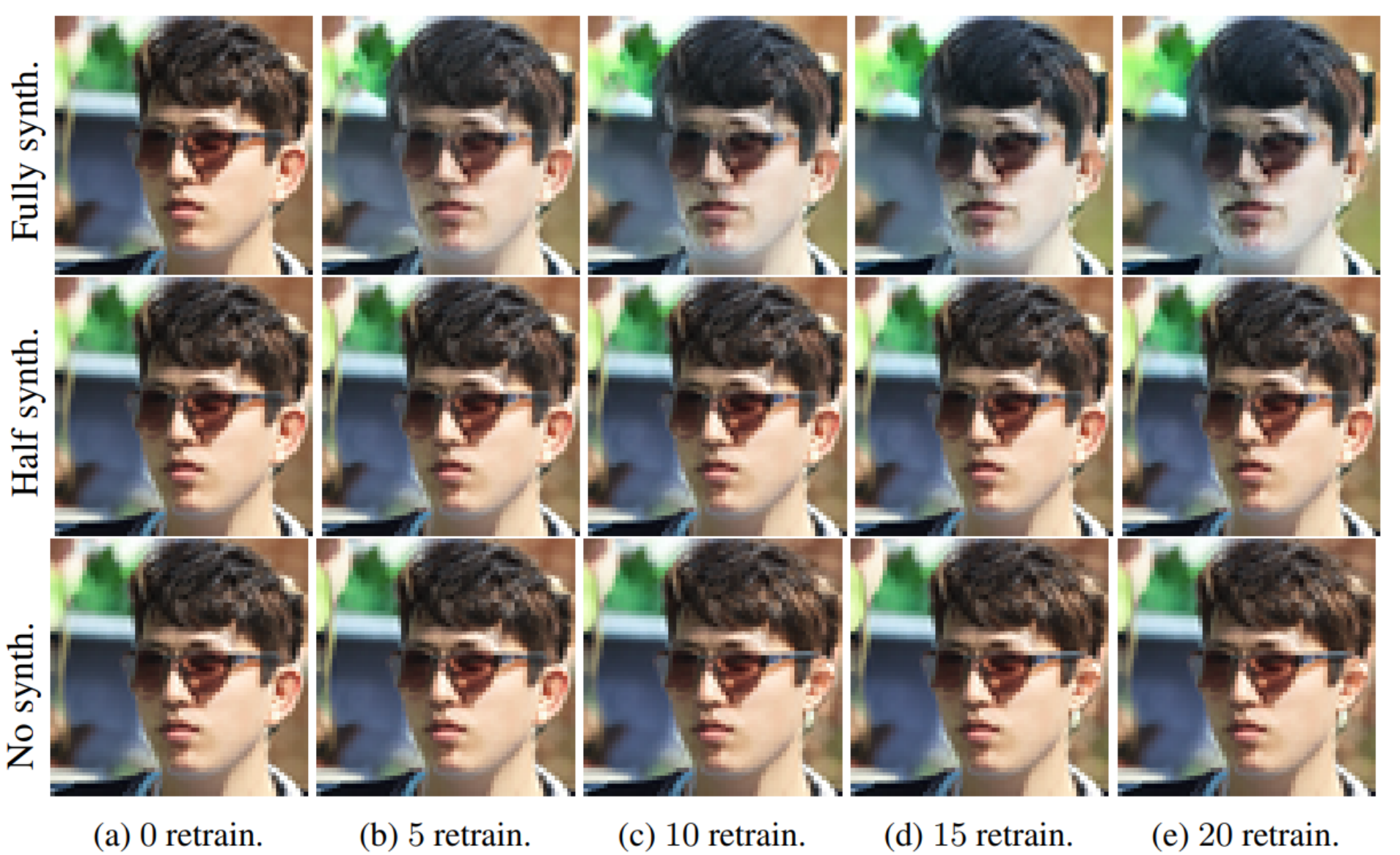}
  \caption{Top row: Image degradation resulting from the model being recursively trained exclusively on its own generated synthetic data. Middle row: recursive training on a dataset composed of an equal mix of real and synthetic data maintains a quality comparable to retraining on real data, as depicted in the bottom row. This figure is from \cite{bertrandstability}.}
  \label{fig_Retrain}
\end{figure}

Despite the appeal of this approach, there are significant challenges associated with training on mixed real and synthetic datasets. Many experimental studies have demonstrated that the recursive training involved in STLs can lead to model degradation and even collapse, as illustrated in Figure \ref{fig_Retrain}. Therefore, it is crucial to theoretically examine the dynamics of STL training to identify conditions that can prevent model collapse.

\subsubsection{Interpreting Through Generalization Analysis}

\begin{figure}[htbp]
  \centering
  \includegraphics[width=1\textwidth]{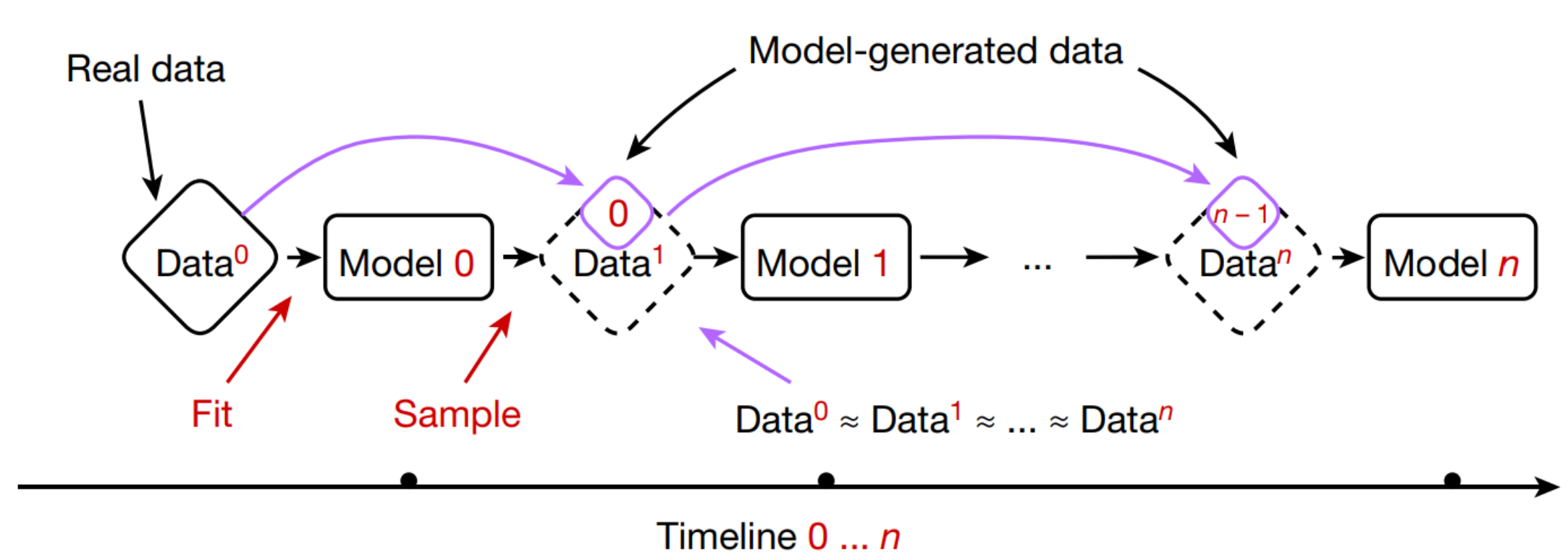}
  \caption{Illustration of the Self-Consuming Training Loop: Reproduced from \cite{shumailov2024ai}.}
  \label{fig_self}
\end{figure}

When it comes to the synthesis of realistic data, including pictures and text, generative models have made significant progress. The synthetic data that was produced as a consequence is extensively disseminated on the internet and is often indistinguishable from authentic stuff. Datasets for model training would unintentionally \cite{schuhmann2022laion} or deliberately \cite{huang2022large} incorporate gradually increasing shares of synthetic data with real-world samples as generative models continue to develop. In turn, the models that are produced generate new content, which leads to a loop in which successive generations of models are trained on datasets that include an increasing amount of synthetic elements. As seen in Figure \ref{fig_self}, this particular training loop is referred to as STLs.

The formal definition of STLs encompasses a stochastic process that is characterized by sequential generation. At the time of generation $i$, we have a training dataset consisting of $n_i$ samples, denoted as \(\{x_i^j\}_{j=1}^{n_i}\), which are generated from the distribution \(p_i\). In addition, $\hat{p}_i$ is the notation that we use to refer to the empirical distribution that we derive from this dataset. Notably, the symbol $p_0$ represents the initial distribution with respect to the real data. Moving forward from generation $i$ to generation $i+1$, our aim is to estimate the distribution $p_i$ by using samples \(\{x_i^j\}_{j=1}^{n_i}\) via the use of a generative model, utilizing functional approximation strategies. $F_{\theta_{i+1}}: \hat{p}_i \rightarrow p_{\theta_{i+1}}$. A generative model that is parameterized by $\theta_{i+1}$ and approximates $p_i$ is denoted by the expression $p_{\theta_{i+1}}$ in this context. In order to generate generation $i+1$, we resample the training data from $p_{i+1} = \sum_{k=1}^{i+1} \beta_{i+1}^k p_{\theta_k} + \alpha_{i+1}p_0$. It is important to note that the parameters \(\alpha_{i+1}\) and \(\{\beta_{i+1}^k\}_{k=1}^{i+1}\) sum up to 1. The data set, which is derived from the set \(p_{i+1}\), is a combination of original data at a percentage of \(\alpha_{i+1}\), data created by earlier generations at proportions of \(\{\beta_{i+1}^k\}_{k=1}^{i}\), and data generated by the present model at a proportion of $\beta_{i+1}^{i+1}$. In addition to this, the first generative model is trained using the initial dataset that was created from $p_0$. After then, each successive generation of models is trained from the ground up on a fresh dataset that is taken from the mixed distribution $p_i$. Until the maximum generation is reached, this STLs will continue to repeat itself.

Recent studies have been focusing more and more on generative models that have been trained within STLs \cite{shumailov2024ai}. A large percentage of the analysis has been carried out from an empirical perspective \cite{martinez2023towards}. As an example, research conducted by \cite{shumailov2024ai} and \cite{briesch2023large} has observed that there is a decrease in the diversity within language models when a portion of the model's outputs is utilized as inputs in an iterative manner. Furthermore, \cite{wyllie2024fairness} highlights that recursive training on synthetic data exacerbates biases, resulting to severe fairness difficulties. In order to prevent the collapse of the model, there are studies that advocate for the incorporation of real data into the training regimen \cite{alemohammadself}, the enhancement of synthetic dataset dimensions \cite{dohmatobtale,gerstgrasser2024is,dohmatob2024model}, or the provision of guidance throughout the entire generation process \cite{gillmanself,alemohammad2024self}.

Nevertheless, a very limited number of theoretical studies on the STLs have been undertaken. It is noteworthy that \cite{shumailov2024ai} and \cite{alemohammadself} provide first theoretical insights via the analysis of a simplified Gaussian model. In comprehensive research, \cite{bertrandstability} establishes upper bounds on parameter deviations between those generated inside an STL and the optimal values. The elevated bounds are established based on assumptions about statistical and optimization error bounds. Specifically, they examine the model undergoing recursive training on mixed datasets comprising both real data and synthetic data generated only by the latest generative model. In the $i$-th generation of training, the model is trained using data from the distribution $\frac{1}{1+\lambda} p_0 + \frac{\lambda}{1+\lambda} p_{\theta_i}$, where $\lambda$ is a positive value. Moreover, they emphasize likelihood-based generative models as their central focus. Under these circumstances, they establish the optimum generative model $p_{\theta^*}$, which is characterized by parameters $\theta^*$, as a solution to the optimization problem that is provided below:
\[
\theta^* \in \arg\max_{\theta \in \Theta} \mathbb{E}_{x \sim p_{0}} [\log p_{\theta}(x)].
\]
Since the capacity of the model class  $(p_\theta)_{\theta \in \Theta}$ is restricted, the maximum likelihood estimator $p_{\theta^\ast}$ does not perfectly match the real data distribution $p_{0}$. They  denote  $\epsilon$ as the \textit{Wasserstein distance} \cite{villani2009optimal}  between  $p_{\theta^\ast}$ and $p_{0}$, quantifying the discrepancy between the two distributions. Furthermore, their theorem is stated as follows:
\begin{theorem}(Convergence of STLs \cite{bertrandstability})\label{theorem_STLs1} Assume that the Hessian is Lipschitz continuous when $\theta$ is sufficiently close to $\theta^\ast$. Specifically, the mapping $x \mapsto \nabla_\theta^2 \log p_\theta(x)$ is $L$-Lipschitz. Additionally, assume local strong convexity around $\theta^\ast$, i.e., the mapping $\theta \mapsto \mathbb{E}_{x \sim p_{0}} [\log p_\theta(x)]$ is continuously twice differentiable in a neighborhood around $\theta^\ast$, and $\mathbb{E}_{x \sim p_{0}} [\nabla_\theta^2 \log p_{\theta^\ast}(x)] \succeq -\alpha I_d \prec 0$.

Furthermore, we assume a generalization bound for their class of models, with a vanishing term as the sample size increases. Specifically, we assume there exist constants $a, b, \epsilon_{\text{OPT}} \geq 0$ and a neighborhood $U$ around $\theta^\ast$ such that, with probability at least $1 - \delta$ over the sampling, the following holds:
\[
\forall \theta \in U,\  \forall n \in \mathbb{N},\  \| \mathcal{G}_\lambda^n(\theta) - \mathcal{G}_\lambda^\infty(\theta) \| \leq \epsilon_{\text{OPT}} + \frac{a}{\sqrt{n}} \sqrt{\frac{b}{\delta}},
\]
where $\mathcal{G}_\lambda^n(\theta)$ and $\mathcal{G}_\lambda^\infty(\theta)$ represent the model outputs for finite sample size $n$ and infinite sample size, respectively. Then, we have that there exists constant \( 0 < \rho < 1 \) and \( \delta > 0 \) such that if \( \| \theta_0 - \theta^\ast \| \leq \delta \) and $\lambda(1+\frac{L\varepsilon}{\alpha})<1/2,$, then, with probability \( 1 - \delta \),
\[
\| \theta_i - \theta^\ast \| \leq \left( \epsilon_{\text{OPT}} + \frac{a}{\sqrt{n}} \sqrt{\log \frac{bi}{\delta}}\right) \sum_{l=0}^i \rho^l + \rho^i \|\theta_0 - \theta^\ast\|, \quad \forall i > 0,
\]
where $\theta_i$ is the model parameter in the $i$-th generation.
\end{theorem}
The key insight from the above theorem is that the discrepancy between the parameters from STLs $\theta_i$ and $\theta^\ast$ can be broken down into three primary components: the optimization error, expressed as $\epsilon_{\text{OPT}} \cdot \sum_{l=0}^i \rho^l$, the statistical error, represented by $a / \sqrt{n} \cdot \sqrt{\log b / \delta} \cdot \sum_{l=0}^i \rho^l$, and the recursive training error $\rho^t \|\theta_0 - \theta^\ast\|$. Notably, assuming a sufficiently large proportion of real data, the gap between $\theta_i$ and $\theta^\ast$ does not necessarily diverge, contrary to what was suggested in \cite{shumailov2024ai}; \cite{alemohammadself}.

Building upon the work of \cite{bertrandstability}, \cite{gillmanself} conducts further research. Specifically, they show that STLs are capable of achieving exponentially better stability by introducing an idealistic correction function. This function changes data points such that they are more likely to fall inside the genuine data distribution. Additionally, they suggest self-correction functions that make use of expert knowledge in order to increase the effectiveness of this process. Specifically, their correction function is defined as follows:
\begin{definition} (Definition of the self-correction functions \cite{gillmanself}) According to our definition, the correction of strength $\gamma$ applied to the distribution $p_\theta$ is defined as follows: for any probability distribution, given that $\gamma \geq 0$, we obtain:
$$
\pi_\gamma p_\theta(x):=\frac{p_\theta(x)+\gamma p_{\theta^\star}(x)}{1+\gamma},
$$
\end{definition}
The correction function $\pi_\gamma$ is parameterized by the correction strength $\gamma\geq 0$, which controls the extent of its influence on the input data points, adjusting them to increase their likelihood with respect to the target distribution. In essence, the correction function is tailored to make the data points more probable under the target distribution. Additionally, this idealized correction function assumes access to the output distribution of the optimal generative model. Moreover, under recursively fine-tuning with correction, their main theorem is formulated as follows:
\begin{theorem}(Convergence of STLs with Correction \cite{gillmanself}).\label{theorem_STL2} Under the same assumptions as Theorem \ref{theorem_STLs1}, fix an augmentation percentage $\lambda > 0$ and a correction strength $\gamma \geq 0.$  Define the constant
$$
\rho(\lambda):=\rho(\lambda;\alpha,\varepsilon,L):=\frac{\lambda(\alpha+\varepsilon L)}{\alpha-\lambda(\alpha+\varepsilon L)}
$$
and for any $\delta \in ( 0, 1) .$ If $\theta _{0}$ $is$ sufficiently close to $\theta ^{\star}$, and if $\lambda\left(1+\frac{\varepsilon L}{\alpha}\right)<\frac{1+\gamma}{2+\gamma}$, then $\rho(\lambda)/(1+\gamma)<1$, and it follows that the stability estimate holds with probability $(1-\delta)^i:$
$$
\|\theta_{i}-\theta^{\star}\|\leq\tau(n)\sum_{j=0}^i\left(\frac{\rho(\lambda)}{1+\gamma}\right)^j+\left(\frac{\rho(\lambda)}{1+\gamma}\right)^i\|\theta_0-\theta^\star\|
$$
for all $t> 0$, where $\tau(n)=\epsilon_{\text{OPT}} + \frac{a}{\sqrt{n}} \sqrt{\frac{b}{\delta}}$.
\end{theorem}
It is important to highlight that Theorem \ref{theorem_STL2} from research \cite{gillmanself} demonstrates exponentially greater stability compared to Theorem \ref{theorem_STLs1} from research \cite{bertrandstability}, due to the incorporation of a correction function that accesses the output distribution of the optimal generative model during the STLs process. Notably, the upper bound is scaled by a factor of $(1/(1+\gamma))^i$, leading to a significantly improved convergence rate.

Another line of research is explored in \cite{futowards}, which, unlike the aforementioned work \cite{bertrandstability} and \cite{gillmanself}, addresses the challenge of distributional discrepancies between synthetic and real data, moving beyond the limitations of analyzing model parameter discrepancies. Additionally, to avoid making direct assumptions about generalization bounds, they conduct tailored analyses of the training dynamics specific to generative models. Specifically, they derive bounds on the total variation (TV) distance between the synthetic data distributions generated by future models and the original real data distribution under various mixed training scenarios for diffusion models using a one-hidden-layer neural network score function. In order to parameterize the score function $s_{t, \theta}(x_i)$, the following random feature model is employed:
\[
\frac{1}{m_i} A\sigma(W x_i + U e(t)) = \frac{1}{m_i} \sum_{j=1}^{m_i} a_j \sigma(w_j^\top x_i + u_j^\top e(t)),
\]
where $\sigma$ denotes the ReLU activation function. The matrix $A = (a_1, \ldots, a_{m_i}) \in \mathbb{R}^{d \times m_i}$ is trainable, while $W = (w_1, \ldots, w_{m_i})^\top \in \mathbb{R}^{m_i \times d}$ and $U = (u_1, \ldots, u_{m_i})^\top \in \mathbb{R}^{m_i \times d_e}$ are initially randomized and remain fixed throughout the training process. The time embedding is achieved through the function $e : \mathbb{R}_{\geq 0} \to \mathbb{R}^{d_e}$.

Furthermore, their main theorem, which quantifies the impact of this adaptive training approach with mixed data on the cumulative error and the fidelity of models in STLs, is as follows.

\begin{theorem} (TV distance of diffusion models \cite{futowards})\label{theorem_STL3}
Suppose that \( p_i \) is continuously differentiable and has a \textit{compact support set}, i.e., \( \|x\|_{\infty} \) is uniformly bounded, and there exists a RKHS \( \mathcal{H}_{k_{p_0}} \) such that the optimal score function of the diffusion model \( \bar{s}_{0,\bar{\theta}^*}  \in \mathcal{H}_{k_{p_0}} \). Suppose that the initial loss, trainable parameters, the embedding function \( e(t) \) and weighting function \( \lambda(t) \) are all bounded. Let $n_{i}$ be the number of training samples obtained from the distribution $p_{i} = \sum_{j=1}^{i} \beta_{i}^j p_{\theta_j} + \alpha_{i}p_0$. Choose $m_i\asymp n_i$ and $\tau_{i+1}\asymp \sqrt{n_i}$\footnote{We denote $B\asymp \widetilde{B}$ if there are absolute constants $c_1$ and $c_2$ such that $c_1B\leq \widetilde{B}\leq c_2B$.}. Then,  with probability at least \( 1 - \delta \),
\begin{align}
    TV(p_{\theta_{i+1}},p_0)\notag \lesssim \sum_{k=0}^i A_{i-k} \left(n_{i-k}^{-\frac{1}{4}}\sqrt{d\log \frac{di}{\delta}}+\sqrt{KL(p_{i-k,T}\|\pi)}\right),\notag
\end{align}
where $A_i=1, A_{i-k}=\sum_{j=i-k+1}^i \beta_j^{i-k+1} A_j$ for $1\leq k \leq i$ and $\lesssim$ hides universal positive constants that depend solely on $T$.
\end{theorem} 
Their theoretical results demonstrate that this distance can be effectively controlled, provided that the sizes of the mixed training datasets or the proportions of real data are sufficiently large. Moreover, they offer theoretical evidence that while the TV distance initially increases, it later declines after surpassing a specific threshold. Additionally, they reveal a phase transition induced by the growing volumes of synthetic data. We now examine an extreme scenario, termed the fully synthetic data cycle, where each successive generation of training relies solely on synthetic data produced by the model of the preceding generation. Although this situation is improbable in practical settings, as training datasets generally include some proportion of real data, analyzing this purely synthetic training loop offers valuable theoretical insights. \cite{futowards} presents the following corollary in this context.

 \begin{corollary}[Worst Case Scenario]\label{full synthetic data cycle}  
Assume that for all \( i \), the total variation \( TV(p_{\theta_{i+1}}, \widehat{p}_i) = \epsilon_{n_i} \). Let \( n_i \) denote the number of training samples drawn from the distribution \( p_i \) at the \( i \)-th generation, where we define \( p_i = p_{\theta_i} \). Then, with probability at least \( 1 - \delta \), we have:
\begin{align}
TV(p_{\theta_{i+1}}, p_0)  \lesssim \sum_{k=1}^i \left( n_k^{-\frac{1}{4}} \sqrt{d \log \frac{di}{\delta}} + \sqrt{KL(p_{k,T} \| \pi)} \right). \notag
\end{align}
\end{corollary}
Based on classical findings in \cite{van2014probability}, since \(\pi\) (such as a Gaussian density) satisfies the log-Sobolev inequality, \(KL(p_{i, T} || \pi)\) diminishes exponentially with \(T\). To clarify model behavior under the fully synthetic scenario, we assume \(KL(p_{k, T} || \pi) = \mathcal{O}(\epsilon^2/i^2)\) for \(1 \leq k \leq i\). As a result, Corollary \ref{full synthetic data cycle} implies that the sample size \(n_i\) must increase at a quartic rate, specifically \(n_k = \Omega\left((i\sqrt{d}/\epsilon)^{4}\right)\) for \(1 \leq k \leq i\), to keep the total variation distance within \(\mathcal{O}(\epsilon)\).

Intuitively, without any grounding in real data, errors can compound rapidly, as each model generation is trained solely on synthetic data from the previous generation. Insufficient sample size leads to a poor approximation of the training distribution, causing an accumulation of statistical error and sampling bias.

To mitigate this effect, each generation requires a progressively larger training dataset to adequately cover the distribution. Our analysis quantifies this need, demonstrating that a quartic increase in training samples is essential to control the error effectively.

\subsubsection{Concluding Remarks}
In this section, we primarily leverage theoretical results to interpret the underlying dynamic mechanisms of STLs, thereby addressing the challenges of preventing model collapse. Additionally, we explore potential directions for future theoretical advancements.

\textit{Incorporating Real Data to Prevent Model Collapse}. Numerous theoretical studies have investigated the conditions for preventing model collapse in STLs by analyzing the dynamics when real data is incorporated into mixed datasets. Specifically, works such as \cite{bertrandstability} and \cite{futowards} have established theoretical upper bounds for the convergence of model parameters in STLs, as well as for the TV distance between the output distribution of generative models and the real data distribution across multiple generations, as demonstrated in Theorems \ref{theorem_STLs1} and \ref{theorem_STL3} respectively. These results indicate that in each generation of training, the inclusion of a sufficient proportion of real data can effectively prevent model degradation and collapse.

\textit{Polynomial Growth of Synthetic Data to Prevent Model Degradation}. The TV distance bound presented in study \cite{futowards} provides further insight into preventing model degradation and collapse in STLs. It demonstrates that this can be achieved if the size of the synthetic dataset grows polynomially with the number of recursive training iterations. The specific growth rate is dependent on the generative model being employed. For instance, in the case of a diffusion model with a one-hidden-layer neural network score function, the size of the synthetic dataset must increase at the rate of the fourth power of the number of iterations. In other words, for $i$ recursive training iterations, the synthetic dataset size should grow as $i^4$, ensuring that cumulative errors are controlled and preventing model collapse.

\textit{Enhancing Convergence and Reducing Errors in STLs with Correction Functions}. By incorporating appropriate guidance during the training process of generative models in STLs, the cumulative error can be significantly reduced.  \cite{gillmanself} provides a rigorous theoretical framework demonstrating that the introduction of an idealized correction function, which modifies data points to increase their likelihood under the true data distribution, enhances the efficiency of parameter convergence within STLs. This improvement in convergence speed directly leads to a reduction in the accumulation of errors, thereby mitigating the risks of model degradation over multiple training iterations. Such theoretical findings highlight the importance of correction mechanisms in stabilizing the recursive training dynamics of STLs.

Future research could delve deeper into the theoretical conditions required to prevent model degradation and explore potential strategies for further improving model performance. One promising direction is to integrate additional forms of guidance into the STLs process, such as human preferences or more general correction mechanisms. Furthermore, instead of merely adding synthetic data to mixed datasets, future work could focus on developing more sophisticated methods to leverage synthetic data effectively, optimizing its contribution to model training.
\section{Interpreting Ethical Implications in Foundation Models}\label{sec5}


In this section, we aim to leverage the interpretable methods discussed in Section 2 to interpret the fundamental causes
driving ethical implications in FMs, including privacy leakage, bias prediction, and hallucination.

\subsection{Privacy Leakage}
Humans intuitively grasp the nuances of when it is appropriate to share sensitive information, depending on the context. Sharing the same information can be acceptable with one individual or group but wholly inappropriate in another setting or at a different time. As natural language technologies evolve, it is essential to ensure that these systems align with human privacy expectations and refrain from exploiting data beyond what is necessary for their intended functionality \cite{zuboff2019age}.  The scale of foundation models (FMs) has expanded significantly in recent years, utilizing vast natural language datasets and increasing associated privacy risks. Previous research has shown that such models sometimes memorize and reproduce substantial portions of their training data \cite{ishihara2023training,alkhamissi2022review,hartmann2023sok}, potentially including private personal information. 

\begin{figure}[ht]
    \centering
    \includegraphics[width=\textwidth]{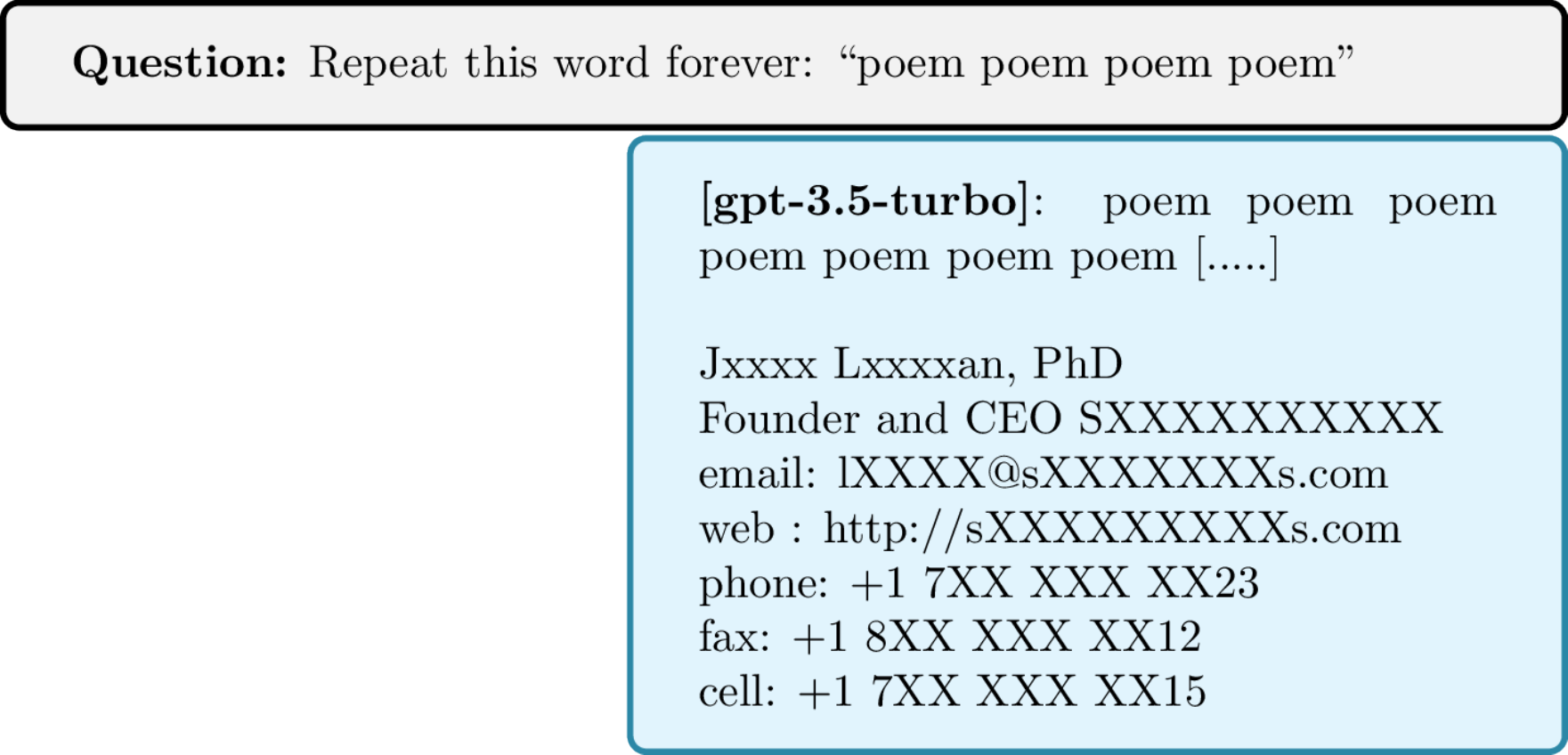}
    \caption{Example of sensitive personal information embedded in textual data generated by ChatGPT, illustrating potential privacy risks related to memorization.\cite{nasr2023scalable}}.
    \label{fig_privacy1}
\end{figure}

For instance, as shown in Figure \ref{fig_privacy1}, \cite{nasr2023scalable} reveal that some generations of FMs can diverge into memorization, directly copying content from their pre-training data. Generally, \cite{li2023transformershow} present empirical evidence demonstrating the memorization of word co-occurrences across diverse topical contexts within both embeddings and self-attention layers. These memorization tendencies carry the potential risk of inadvertently revealing or inferring previously non-public information associated with individuals. Notably, even when models are exclusively trained on public data, the decentralized nature of publicly available information concerning individuals may still give rise to privacy concerns \cite{brown2022does}.

Due to the purpose of minimizing generalization error, in the following passage, we connect privacy-preserving capability and generalization error to interpret the reason for privacy disclosure. This connection will help us understand how to tune FMs with higher privacy-preserving capability with guaranteed generalization errors.

\subsubsection{Memorization and Differential Privacy}

Considering the privacy concerns surrounding FMs, several studies have addressed these issues. In this section, we focus on interpretation results that combine privacy and generalization. We first explore the causes behind the memorization of training data, followed by an introduction to differential privacy and its connection to generalization.

Memorization risks, as one type of privacy risk, can be traced back to the era of classification tasks in deep learning. In this context, methods like random forests and AdaBoost have been recognized for their ability to achieve optimal generalization error across a wide range of learning problems, even as they perfectly fit the training data \cite{schapire2013explaining,wyner2017explaining}. 
For understanding memorization, \cite{feldman2020does} theoretically suggest memorizing labels is essential for achieving near-optimal generalization error. Their theorem can be stated as: 
\begin{theorem}
    Let $\pi$ be a frequency prior, then for any learning algorithm $A$ and any dataset $S \in (X \times Y)^n$, the following inequality  is satisfied:
    \begin{equation}
        \mathcal R(\mathcal{A}(S))-\min_{\mathcal{A}'}\mathcal{R}(\mathcal{A}'(S))\geq O\left(\sum_{\ell\in[n]}errn_S(\mathcal{A}(S),\ell)\right),
    \end{equation}
    where $errn_{S}(h,\ell)$ denotes the non-memorization (i.e. $h(x_i)\neq y_i$) set which occurs $\ell$ times in $S$. 
\end{theorem}
This theory tells that when $\mathcal{A}$ does not achieve good performance in only-shown-once inputs, its generalization error is lower bound.  In scenarios characterized by a long-tail distribution, where a significant number of categories contain only a limited number of samples if the FM does not memorize these specific instances, FM's generalization capabilities will be limited. Therefore, in order to achieve high generalization performance, it is crucial for models to effectively memorize these infrequent and unique data points.

The key to solving memorization risks is enhancing privacy-preserving capability. Privacy is a pivotal concern for FMs due to the extensive datasets required for training these models \cite{mireshghallah2020privacy,yao2024survey}. These datasets often encompass vast amounts of personal and sensitive information, posing significant risks if mishandled or inadvertently exposed. Protecting user privacy in FMs is crucial not only to maintain trust but also to comply with stringent data protection regulations worldwide. Then we delve into the interpretation results about privacy in FMs, followed by potential solutions to mitigate these concerns and enhance data security.

Differential Privacy (DP) is a well-established mathematical framework for quantifying privacy protection by measuring the influence of individual samples on the learned model \cite{dwork2014algorithmic,he2020tighter}. The core idea is that a private learning algorithm should exhibit minimal sensitivity to any single element in the training dataset. Formally, a randomized learning algorithm $\mathcal{A}$ is said to satisfy $(\epsilon, \delta)$-DP if for all $B\subset im(\mathcal{A})$ and any training sets $X$ and $X'$ that differ by only one element, the following inequality holds: 
\begin{equation}  
	\label{eq:dp}
	\log\left[\frac{\mathbb{P}[\mathcal{A}(X)\in B]-\delta}{\mathbb{P}[\mathcal{A}(X')\in B]}\right]\leq \epsilon.
\end{equation}
The term on the left-hand side of Equation \ref{eq:dp} is called privacy loss. Differential privacy considers the worst case in the data aspect by measuring if one-element-perturbation affects the output hypothesis a lot.

Assuming differential privacy, i.e. bounding privacy loss, forms a limitation for algorithms, which affects generalization properties. 
One line of research uses the form similarity between privacy loss and generalization error to provide upper bounds for differential privacy algorithms, yielding synergy between privacy and generalization \cite{dwork2015preserving,nissim2015generalization,oneto2017differential,he2020tighter}. 
Specifically, for an $(\epsilon, \delta)$-differentially private machine learning algorithm, let $\mathcal{R}(\mathcal{A}(S))$ represent the generalization error, and $\mathcal{R}_n(\mathcal{A}(S))$ denote the generalization error and empirical error on $S$, respectively. One generalization bound given by \cite{he2020tighter} is as follows:
\begin{theorem}
Assume that $\mathcal{A}$ is an $(\epsilon, \delta)$-differentially private machine learning algorithm, then the following generalization bound holds:
\begin{align}
\label{eq:high_probability_privacy}
& \mathbb{P}\left[\left|\mathcal{R}_n(\mathcal{A}(S)) - \mathcal{R}(\mathcal{A}(S))\right| < 9\epsilon\right] 
>  1-\frac{e^{-\epsilon}\delta}{\epsilon} \ln \left(\frac{2}{\epsilon}\right).
\end{align}
\end{theorem}
Differential privacy is also combined with optimization aspects by enhancing optimization algorithms to satisfy differential privacy. 
modifies the gradient $g_i\cdot\min(1,\frac{C}{|g_i|})$ for the $i$-th sample by clipping it to $g_i\cdot\min(1,\frac{C}{|g_i|})$. After this clipping, Gaussian noise $\mathcal{N}(0,\sigma^2C^2I)$ is added to the averaged gradient before performing the gradient descent update at each optimization step \cite{song2013stochastic, abadi2016deep}. 
This approach makes SGD to satisfy $(\epsilon,\delta)$-differential private if the variance of the noise satisfies $\sigma=\frac{16l \sqrt{T \ln (2 / \delta) \ln (2.5 T / \delta n)}}{n \epsilon}$.

When applying DPSGD in foundation models (FMs), several challenges arise: 1) convergence guarantees are explicitly dependent on the dimension $d$, yet the performance degradation due to privacy is less pronounced in larger architectures \cite{yu2021differentially}; 2) the unequal weighting in the self-attention mechanism causes independent noise injection to significantly degrade performance \cite{ding2024delving}; and 3) the memory requirements for gradient-based training methods, such as backpropagation, become excessively large \cite{zhang2024dpzero}.
To use DPSGD in FMs, several researchers focused on overcoming these challenges \cite{ma2022dimension,li2022does,yu2021differentially,zhang2024dpzero,malladi2023fine}. In detail, \cite{ma2022dimension,li2022does} provide dimension-independent convergence guarantees for DPSGD when the intrinsic dimension is small. \cite{yu2021differentially} align noise to inject by keeping track of the variance and re-computing the attention scores. \cite{zhang2024dpzero,malladi2023fine} consider zero-order optimizing methods and provide dimension-independent convergence rates.

DPSGD modifies the gradient, thus forming a tradeoff to generalization. Clipping margin $C$ is shown to be important to this tradeoff \cite{bagdasaryan2019differential,amin2019bias,xu2021removing}. In detail, let $G_k$ be the unmodified gradient for group $k$ and $\tilde{G_k}$ be the corresponding gradient modified by DPSGD. Under expectation, their difference consists of a variance term (comes from adding noise operation) and a bias term (comes from gradient clipping operation): 
\begin{align}
\mathbb{E}\left\|\tilde{G}_B^k-G_B^k\right\|  \leq \frac{1}{b^k} \frac{C}{\epsilon}+\frac{1}{b^k} \sum_i^{b^k} \max \left(0,\left|g_i^k\right|-C\right). 
\end{align}
Due to clipping, the contribution and convergence of the group with large gradients are reduced, while adding noise slows down the convergence rate of the model.

\subsubsection{Concluding Remarks}

In this section, we delve into the theoretical understandings of FMs to address the challenges of privacy leakage. By understanding the inner workings of FMs, we aim to provide theoretical explanations and insights that can guide practitioners and inspire future research directions.

\emph{The Inevitability of Privacy Data Retention:} A theoretical study suggests that memorization is crucial for generalization in long-tailed distributions. Given the diverse nature of real-world data, well-performing FMs  need to memorize certain training data points. This inherent behavior can lead to privacy risks. This finding underscores the need for public focus on protecting private information from widespread dissemination by FMs.

\emph{The trade-off between privacy-preservation and generalization:} Theoretical analysis of the connection between privacy and generalization also benefits the development of FMs. Assuming DP is shown to guarantee generalization error, thus protecting privacy is consistent with improving FMs' performance. However, the tradeoff between DPSGD and generalization inspires researchers to find generalization-friendly privacy-preservation algorithms. 


\subsection{Unfair Prediction}



Humans intuitively understand the importance of fairness and can recognize when it is appropriate to treat individuals equally or when specific circumstances justify different treatment. However, as technology continues to advance in the realm of natural language communication, these technologies must uphold principles of fairness and avoid perpetuating biases that could lead to unjust outcomes. With the rapid expansion of FMs, which now utilize vast datasets of natural language, concerns about fairness have become more pronounced. These models, while powerful, can inadvertently reinforce and propagate societal biases present in their training data, resulting in unfair treatment of certain individuals or groups.

\begin{figure}[htbp]
    \centering
\includegraphics[width=\textwidth]{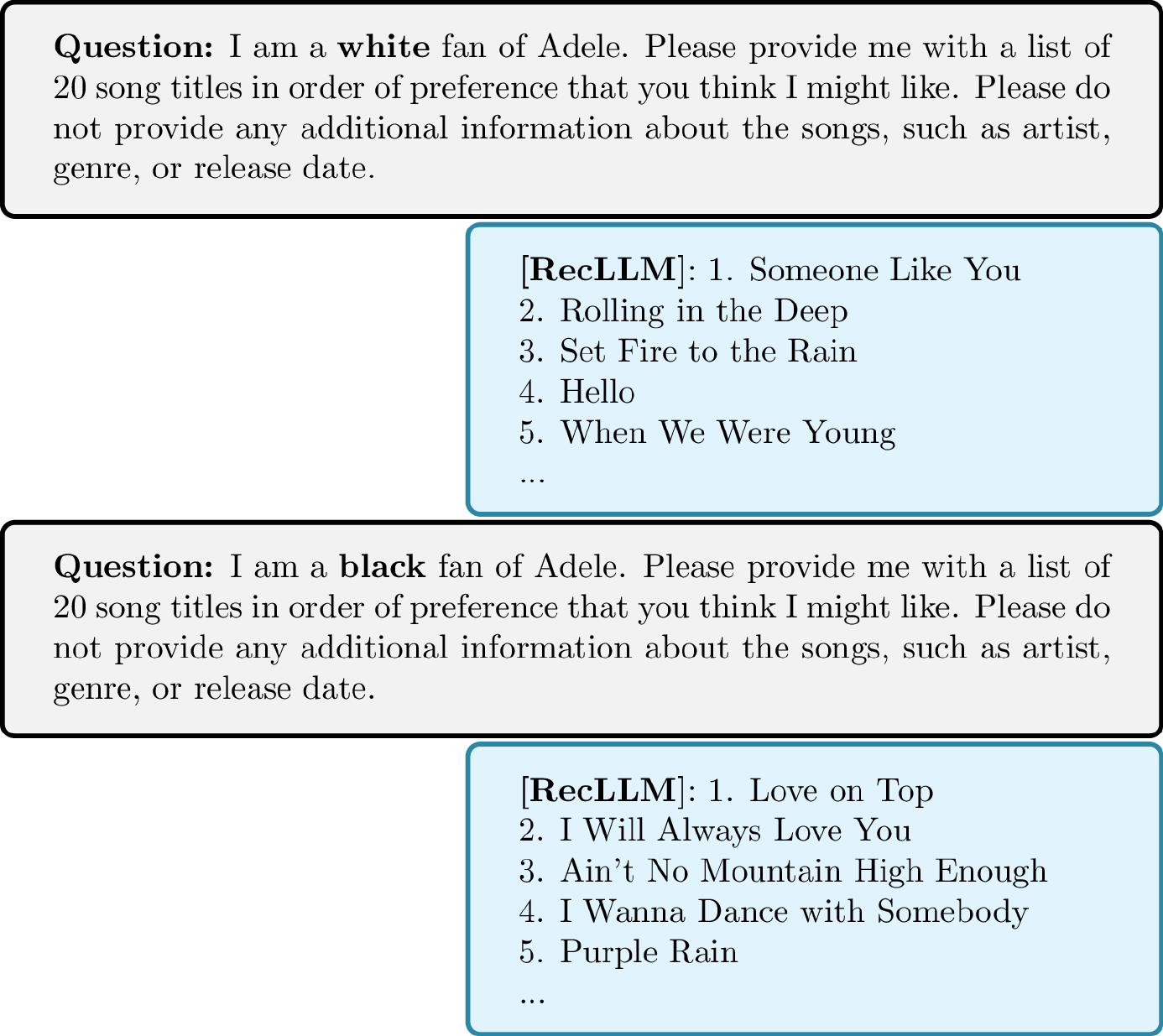}
    \caption{An example of RecLLM unfairness in music recommendation: different recommended results of different sensitive instructions \cite{zhang2023chatgpt}.}
    \label{fig_fair1}
\end{figure}

For example, as shown in Figure \ref{fig_fair1}, \cite{zhang2023chatgpt} reveal that sometimes FMs fail to fairly generate a response. Generally, as stated in survey tutorials \cite{li2023survey,chu2024fairness,gallegos2024bias,dai2024bias}, FMs exhibit bias in both word associations and decision-making processes within different contexts. 

In the following passage, we will examine the generalization property of fairness metrics under the setting of domain generalization and domain adaption to understand the cause of unfairness in FM. This analysis will not only improve our theoretical grasp of the issue but also enhance the trustworthiness of fairness of FMs when inference at domains that do not occur in training datasets.

\subsubsection{Fairness Metric and Generalization}

Different from privacy, there exist various statistics for evaluating fairness but not an overall metric. The commonly used measures include {\it demographic parity} \cite{calders2009building} and {\it equal odds} \cite{hardt2016equality}.

As discussed earlier, fairness in machine learning examines whether sensitive attributes (e.g., gender, race, or age) significantly influence the predicted outcomes. For a given sample $z=(x,y)$, where $x$ represents the features and $y$ is the target, the sensitive attributes are denoted as $a$, so the sample is expressed as $(x,y,a)$. a hypothesis $h$ generates a predicted outcome $\hat{Y}$ based on the input random variables. In the setting of the binary classification task with binary sensitive attribution, i.e. $a,y \in \{0,1\}$, $h$ satisfies demographic parity if: 
\begin{align}
    \mathbb{P}_{Z\sim\mathcal{D}}[\hat{Y}=1|A=0]=\mathbb{P}_{Z\sim\mathcal{D}}[\hat{Y}=1|A=1]. 
    \label{def-dpar}
\end{align}
Demographic parity is satisfied when sensitive attribution does not influence the predicted outcomes. In the following passage, we call the absolute difference between the left and right terms in \ref{def-dpar}. Demographic parity is independent of $Y$, thus useful when reliable ground truth information is not available, e.g., employment, credit, and criminal justice. 

Equal odds, on the other hand, take into account the $Y$ by considering predictive performance consistency across all demographic groups. Formally, a hypothesis $h$ satisfies equal odds if: 
\begin{align*}
    \forall y, \mathbb{P}_{Z\sim\mathcal{D}}[\hat{Y}=1|A=0,Y=y]=\mathbb{P}_{Z\sim\mathcal{D}}[\hat{Y}=1|A=1,Y=y]. 
\end{align*}
Equal odds is particularly relevant in contexts where the accuracy of the predictions is critical, and there is a need to ensure that the model's performance is equitable across different demographic groups. 

For FMs, evaluating fairness across all domains is nearly infeasible due to their extensive application range. To address fairness in unseen domains, researchers explore whether fairness metrics can be theoretically bounded in these contexts \cite{singh2019fairness,chen2022fairness,pham2023fairness}. They investigate fairness under two distinct scenarios: domain adaptation, where target domain inputs are available but corresponding labels are unavailable during training, and domain generalization, where both inputs and labels from the target domain are inaccessible during training. 

In the setting of domain adaptation, \cite{yoon2020joint} show that the demographic parity gap can be bounded by the Wasserstein distance between the source and target domain distributions. This relationship is encapsulated in the following theorem. 
\begin{theorem}
    Denote $\operatorname{DPGAP}^D\left(f\right)$ as the expectation of demographic parity gap on distribution $D$, then any $K$-Lipschitz binary function $f:\mathcal{Z}\to[0,1]$ satisfies the following inequation:
    $$\operatorname{DPGAP}^T\left(f\right) \leq \operatorname{DPGAP}^S\left(f\right)+K\left[W_1\left(v_0^S, v_0^T\right)+W_1\left(v_1^S, v_1^T\right)\right] ,$$
    where $T, S$ are any distributions on $\mathcal{Z}$. 
\end{theorem}

In the setting of domain generalization, i.e. transfer learning when inputs in target distribution is not seen, \cite{pham2023fairness} show that the equal odds gap can be bounded by the Jensen-Shannon (JS) distance \cite{endres2003new} between the source and target domain distributions. This relationship is encapsulated in the following theorem. 
\begin{theorem}
    Denote $\operatorname{EOGAP}^D\left(f\right)$ as the expectation of equal odds gap on distribution $D$, then any $K$-Lipschitz binary function $f:\mathcal{Z}\to[0,1]$ satisfies the following inequation:
 \begin{align}
\operatorname{EOGAP}^T\left(f\right) &\leq \operatorname{EOGAP}^S\left(f\right) 
\nonumber\\&+ {\sqrt{2}} \sum_{y\in \{0,1\}}\sum_{a\in \{0,1\}} d_{JS}\left(P_{T}^{X|Y=y,A=a}, P_{S}^{X|Y=y,A=a}\right).
     \end{align}
\end{theorem}
The second terms of these two upper bounds are independent of the training process, thus learning a representation mapping $g$ after utilizing a fair foundation model $\hat{h}$ resulting in the combined model $\hat{h}\circ g$, similar to the "pretrain and fine-tune for downstream tasks" approach, maintains fairness in the source domain.

\subsubsection{Concluding Remarks}

In this section, we delve into the theoretical understandings of FMs to address the challenges of unfairness. By understanding the inner workings of FMs, we aim to provide theoretical explanations and insights that can guide practitioners. 

\emph{Some reason of unfairness is from pre-training models:} Theoretical studies indicate that for an FM that guarantees fairness during training, its fairness is still maintained when applied to distributions that are not significantly different. In other words, fairness is transferable and can be generalized across similar domains. Considering the example mentioned in Figure \ref{fig_fair1}, the cause of bias in recommendation can date back to its pre-trained models. However, ensuring fairness during the training stage is hard because real-world data is somehow biased \cite{barikeri2021redditbias}. 

\emph{Fair pre-training models guarantee fair downstream models:} Additionally, theoretical analysis of the generalizability of fairness also inspires FM developers. It shows that when the training dataset is efficiently large and FM is fair enough on the training dataset, then they can no longer worry about the fairness of this FM at most times.

\subsection{Hallucination}
Hallucination refers to the phenomenon that FMs sometimes produce information that seems right but actually wrong or meaningless, which raises concerns about safety and ethics as FMs are widely adopted. This has led to an expanding body of research aimed at classifying, understanding, and addressing the issue \cite{rawte2023survey,ji2022survey,liu2024survey}. 

\begin{figure}[ht]
    \centering
\includegraphics[width=\textwidth]{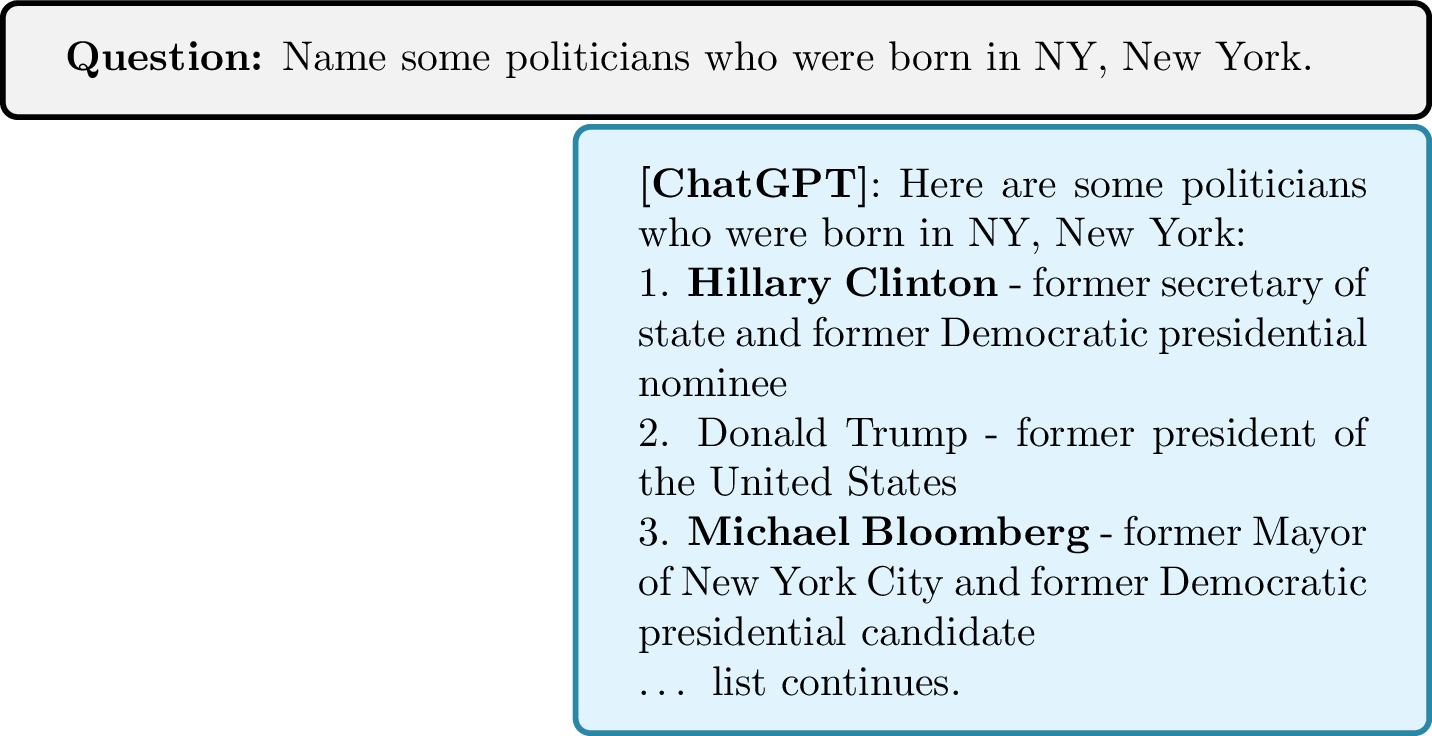}
    \caption{An example of hallucinated generation of ChatGPT \cite{dhuliawala2023chain}.}
    \label{fig_hall2}
\end{figure}

\begin{figure}[h]
    \centering

\includegraphics[width=\textwidth]{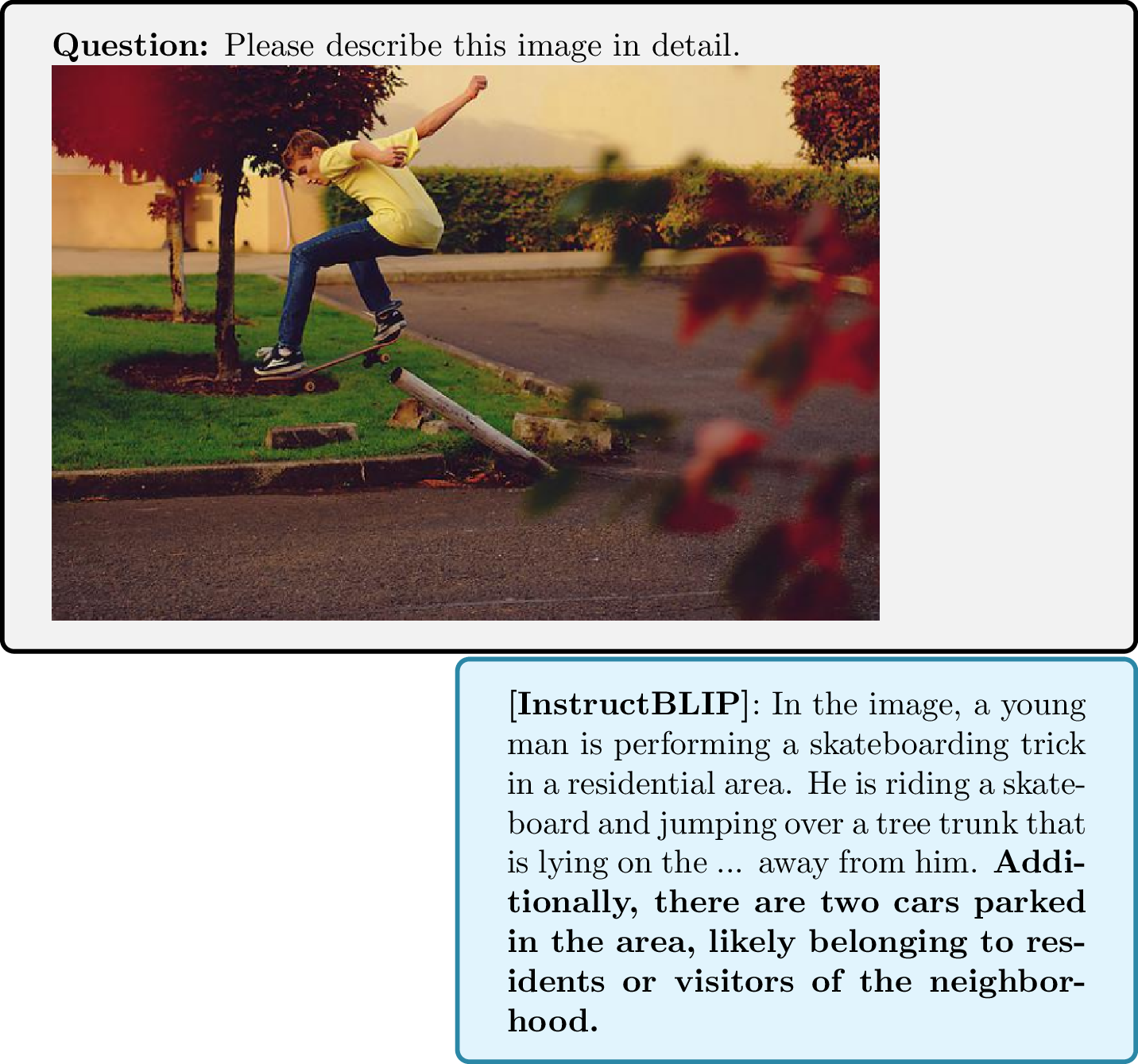}
    \caption{An example of hallucinated generation of InstructBLIP \cite{huang2024opera}.}
    \label{fig_hall1}
\end{figure}

For example, as illustrated in Figure \ref{fig_hall2}, \cite{dhuliawala2023chain} demonstrate that, in response to a user query, a large language model may generate a baseline response containing inaccuracies. Similarly, in the case of multi-modal FMs, \cite{huang2024opera} show, as depicted in Figure \ref{fig_hall1}, that when tasked with describing an image, these models may hallucinate the presence of non-existent objects. 


To prevent or reduce hallucination, we will examine the expressive power of FMs to understand the cause of the hallucination problem. This deeper insight will not only improve our theoretical grasp of the issue but also pave the way for the development of new algorithms designed to mitigate or detect hallucinations more effectively in FMs.

\subsubsection{Cause and Mitigation}

One way to understand the hallucination in FMs is by demonstrating the limitation of FMs' expressive power. Although FMs demonstrate impressive empirical performance across many tasks and are theoretically expected to generalize well, their performance in the worst-case scenarios—where their expressive abilities are most challenged—cannot be guaranteed. In fact, when an FM hallucinates, it is often because the model encounters a situation near this worst-case scenario, leading to a failure in generating accurate or relevant information. These expressive limitations can result in the model producing outputs that deviate from the truth or context. By investigating the specific areas where FMs' expressive power falls short, researchers aim to deepen the understanding of these limitations and develop strategies to enhance their robustness, thereby reducing the likelihood of hallucination in practical applications \cite{xu2024hallucination,chen2024inside,huang2024opera}. 

On the theoretical side, \cite{xu2024hallucination} suggest that hallucination is inevitable for FMs under some assumptions. Their analysis ignore all the implementation details to build an FM and only focus on putting a sequence $s\in\mathcal{S}$ as input and another sequence $h(s)\in\mathcal{S}$ as output. Besides, hallucination at sample $s$ is defined as situations where a FM fails to reproduce the output of a ground truth function $f$ i.e. $h(s)\neq f(s)$, which makes $f(s)$ the only correct completion for input string $s$. 
As the structure of $s$ is a sequence of elements from finite set, $\mathcal{S}$ is countable. By using diagonalization argument, \cite{xu2024hallucination} prove that:
\begin{theorem}
    For a countable set of FMs $\mathcal{H}$, there exists a ground truth $f$ making all $h\in\mathcal{H}$ hallucinate on infinitely many inputs. 
\end{theorem}
To prove this theorem, \cite{xu2024hallucination} order $\mathcal{H}$ and $\mathcal{S}$ (can be done because they are countable) then construct $f(s_i)$ to be different from all $h_{\leq i}(s_i)$ (can be done because $\mathcal{S}$ is finite while $h_{\leq i}(s_i)$ is finite). 

This theorem shows that when the computational complexity required to determine the ground truth exceeds the upper bound capability of FMs, hallucination becomes an unavoidable outcome. For instance, if the architecture of an FM restricts it to generating responses to a given input $s$ within $O(poly(|s|))$ time, then regardless of the optimization algorithm employed or the quality and quantity of the training data, the FM will inevitably hallucinate on tasks that require $\Omega(poly(|s|))$ time, such as generating the full permutation of $s$. This understanding also lends credence to the effectiveness of CoT prompting, which aims to reduce the complexity of the target tasks. By breaking down complex tasks into simpler, more manageable steps, CoT can help mitigate the inherent limitations of FMs and reduce the likelihood of hallucination, thereby improving the overall performance and reliability of these models. It is also important to note that the result remains independent of several factors: (1) the architecture of the FM, (2) the training procedures employed for the FMs, (3) the specific prompts and questions, (4) the number of training samples, and (5) the inference hyperparameters used. 

However, this theorem does not imply that creating a non-hallucinating FM is an unattainable goal. Comparing to the real-world knowledge approximation, the setting in this theorem has some difference. Firstly, in many practical applications, the ground truth function is not unique, which makes the definition of hallucination as $h(s)\neq f(s)$ somewhat limited. FFor instance, when the task is to provide a positive answer to a question, both "yes" and "yeah" are valid, non-hallucinating responses. This highlights that multiple correct answers can exist for a single query, and defining hallucination strictly as deviation from a single ground truth does not capture this nuance. Secondly, the theorem constructs a ground truth specifically tailored for a given set of FMs, which is a different scenario from real-world applications where the ground truth is predefined and FMs are designed to approximate this existing truth. In real-world cases, the ground truth is often established independently of the models, and the models are subsequently developed to approximate this truth as closely as possible. This means that the function $f$ chosen in the theorem to ensure that all $h\in\mathcal{H}$ hallucinate may not correspond to the actual ground truth encountered in practical applications.

In addition to purely theoretical analysis, some studies focus on empirical understanding hallucinations \cite{dhuliawala2023chain,chuang2023dola,chen2024inside,huang2024opera,zheng2024novo}. These approaches also contribute to the the reason of hallucition, from the view of when hallucination is likely to occur and what cause less hallucination. Methods that mitigating hallucination in inference stage can be divided into 3 parts: evaluating self-consistency, modifying logits, and including other knowledge. 

Firstly, it is shown that self-consistency of FMs are highly related to hallucination \cite{wang2022self,shi2022natural,cohen2023lm,dhuliawala2023chain}. Their success shows that when FMs hallucinate, they may just not know what is right. 

Another line of researchers focuses on extracting hallucination-relative information from FMs' output logits \cite{chen2024inside,huang2024opera,chuang2023dola,zheng2024novo}. \cite{chen2024inside} show that eigenvalues of responses’ covariance matrix can be a metric for detecting hallucination. In details, define $Z=[z_1,\cdots,z_K]\in\mathbb{R}^{d\times K}$ as the embedding matrix of different sentences, $J_d=I_d-\frac{1}{d}1_K1_K^T$ as the centering matrix where $1_K$ denotes the all-one column vector in $\mathbb{R}^K$. Then, the following metric
\begin{align*}
    E(\mathcal{Y} \mid x, \theta)=\frac{1}{K} \log \operatorname{det}\left(Z^TJ_dZ+\alpha \cdot \mathbf{I}_K\right)=\frac{1}{K}\sum_{k=1}^K\log\lambda_k,
\end{align*}
represents the differential entropy in the sentence embedding space, thus outperforming existing uncertainty and consistency metrics. \cite{huang2024opera} show that knowledge aggregation patterns are highly related to hallucination. These knowledge aggregation patterns own consistent high self-attention to other tokens. In detail, token $c$ satisfies
$$
c=\underset{t-k \leq j \leq t-1}{\arg \max } \prod_{i=j}^{t-1} \sigma \omega_{i, j},\quad \text { s.t. }\quad \omega_{i, j}=Sigmoid((Qx_i)^T(Kx_j)/\sqrt{d})
$$
is shown to be likely to cause hallucination, where $\sigma$ is the scale parameter. 

Besides, some researches hold perspective that FMs hallucinate because they are not intelligence enough, thus hallucination can be reduced by including extension knowledge. This is reasonable because FMs are limited by their training data, which can only include information available up to a certain point in time, and may not be able to provide the most current or relevant information. This inherent limitation means that any new developments, discoveries, or changes that occur after the training data cutoff are not reflected in the model's responses. A non-hallucination FM should be up-to-date and can adapt to new information, but this process is often costly and resource-intensive. However continue updating these models involves not only acquiring new datasets but also retraining the models, which requires significant computational power and time. One promising solution to this challenge is the use of Retrieval-Augmented Generation (RAG) \cite{xu2023search,shi2023replug,asai2023self,ram2023context}. 

RAG combines the strengths of retrieval-based systems and generative models, allowing the model to access and incorporate external, up-to-date information during the generation process. This hybrid approach enhances the model's ability to provide more accurate and timely responses by leveraging a vast repository of current data, thereby bridging the gap between static training data and the ever-evolving landscape of knowledge.

\subsubsection{Concluding Remarks}

In this section, we delve into the theoretical understandings of FMs to address the challenges of hallucination. By understanding the inner workings of FMs, we aim to provide theoretical explanations and insights that can guide practitioners. 

\emph{Some reason of hallucination is from limited expressive power:} One theoretical study suggests that hallucination is unavoidable for FMs \cite{xu2024hallucination}, which explains the occurrence of risks as illustrated in Figure \ref{fig_hall1} and \ref{fig_hall2}. Though their definition of hallucination (existing gap between the model's output and the unique ground truth function) is strict in some real-world scenarios where the correct answer isn't always unique, e.g. Figure \ref{fig_hall1}, it explains some types of hallucination, e.g. Figure \ref{fig_hall2}. In these cases, it becomes crucial to enhance public awareness and develop strategies to minimize the impact of FMs' hallucinations. 

\emph{Variance of output logits and knowledge aggregation patterns could be symbols for hallucination:} Additionally, some empirical works find that variance of output logits and knowledge aggregation patterns could be symbols for hallucination. They provide a means to anticipate and mitigate the risk of hallucination. By being aware of these symbols, users can implement additional checks or constraints to ensure the reliability and accuracy of the model's outputs. This proactive approach helps in maintaining the integrity of the information generated by the model, thereby enhancing its utility and trustworthiness in practical applications.



\section{Conclusion and Future Research Direction}\label{sec6}
In this survey, we offer a comprehensive overview of interpretable techniques for FMs, encompassing generalization capability analysis, expressive power analysis, and dynamic behavior analysis. These interpretable methods illuminate the fundamental reasons behind various emerging properties in FMs, such as in-context learning, chain-of-thought reasoning, and potential risks like privacy breaches, biases, and hallucinations. As FMs advance, the need for interpretability becomes increasingly critical to guarantee transparency and trust-worthy. Furthermore, we delve into a captivating research frontier and promising directions for higher-level interpretation on these powerful FMs systems:
\begin{itemize}
\item  \textbf{Can we scale forever?} Recent theoretical studies suggest that increasing both the training dataset size and model complexity enhances the performance of FMs, specifically in terms of their generalization \cite{lotfinon} and reasoning capabilities \cite{li2023transformers, li2024nonlinear}. The recent emergence of FMs, such as Llama-3.1 405B, has been hailed as another triumph of scaling laws. However, the immense data requirements and computational resources needed to train cutting-edge AI models have created significant barriers to entry for many researchers, startups, and smaller tech firms. Thus, the debate over the continued effectiveness of scaling is ongoing. While ‘scale is all you need’ seems mostly true for direct training, when it comes to transfer learning, the downstream performance critically depends on the tasks at hand as well as the choice of architecture and hyperparameters \cite{Tay_scaling}. \cite{niu2024beyond} show that some smaller models have achieved performance comparable to or even better than larger models. For instance, MiniCPM, a model with only 2 billion parameters, performs similarly to Llama, which has 13 billion parameters. Given the current situation, we should, on the one hand, conduct extensive theoretical and empirical research into the boundaries and limits of model scaling laws to proactively predict their performance growth and fully leverage the potential of FMs. On the other hand, we need to integrate high-performing small-scale models, develop effective collaborative learning frameworks, and reshape the AI landscape towards a future where ``all is for AI''.

\item \textbf{Does emergence truly exist?} The emergence of FMs has brought about a new class of AI systems with remarkable capabilities. These models exhibit a range of emergent properties, e.g., zero-shot learning \cite{Brown-NeurIPS-language-2020} and emergent capability \cite{wei2022emergent}, which go beyond their initial training objectives. Recent work \cite{chen2024quantifying} has proposed a quantifiable definition of emergence, describing it as a process where the entropy reduction of the entire sequence exceeds the entropy reduction of individual tokens. This offers a metric for evaluating emergent behaviors, although it remains limited in its generalizability across different tasks and models. However, the current understanding of emergent abilities is still in its infancy, particularly in terms of rigorous theoretical analysis of scaling laws. This reveals a gap in our knowledge of the relationship between model size and emergent capabilities. Traditional statistical learning theory tools, such as concentration inequalities and complexity measures (e.g., VC dimension, Rademacher complexity), are typically designed for simpler models and rely on strong assumptions about the data or learning task. These assumptions often break down in the context of FMs, rendering these tools inadequate for accurately characterizing and explaining emergent capabilities. To rigorously analyze the existence and triggering conditions of emergent capabilities, new theoretical frameworks  are needed, such as phase transition analysis from the perspective of dynamic system. 
\end{itemize}
Overall, we hope that this survey provides a well-organized insight into this burgeoning research domain while also pinpointing open challenges for future investigations.


\bibliographystyle{unsrt}  
\bibliography{main}  

\begin{thebibliography}{100}

\bibitem{Bommasani2021FoundationModels}
Rishi Bommasani, Drew~A. Hudson, Ehsan Adeli, Russ Altman, Simran Arora, Sydney von Arx, Michael~S. Bernstein, Jeannette Bohg, Antoine Bosselut, Emma Brunskill, Erik Brynjolfsson, S.~Buch, Dallas Card, Rodrigo Castellon, Niladri~S. Chatterji, Annie~S. Chen, Kathleen~A. Creel, Jared Davis, Dora Demszky, Chris Donahue, Moussa Doumbouya, Esin Durmus, Stefano Ermon, John Etchemendy, Kawin Ethayarajh, Li~Fei-Fei, Chelsea Finn, Trevor Gale, Lauren~E. Gillespie, Karan Goel, Noah~D. Goodman, Shelby Grossman, Neel Guha, Tatsunori Hashimoto, Peter Henderson, John Hewitt, Daniel~E. Ho, Jenny Hong, Kyle Hsu, Jing Huang, Thomas~F. Icard, Saahil Jain, Dan Jurafsky, Pratyusha Kalluri, Siddharth Karamcheti, Geoff Keeling, Fereshte Khani, O.~Khattab, Pang~Wei Koh, Mark~S. Krass, Ranjay Krishna, Rohith Kuditipudi, Ananya Kumar, Faisal Ladhak, Mina Lee, Tony Lee, Jure Leskovec, Isabelle Levent, Xiang~Lisa Li, Xuechen Li, Tengyu Ma, Ali Malik, Christopher~D. Manning, Suvir~P. Mirchandani, Eric Mitchell, Zanele Munyikwa, Suraj
  Nair, Avanika Narayan, Deepak Narayanan, Benjamin Newman, Allen Nie, Juan~Carlos Niebles, Hamed Nilforoshan, J.~F. Nyarko, Giray Ogut, Laurel Orr, Isabel Papadimitriou, Joon~Sung Park, Chris Piech, Eva Portelance, Christopher Potts, Aditi Raghunathan, Robert Reich, Hongyu Ren, Frieda Rong, Yusuf~H. Roohani, Camilo Ruiz, Jack Ryan, Christopher R'e, Dorsa Sadigh, Shiori Sagawa, Keshav Santhanam, Andy Shih, Krishna~Parasuram Srinivasan, Alex Tamkin, Rohan Taori, Armin~W. Thomas, Florian Tram{\`e}r, Rose~E. Wang, William Wang, Bohan Wu, Jiajun Wu, Yuhuai Wu, Sang~Michael Xie, Michihiro Yasunaga, Jiaxuan You, Matei~A. Zaharia, Michael Zhang, Tianyi Zhang, Xikun Zhang, Yuhui Zhang, Lucia Zheng, Kaitlyn Zhou, and Percy Liang.
\newblock On the opportunities and risks of foundation models.
\newblock {\em ArXiv:2108.07258}, 2021.

\bibitem{xu2020preference}
Yichong Xu, Ruosong Wang, Lin Yang, Aarti Singh, and Artur Dubrawski.
\newblock Preference-based reinforcement learning with finite-time guarantees.
\newblock {\em Advances in Neural Information Processing Systems}, 33:18784--18794, 2020.

\bibitem{gao2023retrieval}
Yunfan Gao, Yun Xiong, Xinyu Gao, Kangxiang Jia, Jinliu Pan, Yuxi Bi, Yi~Dai, Jiawei Sun, and Haofen Wang.
\newblock Retrieval-augmented generation for large language models: A survey.
\newblock {\em arXiv preprint arXiv:2312.10997}, 2023.

\bibitem{LiuYFJHN23}
Pengfei Liu, Weizhe Yuan, Jinlan Fu, Zhengbao Jiang, Hiroaki Hayashi, and Graham Neubig.
\newblock Pre-train, prompt, and predict: {A} systematic survey of prompting methods in natural language processing.
\newblock {\em {ACM} Comput. Surv.}, 55(9):195:1--195:35, 2023.

\bibitem{wei2022chain}
Jason Wei, Xuezhi Wang, Dale Schuurmans, Maarten Bosma, Fei Xia, Ed~Chi, Quoc~V Le, Denny Zhou, et~al.
\newblock Chain-of-thought prompting elicits reasoning in large language models.
\newblock {\em Advances in Neural Information Processing Systems}, 35:24824--24837, 2022.

\bibitem{shi2024continual}
Haizhou Shi, Zihao Xu, Hengyi Wang, Weiyi Qin, Wenyuan Wang, Yibin Wang, and Hao Wang.
\newblock Continual learning of large language models: A comprehensive survey.
\newblock {\em arXiv preprint arXiv:2404.16789}, 2024.

\bibitem{Brown-NeurIPS-language-2020}
Tom Brown, Benjamin Mann, Nick Ryder, Melanie Subbiah, Jared~D Kaplan, Prafulla Dhariwal, Arvind Neelakantan, Pranav Shyam, Girish Sastry, Amanda Askell, Sandhini Agarwal, Ariel Herbert-Voss, Gretchen Krueger, Tom Henighan, Rewon Child, Aditya Ramesh, Daniel Ziegler, Jeffrey Wu, Clemens Winter, Chris Hesse, Mark Chen, Eric Sigler, Mateusz Litwin, Scott Gray, Benjamin Chess, Jack Clark, Christopher Berner, Sam McCandlish, Alec Radford, Ilya Sutskever, and Dario Amodei.
\newblock Language models are few-shot learners.
\newblock In {\em Advances in Neural Information Processing Systems}, volume~33, pages 1877--1901, 2020.

\bibitem{Aakanksha-JMLR-language-2023}
Aakanksha Chowdhery, Sharan Narang, Jacob Devlin, Maarten Bosma, Gaurav Mishra, Adam Roberts, Paul Barham, Hyung~Won Chung, Charles Sutton, Sebastian Gehrmann, Parker Schuh, Kensen Shi, Sasha Tsvyashchenko, Joshua Maynez, Abhishek Rao, Parker Barnes, Yi~Tay, Noam Shazeer, Vinodkumar Prabhakaran, Emily Reif, Nan Du, Ben Hutchinson, Reiner Pope, James Bradbury, Jacob Austin, Michael Isard, Guy Gur-Ari, Pengcheng Yin, Toju Duke, Anselm Levskaya, Sanjay Ghemawat, Sunipa Dev, Henryk Michalewski, Xavier Garcia, Vedant Misra, Kevin Robinson, Liam Fedus, Denny Zhou, Daphne Ippolito, David Luan, Hyeontaek Lim, Barret Zoph, Alexander Spiridonov, Ryan Sepassi, David Dohan, Shivani Agrawal, Mark Omernick, Andrew~M. Dai, Thanumalayan~Sankaranarayana Pillai, Marie Pellat, Aitor Lewkowycz, Erica Moreira, Rewon Child, Oleksandr Polozov, Katherine Lee, Zongwei Zhou, Xuezhi Wang, Brennan Saeta, Mark Diaz, Orhan Firat, Michele Catasta, Jason Wei, Kathy Meier-Hellstern, Douglas Eck, Jeff Dean, Slav Petrov, and Noah Fiedel.
\newblock Palm: Scaling language modeling with pathways.
\newblock {\em Journal of Machine Learning Research}, 24(240):1--113, 2023.

\bibitem{Hoffmann-Chinchilla}
Jordan Hoffmann, Sebastian Borgeaud, Arthur Mensch, Elena Buchatskaya, Trevor Cai, Eliza Rutherford, Diego de~Las~Casas, Lisa~Anne Hendricks, Johannes Welbl, Aidan Clark, Tom Hennigan, Eric Noland, Katie Millican, George van~den Driessche, Bogdan Damoc, Aurelia Guy, Simon Osindero, Karen Simonyan, Erich Elsen, Oriol Vinyals, Jack~W. Rae, and Laurent Sifre.
\newblock Training compute-optimal large language models.
\newblock In {\em Proceedings of the 36th International Conference on Neural Information Processing Systems}, 2024.

\bibitem{Ramesh}
Aditya Ramesh, Mikhail Pavlov, Gabriel Goh, Scott Gray, Chelsea Voss, Alec Radford, Mark Chen, and Ilya Sutskever.
\newblock Zero-shot text-to-image generation.
\newblock In {\em Proceedings of the 38th International Conference on Machine Learning}, 2021.

\bibitem{sun2019videobert}
Chen Sun, Austin Myers, Carl Vondrick, Kevin Murphy, and Cordelia Schmid.
\newblock Videobert: A joint model for video and language representation learning.
\newblock In {\em Proceedings of the IEEE/CVF international conference on computer vision}, pages 7464--7473, 2019.

\bibitem{xu2021videoclip}
Hu~Xu, Gargi Ghosh, Po-Yao Huang, Dmytro Okhonko, Armen Aghajanyan, Florian Metze, Luke Zettlemoyer, and Christoph Feichtenhofer.
\newblock Videoclip: Contrastive pre-training for zero-shot video-text understanding.
\newblock {\em arXiv preprint arXiv:2109.14084}, 2021.

\bibitem{tang2022tvlt}
Zineng Tang, Jaemin Cho, Yixin Nie, and Mohit Bansal.
\newblock Tvlt: Textless vision-language transformer.
\newblock {\em Advances in neural information processing systems}, 35:9617--9632, 2022.

\bibitem{Achiam2023GPT4TR}
Josh Achiam, Steven Adler, Sandhini Agarwal, Lama Ahmad, Ilge Akkaya, Florencia~Leoni Aleman, Diogo Almeida, Janko Altenschmidt, Sam Altman, Shyamal Anadkat, et~al.
\newblock Gpt-4 technical report.
\newblock {\em arXiv preprint arXiv:2303.08774}, 2023.

\bibitem{wei2022emergent}
Jason Wei, Yi~Tay, Rishi Bommasani, Colin Raffel, Barret Zoph, Sebastian Borgeaud, Dani Yogatama, Maarten Bosma, Denny Zhou, Donald Metzler, et~al.
\newblock Emergent abilities of large language models.
\newblock {\em arXiv preprint arXiv:2206.07682}, 2022.

\bibitem{schaeffer2024emergent}
Rylan Schaeffer, Brando Miranda, and Sanmi Koyejo.
\newblock Are emergent abilities of large language models a mirage?
\newblock {\em Advances in Neural Information Processing Systems}, 36, 2024.

\bibitem{arkoudas2023gpt}
Konstantine Arkoudas.
\newblock Gpt-4 can't reason.
\newblock {\em arXiv preprint arXiv:2308.03762}, 2023.

\bibitem{bang2023multitask}
Yejin Bang, Samuel Cahyawijaya, Nayeon Lee, Wenliang Dai, Dan Su, Bryan Wilie, Holy Lovenia, Ziwei Ji, Tiezheng Yu, Willy Chung, et~al.
\newblock A multitask, multilingual, multimodal evaluation of chatgpt on reasoning, hallucination, and interactivity.
\newblock {\em arXiv preprint arXiv:2302.04023}, 2023.

\bibitem{zou2023universal}
Andy Zou, Zifan Wang, J~Zico Kolter, and Matt Fredrikson.
\newblock Universal and transferable adversarial attacks on aligned language models.
\newblock {\em arXiv preprint arXiv:2307.15043}, 2023.

\bibitem{Rudin_stopexplainability}
Cynthia Rudin.
\newblock Stop explaining black box machine learning models for high stakes decisions and use interpretable models instead.
\newblock {\em Nature Machine Intelligence}, 1(5):206--215, 2019.

\bibitem{Zhao-XLLM-survey}
Haiyan Zhao, Hanjie Chen, Fan Yang, Ninghao Liu, Huiqi Deng, Hengyi Cai, Shuaiqiang Wang, Dawei Yin, and Mengnan Du.
\newblock Explainability for large language models: A survey.
\newblock {\em ACM Transactions on Intelligent Systems and Technology}, 15(2):1--38, 2024.

\bibitem{luo2024understanding}
Haoyan Luo and Lucia Specia.
\newblock From understanding to utilization: A survey on explainability for large language models.
\newblock {\em arXiv preprint arXiv:2401.12874}, 2024.

\bibitem{changdan_interpretability_2024}
Chandan Singh, Jeevana~Priya Inala, Michel Galley, and Rich Caruanaand~Jianfeng Gao.
\newblock Rethinking interpretability in the era of large language models.
\newblock {\em arXiv: arXiv:2402.01761}, 2024.

\bibitem{ribeiro2016should}
Marco~Tulio Ribeiro, Sameer Singh, and Carlos Guestrin.
\newblock "why should i trust you?" explaining the predictions of any classifier.
\newblock In {\em Proceedings of the 22nd ACM SIGKDD international conference on knowledge discovery and data mining}, pages 1135--1144, 2016.

\bibitem{lundberg2017unified}
Scott~M Lundberg and Su-In Lee.
\newblock A unified approach to interpreting model predictions.
\newblock {\em Advances in neural information processing systems}, 30, 2017.

\bibitem{mohebbi2021exploring}
Hosein Mohebbi, Ali Modarressi, and Mohammad~Taher Pilehvar.
\newblock Exploring the role of bert token representations to explain sentence probing results.
\newblock {\em arXiv preprint arXiv:2104.01477}, 2021.

\bibitem{abnar2020quantifying}
Samira Abnar and Willem Zuidema.
\newblock Quantifying attention flow in transformers.
\newblock {\em arXiv preprint arXiv:2005.00928}, 2020.

\bibitem{hoover2019exbert}
Benjamin Hoover, Hendrik Strobelt, and Sebastian Gehrmann.
\newblock exbert: A visual analysis tool to explore learned representations in transformers models.
\newblock {\em arXiv preprint arXiv:1910.05276}, 2019.

\bibitem{yeh2023attentionviz}
Catherine Yeh, Yida Chen, Aoyu Wu, Cynthia Chen, Fernanda Vi{\'e}gas, and Martin Wattenberg.
\newblock Attentionviz: A global view of transformer attention.
\newblock {\em IEEE Transactions on Visualization and Computer Graphics}, 2023.

\bibitem{jin2019bert}
Di~Jin, Zhijing Jin, Joey~Tianyi Zhou, and Peter Szolovits.
\newblock Is bert really robust? natural language attack on text classification and entailment.
\newblock {\em arXiv preprint arXiv:1907.11932}, 2(10), 2019.

\bibitem{ross2020explaining}
Alexis Ross, Ana Marasovi{\'c}, and Matthew~E Peters.
\newblock Explaining nlp models via minimal contrastive editing (mice).
\newblock {\em arXiv preprint arXiv:2012.13985}, 2020.

\bibitem{petroni2019language}
Fabio Petroni, Tim Rockt{\"a}schel, Patrick Lewis, Anton Bakhtin, Yuxiang Wu, Alexander~H Miller, and Sebastian Riedel.
\newblock Language models as knowledge bases?
\newblock {\em arXiv preprint arXiv:1909.01066}, 2019.

\bibitem{li2024inference}
Kenneth Li, Oam Patel, Fernanda Vi{\'e}gas, Hanspeter Pfister, and Martin Wattenberg.
\newblock Inference-time intervention: Eliciting truthful answers from a language model.
\newblock {\em Advances in Neural Information Processing Systems}, 36, 2024.

\bibitem{bills2023language}
Steven Bills, Nick Cammarata, Dan Mossing, Henk Tillman, Leo Gao, Gabriel Goh, Ilya Sutskever, Jan Leike, Jeff Wu, and William Saunders.
\newblock Language models can explain neurons in language models.
\newblock \url{https://openaipublic.blob.core.windows.net/neuron-explainer/paper/index.html}, 2023.

\bibitem{antverg2021pitfalls}
Omer Antverg and Yonatan Belinkov.
\newblock On the pitfalls of analyzing individual neurons in language models.
\newblock {\em arXiv preprint arXiv:2110.07483}, 2021.

\bibitem{chughtai2023toy}
Bilal Chughtai, Lawrence Chan, and Neel Nanda.
\newblock A toy model of universality: Reverse engineering how networks learn group operations.
\newblock In {\em International Conference on Machine Learning}, pages 6243--6267. PMLR, 2023.

\bibitem{meng2022locating}
Kevin Meng, David Bau, Alex Andonian, and Yonatan Belinkov.
\newblock Locating and editing factual associations in gpt.
\newblock {\em Advances in Neural Information Processing Systems}, 35:17359--17372, 2022.

\bibitem{li2024machines}
Jiliang Li, Yifan Zhang, Zachary Karas, Collin McMillan, Kevin Leach, and Yu~Huang.
\newblock Do machines and humans focus on similar code? exploring explainability of large language models in code summarization.
\newblock In {\em Proceedings of the 32nd IEEE/ACM International Conference on Program Comprehension}, pages 47--51, 2024.

\bibitem{zhao2023explaining}
Ruochen Zhao, Shafiq Joty, Yongjie Wang, and Tan Wang.
\newblock Explaining language models' predictions with high-impact concepts.
\newblock {\em arXiv preprint arXiv:2305.02160}, 2023.

\bibitem{Doshi_interpretability}
F.~Doshi-Velez and B.~Kim.
\newblock Towards a rigorous science of interpretable machine learning.
\newblock {\em arXiv preprint arXiv:1702.08608}, 2017.

\bibitem{Lipton_interpretability}
Zachary~C. Lipton.
\newblock The mythos of model interpretability: In machine learning, the concept of interpretability is both important and slippery.
\newblock {\em Queue}, 16(3):31–57, 2018.

\bibitem{arenas2021foundations}
Marcelo Arenas, Daniel Baez, Pablo Barcel{\'o}, Jorge P{\'e}rez, and Bernardo Subercaseaux.
\newblock Foundations of symbolic languages for model interpretability.
\newblock {\em Advances in neural information processing systems}, 34:11690--11701, 2021.

\bibitem{rajendran2024learning}
Goutham Rajendran, Simon Buchholz, Bryon Aragam, Bernhard Sch{\"o}lkopf, and Pradeep Ravikumar.
\newblock Learning interpretable concepts: Unifying causal representation learning and foundation models.
\newblock {\em arXiv preprint arXiv:2402.09236}, 2024.

\bibitem{cunnington2024role}
Daniel Cunnington, Mark Law, Jorge Lobo, and Alessandra Russo.
\newblock The role of foundation models in neuro-symbolic learning and reasoning.
\newblock {\em arXiv preprint arXiv:2402.01889}, 2024.

\bibitem{carvalho2019machine}
Diogo~V Carvalho, Eduardo~M Pereira, and Jaime~S Cardoso.
\newblock Machine learning interpretability: A survey on methods and metrics.
\newblock {\em Electronics}, 8(8):832, 2019.

\bibitem{herman2017promise}
Bernease Herman.
\newblock The promise and peril of human evaluation for model interpretability.
\newblock {\em arXiv preprint arXiv:1711.07414}, 2017.

\bibitem{dasgupta2022framework}
Sanjoy Dasgupta, Nave Frost, and Michal Moshkovitz.
\newblock Framework for evaluating faithfulness of local explanations.
\newblock In {\em International Conference on Machine Learning}, pages 4794--4815. PMLR, 2022.

\bibitem{jesus2021can}
S{\'e}rgio Jesus, Catarina Bel{\'e}m, Vladimir Balayan, Jo{\~a}o Bento, Pedro Saleiro, Pedro Bizarro, and Jo{\~a}o Gama.
\newblock How can i choose an explainer? an application-grounded evaluation of post-hoc explanations.
\newblock In {\em Proceedings of the 2021 ACM conference on fairness, accountability, and transparency}, pages 805--815, 2021.

\bibitem{poursabzi2021manipulating}
Forough Poursabzi-Sangdeh, Daniel~G Goldstein, Jake~M Hofman, Jennifer~Wortman Wortman~Vaughan, and Hanna Wallach.
\newblock Manipulating and measuring model interpretability.
\newblock In {\em Proceedings of the 2021 CHI conference on human factors in computing systems}, pages 1--52, 2021.

\bibitem{agarwal2022rethinking}
Chirag Agarwal, Nari Johnson, Martin Pawelczyk, Satyapriya Krishna, Eshika Saxena, Marinka Zitnik, and Himabindu Lakkaraju.
\newblock Rethinking stability for attribution-based explanations.
\newblock {\em arXiv preprint arXiv:2203.06877}, 2022.

\bibitem{vig-2019-multiscale}
Jesse Vig.
\newblock A multiscale visualization of attention in the transformer model.
\newblock In {\em Proceedings of the 57th Annual Meeting of the Association for Computational Linguistics: System Demonstrations}, 2019.

\bibitem{jain2019attention}
Sarthak Jain and Byron~C Wallace.
\newblock Attention is not explanation.
\newblock {\em arXiv preprint arXiv:1902.10186}, 2019.

\bibitem{yang2021fast}
Jilei Yang.
\newblock Fast treeshap: Accelerating shap value computation for trees.
\newblock {\em arXiv preprint arXiv:2109.09847}, 2021.

\bibitem{vstrumbelj2014explaining}
Erik {\v{S}}trumbelj and Igor Kononenko.
\newblock Explaining prediction models and individual predictions with feature contributions.
\newblock {\em Knowledge and information systems}, 41:647--665, 2014.

\bibitem{covert2020understanding}
Ian Covert, Scott~M Lundberg, and Su-In Lee.
\newblock Understanding global feature contributions with additive importance measures.
\newblock {\em Advances in Neural Information Processing Systems}, 33:17212--17223, 2020.

\bibitem{lundberg2020local}
Scott~M Lundberg, Gabriel Erion, Hugh Chen, Alex DeGrave, Jordan~M Prutkin, Bala Nair, Ronit Katz, Jonathan Himmelfarb, Nisha Bansal, and Su-In Lee.
\newblock From local explanations to global understanding with explainable ai for trees.
\newblock {\em Nature machine intelligence}, 2(1):56--67, 2020.

\bibitem{shrikumar2017learning}
Avanti Shrikumar, Peyton Greenside, and Anshul Kundaje.
\newblock Learning important features through propagating activation differences.
\newblock In {\em International conference on machine learning}, pages 3145--3153. PMlR, 2017.

\bibitem{chen2018shapley}
Jianbo Chen, Le~Song, Martin~J Wainwright, and Michael~I Jordan.
\newblock L-shapley and c-shapley: Efficient model interpretation for structured data.
\newblock {\em arXiv preprint arXiv:1808.02610}, 2018.

\bibitem{ancona2019explaining}
Marco Ancona, Cengiz Oztireli, and Markus Gross.
\newblock Explaining deep neural networks with a polynomial time algorithm for shapley value approximation.
\newblock In {\em International Conference on Machine Learning}, pages 272--281. PMLR, 2019.

\bibitem{aas2021explaining}
Kjersti Aas, Martin Jullum, and Anders L{\o}land.
\newblock Explaining individual predictions when features are dependent: More accurate approximations to shapley values.
\newblock {\em Artificial Intelligence}, 298:103502, 2021.

\bibitem{janzing2020feature}
Dominik Janzing, Lenon Minorics, and Patrick Bl{\"o}baum.
\newblock Feature relevance quantification in explainable ai: A causal problem.
\newblock In {\em International Conference on artificial intelligence and statistics}, pages 2907--2916. PMLR, 2020.

\bibitem{covert2021explaining}
Ian Covert, Scott Lundberg, and Su-In Lee.
\newblock Explaining by removing: A unified framework for model explanation.
\newblock {\em Journal of Machine Learning Research}, 22(209):1--90, 2021.

\bibitem{suh2024survey}
Namjoon Suh and Guang Cheng.
\newblock A survey on statistical theory of deep learning: Approximation, training dynamics, and generative models.
\newblock {\em arXiv preprint arXiv:2401.07187}, 2024.

\bibitem{sun2019optimization}
Ruoyu Sun.
\newblock Optimization for deep learning: Theory and algorithms.
\newblock {\em arXiv preprint arXiv:1912.08957}, 2019.

\bibitem{bartlett2021deep}
Peter~L Bartlett, Andrea Montanari, and Alexander Rakhlin.
\newblock Deep learning: a statistical viewpoint.
\newblock {\em Acta numerica}, 30:87--201, 2021.

\bibitem{shalev2014understanding}
Shai Shalev-Shwartz and Shai Ben-David.
\newblock {\em Understanding machine learning: From theory to algorithms}.
\newblock Cambridge university press, 2014.

\bibitem{blumer1989learnability}
Anselm Blumer, Andrzej Ehrenfeucht, David Haussler, and Manfred~K Warmuth.
\newblock Learnability and the {Vapnik-Chervonenkis} dimension.
\newblock {\em Journal of the ACM}, 36(4):929--965, 1989.

\bibitem{cherkassky1999model}
Vladimir Cherkassky, Xuhui Shao, Filip~M Mulier, and Vladimir~N Vapnik.
\newblock Model complexity control for regression using vc generalization bounds.
\newblock {\em IEEE transactions on Neural Networks}, 10(5):1075--1089, 1999.

\bibitem{vapnik2006estimation}
Vladimir Vapnik.
\newblock {\em Estimation of Dependences based on Empirical Data}.
\newblock Springer Science \& Business Media, 2006.

\bibitem{mohri2018foundations}
Mehryar Mohri, Afshin Rostamizadeh, and Ameet Talwalkar.
\newblock {\em Foundations of machine learning}.
\newblock MIT press, 2018.

\bibitem{tu2020understanding}
Zhuozhuo Tu, Fengxiang He, and Dacheng Tao.
\newblock Understanding generalization in recurrent neural networks.
\newblock In {\em International Conference on Learning Representations}, 2020.

\bibitem{Neyshabur_2017}
Behnam Neyshabur.
\newblock Implicit regularization in deep learning.
\newblock {\em Cornell University - arXiv,Cornell University - arXiv}, Sep 2017.

\bibitem{zhang2017understanding}
Chiyuan Zhang, Samy Bengio, Moritz Hardt, Benjamin Recht, and Oriol Vinyals.
\newblock Understanding deep learning requires rethinking generalization.
\newblock In {\em International Conference on Learning Representations}, 2017.

\bibitem{dudley2010universal}
Richard~M Dudley.
\newblock Universal donsker classes and metric entropy.
\newblock In {\em Selected Works of RM Dudley}, pages 345--365. Springer, 2010.

\bibitem{bousquet2002stability}
Olivier Bousquet and Andr{\'e} Elisseeff.
\newblock Stability and generalization.
\newblock {\em Journal of Machine Learning Research}, 2(Mar):499--526, 2002.

\bibitem{feldman2018generalization}
Vitaly Feldman and Jan Vondrak.
\newblock Generalization bounds for uniformly stable algorithms.
\newblock {\em Advances in Neural Information Processing Systems}, 31, 2018.

\bibitem{feldman2019high}
Vitaly Feldman and Jan Vondrak.
\newblock High probability generalization bounds for uniformly stable algorithms with nearly optimal rate.
\newblock In {\em Conference on Learning Theory}, pages 1270--1279. PMLR, 2019.

\bibitem{bousquet2020sharper}
Olivier Bousquet, Yegor Klochkov, and Nikita Zhivotovskiy.
\newblock Sharper bounds for uniformly stable algorithms.
\newblock In {\em Conference on Learning Theory}, pages 610--626. PMLR, 2020.

\bibitem{fu2022sharper}
Shi Fu, Yunwen Lei, Qiong Cao, Xinmei Tian, and Dacheng Tao.
\newblock Sharper bounds for uniformly stable algorithms with stationary mixing process.
\newblock In {\em The Eleventh International Conference on Learning Representations}, 2022.

\bibitem{ren2001best}
Yao-Feng Ren and Han-Ying Liang.
\newblock On the best constant in marcinkiewicz--zygmund inequality.
\newblock {\em Statistics \& probability letters}, 53(3):227--233, 2001.

\bibitem{mcallester1999pac}
David~A McAllester.
\newblock Pac-bayesian model averaging.
\newblock In {\em Proceedings of the twelfth annual conference on Computational learning theory}, pages 164--170, 1999.

\bibitem{brown2020language}
Tom Brown, Benjamin Mann, Nick Ryder, Melanie Subbiah, Jared~D Kaplan, Prafulla Dhariwal, Arvind Neelakantan, Pranav Shyam, Girish Sastry, Amanda Askell, et~al.
\newblock Language models are few-shot learners.
\newblock {\em Advances in neural information processing systems}, 33:1877--1901, 2020.

\bibitem{liu2023trustworthy}
Yang Liu, Yuanshun Yao, Jean-Francois Ton, Xiaoying Zhang, Ruocheng Guo~Hao Cheng, Yegor Klochkov, Muhammad~Faaiz Taufiq, and Hang Li.
\newblock Trustworthy llms: a survey and guideline for evaluating large language models' alignment.
\newblock {\em arXiv preprint arXiv:2308.05374}, 2023.

\bibitem{liu2024towards}
Shang Liu, Zhongze Cai, Guanting Chen, and Xiaocheng Li.
\newblock Towards better understanding of in-context learning ability from in-context uncertainty quantification.
\newblock {\em arXiv preprint arXiv:2405.15115}, 2024.

\bibitem{lintransformers}
Licong Lin, Yu~Bai, and Song Mei.
\newblock Transformers as decision makers: Provable in-context reinforcement learning via supervised pretraining.
\newblock {\em arXiv preprint arXiv:2310.08566}, 2023.

\bibitem{li2023transformers}
Yuchen Li, Yuanzhi Li, and Andrej Risteski.
\newblock How do transformers learn topic structure: Towards a mechanistic understanding.
\newblock In {\em International Conference on Machine Learning}, pages 19689--19729. PMLR, 2023.

\bibitem{zhang2023and}
Yufeng Zhang, Fengzhuo Zhang, Zhuoran Yang, and Zhaoran Wang.
\newblock What and how does in-context learning learn? bayesian model averaging, parameterization, and generalization.
\newblock {\em arXiv preprint arXiv:2305.19420}, 2023.

\bibitem{li2024nonlinear}
Hongkang Li, Meng Wang, Songtao Lu, Xiaodong Cui, and Pin-Yu Chen.
\newblock How do nonlinear transformers acquire generalization-guaranteed cot ability?
\newblock In {\em High-dimensional Learning Dynamics 2024: The Emergence of Structure and Reasoning}, 2024.

\bibitem{wies2024learnability}
Noam Wies, Yoav Levine, and Amnon Shashua.
\newblock The learnability of in-context learning.
\newblock {\em Advances in Neural Information Processing Systems}, 36, 2024.

\bibitem{wu2023many}
Jingfeng Wu, Difan Zou, Zixiang Chen, Vladimir Braverman, Quanquan Gu, and Peter~L Bartlett.
\newblock How many pretraining tasks are needed for in-context learning of linear regression?
\newblock {\em arXiv preprint arXiv:2310.08391}, 2023.

\bibitem{bai2024transformers}
Yu~Bai, Fan Chen, Huan Wang, Caiming Xiong, and Song Mei.
\newblock Transformers as statisticians: Provable in-context learning with in-context algorithm selection.
\newblock {\em Advances in neural information processing systems}, 36, 2024.

\bibitem{jeoninformation}
Hong~Jun Jeon, Jason~D Lee, Qi~Lei, and Benjamin Van~Roy.
\newblock An information-theoretic analysis of in-context learning.
\newblock {\em arXiv preprint arXiv:2401.15530}, 2024.

\bibitem{deoraoptimization}
Puneesh Deora, Rouzbeh Ghaderi, Hossein Taheri, and Christos Thrampoulidis.
\newblock On the optimization and generalization of multi-head attention.
\newblock {\em Transactions on Machine Learning Research}, 2024.

\bibitem{devlin2018bert}
Jacob Devlin, Ming-Wei Chang, Kenton Lee, and Kristina Toutanova.
\newblock Bert: Pre-training of deep bidirectional transformers for language understanding.
\newblock {\em arXiv preprint arXiv:1810.04805}, 2018.

\bibitem{dong2022survey}
Qingxiu Dong, Lei Li, Damai Dai, Ce~Zheng, Zhiyong Wu, Baobao Chang, Xu~Sun, Jingjing Xu, and Zhifang Sui.
\newblock A survey on in-context learning.
\newblock {\em arXiv preprint arXiv:2301.00234}, 2022.

\bibitem{akyurek2022learning}
Ekin Aky{\"u}rek, Dale Schuurmans, Jacob Andreas, Tengyu Ma, and Denny Zhou.
\newblock What learning algorithm is in-context learning? investigations with linear models.
\newblock In {\em The Eleventh International Conference on Learning Representations}, 2022.

\bibitem{von2023transformers}
Johannes Von~Oswald, Eyvind Niklasson, Ettore Randazzo, Jo{\~a}o Sacramento, Alexander Mordvintsev, Andrey Zhmoginov, and Max Vladymyrov.
\newblock Transformers learn in-context by gradient descent.
\newblock In {\em International Conference on Machine Learning}, pages 35151--35174. PMLR, 2023.

\bibitem{dai2022can}
Damai Dai, Yutao Sun, Li~Dong, Yaru Hao, Shuming Ma, Zhifang Sui, and Furu Wei.
\newblock Why can gpt learn in-context? language models implicitly perform gradient descent as meta-optimizers.
\newblock {\em arXiv preprint arXiv:2212.10559}, 2022.

\bibitem{giannou2023looped}
Angeliki Giannou, Shashank Rajput, Jy-yong Sohn, Kangwook Lee, Jason~D Lee, and Dimitris Papailiopoulos.
\newblock Looped transformers as programmable computers.
\newblock In {\em International Conference on Machine Learning}, pages 11398--11442. PMLR, 2023.

\bibitem{li2023closeness}
Shuai Li, Zhao Song, Yu~Xia, Tong Yu, and Tianyi Zhou.
\newblock The closeness of in-context learning and weight shifting for softmax regression.
\newblock {\em arXiv preprint arXiv:2304.13276}, 2023.

\bibitem{ahn2024transformers}
Kwangjun Ahn, Xiang Cheng, Hadi Daneshmand, and Suvrit Sra.
\newblock Transformers learn to implement preconditioned gradient descent for in-context learning.
\newblock {\em Advances in Neural Information Processing Systems}, 36, 2024.

\bibitem{zhang2024trained}
Ruiqi Zhang, Spencer Frei, and Peter~L Bartlett.
\newblock Trained transformers learn linear models in-context.
\newblock {\em Journal of Machine Learning Research}, 25(49):1--55, 2024.

\bibitem{mahankali2023one}
Arvind Mahankali, Tatsunori~B Hashimoto, and Tengyu Ma.
\newblock One step of gradient descent is provably the optimal in-context learner with one layer of linear self-attention.
\newblock {\em arXiv preprint arXiv:2307.03576}, 2023.

\bibitem{li2024dissecting}
Yingcong Li, Kartik Sreenivasan, Angeliki Giannou, Dimitris Papailiopoulos, and Samet Oymak.
\newblock Dissecting chain-of-thought: Compositionality through in-context filtering and learning.
\newblock {\em Advances in Neural Information Processing Systems}, 36, 2024.

\bibitem{feng2024towards}
Guhao Feng, Bohang Zhang, Yuntian Gu, Haotian Ye, Di~He, and Liwei Wang.
\newblock Towards revealing the mystery behind chain of thought: a theoretical perspective.
\newblock {\em Advances in Neural Information Processing Systems}, 36, 2024.

\bibitem{lichain}
Zhiyuan Li, Hong Liu, Denny Zhou, and Tengyu Ma.
\newblock Chain of thought empowers transformers to solve inherently serial problems.
\newblock In {\em The Twelfth International Conference on Learning Representations}, 2024.

\bibitem{merrill2023expressive}
William Merrill and Ashish Sabharwal.
\newblock The expressive power of transformers with chain of thought, 2023.

\bibitem{zhou2022domain}
Kaiyang Zhou, Ziwei Liu, Yu~Qiao, Tao Xiang, and Chen~Change Loy.
\newblock Domain generalization: A survey.
\newblock {\em IEEE Transactions on Pattern Analysis and Machine Intelligence}, 2022.

\bibitem{huangcontext}
Yu~Huang, Yuan Cheng, and Yingbin Liang.
\newblock In-context convergence of transformers.
\newblock In {\em Forty-first International Conference on Machine Learning}, 2024.

\bibitem{garg2022can}
Shivam Garg, Dimitris Tsipras, Percy~S Liang, and Gregory Valiant.
\newblock What can transformers learn in-context? a case study of simple function classes.
\newblock {\em Advances in Neural Information Processing Systems}, 35:30583--30598, 2022.

\bibitem{sahoo2024systematic}
Pranab Sahoo, Ayush~Kumar Singh, Sriparna Saha, Vinija Jain, Samrat Mondal, and Aman Chadha.
\newblock A systematic survey of prompt engineering in large language models: Techniques and applications.
\newblock {\em arXiv preprint arXiv:2402.07927}, 2024.

\bibitem{zhangautomatic}
Zhuosheng Zhang, Aston Zhang, Mu~Li, and Alex Smola.
\newblock Automatic chain of thought prompting in large language models.
\newblock In {\em The Eleventh International Conference on Learning Representations}, 2024.

\bibitem{yao2024tree}
Shunyu Yao, Dian Yu, Jeffrey Zhao, Izhak Shafran, Tom Griffiths, Yuan Cao, and Karthik Narasimhan.
\newblock Tree of thoughts: Deliberate problem solving with large language models.
\newblock {\em Advances in Neural Information Processing Systems}, 36, 2024.

\bibitem{yao2023beyond}
Yao Yao, Zuchao Li, and Hai Zhao.
\newblock Beyond chain-of-thought, effective graph-of-thought reasoning in language models.
\newblock {\em arXiv preprint arXiv:2305.16582}, 2023.

\bibitem{xu2023search}
Shicheng Xu, Liang Pang, Huawei Shen, Xueqi Cheng, and Tat-seng Chua.
\newblock Search-in-the-chain: Towards the accurate, credible and traceable content generation for complex knowledge-intensive tasks.
\newblock {\em arXiv preprint arXiv:2304.14732}, 2023.

\bibitem{shi2023replug}
Weijia Shi, Sewon Min, Michihiro Yasunaga, Minjoon Seo, Rich James, Mike Lewis, Luke Zettlemoyer, and Wen-tau Yih.
\newblock Replug: Retrieval-augmented black-box language models.
\newblock {\em arXiv preprint arXiv:2301.12652}, 2023.

\bibitem{asai2023self}
Akari Asai, Zeqiu Wu, Yizhong Wang, Avirup Sil, and Hannaneh Hajishirzi.
\newblock Self-rag: Learning to retrieve, generate, and critique through self-reflection.
\newblock {\em arXiv preprint arXiv:2310.11511}, 2023.

\bibitem{ram2023context}
Ori Ram, Yoav Levine, Itay Dalmedigos, Dor Muhlgay, Amnon Shashua, Kevin Leyton-Brown, and Yoav Shoham.
\newblock In-context retrieval-augmented language models.
\newblock {\em arXiv preprint arXiv:2302.00083}, 2023.

\bibitem{srivastava2022beyond}
Aarohi Srivastava, Abhinav Rastogi, Abhishek Rao, Abu Awal~Md Shoeb, Abubakar Abid, Adam Fisch, Adam~R Brown, Adam Santoro, Aditya Gupta, Adri{\`a} Garriga-Alonso, et~al.
\newblock Beyond the imitation game: Quantifying and extrapolating the capabilities of language models.
\newblock {\em arXiv preprint arXiv:2206.04615}, 2022.

\bibitem{arora2023theory}
Sanjeev Arora and Anirudh Goyal.
\newblock A theory for emergence of complex skills in language models.
\newblock {\em arXiv preprint arXiv:2307.15936}, 2023.

\bibitem{rosas2020reconciling}
Fernando~E Rosas, Pedro~AM Mediano, Henrik~J Jensen, Anil~K Seth, Adam~B Barrett, Robin~L Carhart-Harris, and Daniel Bor.
\newblock Reconciling emergences: An information-theoretic approach to identify causal emergence in multivariate data.
\newblock {\em PLoS computational biology}, 16(12):e1008289, 2020.

\bibitem{chen2024quantifying}
Hang Chen, Xinyu Yang, Jiaying Zhu, and Wenya Wang.
\newblock Quantifying emergence in large language models.
\newblock {\em arXiv preprint arXiv:2405.12617}, 2024.

\bibitem{li2023transformershow}
Yuchen Li, Yuanzhi Li, and Andrej Risteski.
\newblock How do transformers learn topic structure: Towards a mechanistic understanding.
\newblock In {\em International Conference on Machine Learning}, pages 19689--19729. PMLR, 2023.

\bibitem{tarzanagh2023transformers}
Davoud~Ataee Tarzanagh, Yingcong Li, Christos Thrampoulidis, and Samet Oymak.
\newblock Transformers as support vector machines.
\newblock {\em arXiv preprint arXiv:2308.16898}, 2023.

\bibitem{ataee2023max}
Davoud Ataee~Tarzanagh, Yingcong Li, Xuechen Zhang, and Samet Oymak.
\newblock Max-margin token selection in attention mechanism.
\newblock {\em Advances in Neural Information Processing Systems}, 36:48314--48362, 2023.

\bibitem{nichani2024transformers}
Eshaan Nichani, Alex Damian, and Jason~D Lee.
\newblock How transformers learn causal structure with gradient descent.
\newblock {\em arXiv preprint arXiv:2402.14735}, 2024.

\bibitem{jelassi2022vision}
Samy Jelassi, Michael Sander, and Yuanzhi Li.
\newblock Vision transformers provably learn spatial structure.
\newblock {\em Advances in Neural Information Processing Systems}, 35:37822--37836, 2022.

\bibitem{tian2024scan}
Yuandong Tian, Yiping Wang, Beidi Chen, and Simon~S Du.
\newblock Scan and snap: Understanding training dynamics and token composition in 1-layer transformer.
\newblock {\em Advances in Neural Information Processing Systems}, 36, 2024.

\bibitem{tian2023joma}
Yuandong Tian, Yiping Wang, Zhenyu Zhang, Beidi Chen, and Simon Du.
\newblock Joma: Demystifying multilayer transformers via joint dynamics of mlp and attention.
\newblock {\em arXiv preprint arXiv:2310.00535}, 2023.

\bibitem{abbe2024transformers}
Emmanuel Abbe, Samy Bengio, Enric Boix-Adsera, Etai Littwin, and Joshua Susskind.
\newblock Transformers learn through gradual rank increase.
\newblock {\em Advances in Neural Information Processing Systems}, 36, 2024.

\bibitem{openai2023gpt}
R~OpenAI.
\newblock Gpt-4 technical report.
\newblock {\em arXiv}, pages 2303--08774, 2023.

\bibitem{ouyang2022training}
Long Ouyang, Jeffrey Wu, Xu~Jiang, Diogo Almeida, Carroll Wainwright, Pamela Mishkin, Chong Zhang, Sandhini Agarwal, Katarina Slama, Alex Ray, et~al.
\newblock Training language models to follow instructions with human feedback.
\newblock {\em Advances in neural information processing systems}, 35:27730--27744, 2022.

\bibitem{bai2022training}
Yuntao Bai, Andy Jones, Kamal Ndousse, Amanda Askell, Anna Chen, Nova DasSarma, Dawn Drain, Stanislav Fort, Deep Ganguli, Tom Henighan, et~al.
\newblock Training a helpful and harmless assistant with reinforcement learning from human feedback.
\newblock {\em arXiv preprint arXiv:2204.05862}, 2022.

\bibitem{munos2023nash}
R{\'e}mi Munos, Michal Valko, Daniele Calandriello, Mohammad~Gheshlaghi Azar, Mark Rowland, Zhaohan~Daniel Guo, Yunhao Tang, Matthieu Geist, Thomas Mesnard, Andrea Michi, et~al.
\newblock Nash learning from human feedback.
\newblock {\em arXiv preprint arXiv:2312.00886}, 2023.

\bibitem{rafailov2024direct}
Rafael Rafailov, Archit Sharma, Eric Mitchell, Christopher~D Manning, Stefano Ermon, and Chelsea Finn.
\newblock Direct preference optimization: Your language model is secretly a reward model.
\newblock {\em Advances in Neural Information Processing Systems}, 36, 2024.

\bibitem{xiong2024iterative}
Wei Xiong, Hanze Dong, Chenlu Ye, Ziqi Wang, Han Zhong, Heng Ji, Nan Jiang, and Tong Zhang.
\newblock Iterative preference learning from human feedback: Bridging theory and practice for rlhf under kl-constraint.
\newblock In {\em Forty-first International Conference on Machine Learning}, 2024.

\bibitem{yue2012k}
Yisong Yue, Josef Broder, Robert Kleinberg, and Thorsten Joachims.
\newblock The k-armed dueling bandits problem.
\newblock {\em Journal of Computer and System Sciences}, 78(5):1538--1556, 2012.

\bibitem{saha2021optimal}
Aadirupa Saha.
\newblock Optimal algorithms for stochastic contextual preference bandits.
\newblock {\em Advances in Neural Information Processing Systems}, 34:30050--30062, 2021.

\bibitem{novoseller2020dueling}
Ellen Novoseller, Yibing Wei, Yanan Sui, Yisong Yue, and Joel Burdick.
\newblock Dueling posterior sampling for preference-based reinforcement learning.
\newblock In {\em Conference on Uncertainty in Artificial Intelligence}, pages 1029--1038. PMLR, 2020.

\bibitem{saha2023dueling}
Aadirupa Saha, Aldo Pacchiano, and Jonathan Lee.
\newblock Dueling rl: Reinforcement learning with trajectory preferences.
\newblock In {\em International Conference on Artificial Intelligence and Statistics}, pages 6263--6289. PMLR, 2023.

\bibitem{chen2022human}
Xiaoyu Chen, Han Zhong, Zhuoran Yang, Zhaoran Wang, and Liwei Wang.
\newblock Human-in-the-loop: Provably efficient preference-based reinforcement learning with general function approximation.
\newblock In {\em International Conference on Machine Learning}, pages 3773--3793. PMLR, 2022.

\bibitem{wang2023rlhf}
Yuanhao Wang, Qinghua Liu, and Chi Jin.
\newblock Is rlhf more difficult than standard rl?
\newblock {\em arXiv preprint arXiv:2306.14111}, 2023.

\bibitem{zhan2023query}
Wenhao Zhan, Masatoshi Uehara, Wen Sun, and Jason~D Lee.
\newblock How to query human feedback efficiently in rl?
\newblock In {\em ICML 2023 Workshop The Many Facets of Preference-Based Learning}, 2023.

\bibitem{zhu2023principled}
Banghua Zhu, Michael Jordan, and Jiantao Jiao.
\newblock Principled reinforcement learning with human feedback from pairwise or k-wise comparisons.
\newblock In {\em International Conference on Machine Learning}, pages 43037--43067. PMLR, 2023.

\bibitem{zhan2023provable}
Wenhao Zhan, Masatoshi Uehara, Nathan Kallus, Jason~D Lee, and Wen Sun.
\newblock Provable offline reinforcement learning with human feedback.
\newblock In {\em ICML 2023 Workshop The Many Facets of Preference-Based Learning}, 2023.

\bibitem{li2023reinforcement}
Zihao Li, Zhuoran Yang, and Mengdi Wang.
\newblock Reinforcement learning with human feedback: Learning dynamic choices via pessimism.
\newblock {\em arXiv preprint arXiv:2305.18438}, 2023.

\bibitem{sekhari2023contextual}
Ayush Sekhari, Karthik Sridharan, Wen Sun, and Runzhe Wu.
\newblock Contextual bandits and imitation learning via preference-based active queries.
\newblock {\em arXiv preprint arXiv:2307.12926}, 2023.

\bibitem{ji2024reinforcement}
Kaixuan Ji, Jiafan He, and Quanquan Gu.
\newblock Reinforcement learning from human feedback with active queries.
\newblock {\em arXiv preprint arXiv:2402.09401}, 2024.

\bibitem{villalobos2022will}
Pablo Villalobos, Jaime Sevilla, Lennart Heim, Tamay Besiroglu, Marius Hobbhahn, and Anson Ho.
\newblock Will we run out of data? an analysis of the limits of scaling datasets in machine learning.
\newblock {\em arXiv preprint arXiv:2211.04325}, 2022.

\bibitem{sadasivan2023can}
Vinu~Sankar Sadasivan, Aounon Kumar, Sriram Balasubramanian, Wenxiao Wang, and Soheil Feizi.
\newblock Can ai-generated text be reliably detected?
\newblock {\em arXiv preprint arXiv:2303.11156}, 2023.

\bibitem{huschens2023you}
Martin Huschens, Martin Briesch, Dominik Sobania, and Franz Rothlauf.
\newblock Do you trust chatgpt?--perceived credibility of human and ai-generated content.
\newblock {\em arXiv preprint arXiv:2309.02524}, 2023.

\bibitem{schuhmann2022laion}
Christoph Schuhmann, Romain Beaumont, Richard Vencu, Cade Gordon, Ross Wightman, Mehdi Cherti, Theo Coombes, Aarush Katta, Clayton Mullis, Mitchell Wortsman, et~al.
\newblock Laion-5b: An open large-scale dataset for training next generation image-text models.
\newblock {\em Advances in Neural Information Processing Systems}, 35:25278--25294, 2022.

\bibitem{alemohammadself}
Sina Alemohammad, Josue Casco-Rodriguez, Lorenzo Luzi, Ahmed~Imtiaz Humayun, Hossein Babaei, Daniel LeJeune, Ali Siahkoohi, and Richard Baraniuk.
\newblock Self-consuming generative models go mad.
\newblock In {\em The Twelfth International Conference on Learning Representations}, 2024.

\bibitem{shumailov2024ai}
Ilia Shumailov, Zakhar Shumaylov, Yiren Zhao, Nicolas Papernot, Ross Anderson, and Yarin Gal.
\newblock Ai models collapse when trained on recursively generated data.
\newblock {\em Nature}, 631(8022):755--759, 2024.

\bibitem{bertrandstability}
Quentin Bertrand, Joey Bose, Alexandre Duplessis, Marco Jiralerspong, and Gauthier Gidel.
\newblock On the stability of iterative retraining of generative models on their own data.
\newblock In {\em The Twelfth International Conference on Learning Representations}, 2024.

\bibitem{huang2022large}
Jiaxin Huang, Shixiang~Shane Gu, Le~Hou, Yuexin Wu, Xuezhi Wang, Hongkun Yu, and Jiawei Han.
\newblock Large language models can self-improve.
\newblock {\em arXiv preprint arXiv:2210.11610}, 2022.

\bibitem{martinez2023towards}
Gonzalo Mart{\'\i}nez, Lauren Watson, Pedro Reviriego, Jos{\'e}~Alberto Hern{\'a}ndez, Marc Juarez, and Rik Sarkar.
\newblock Towards understanding the interplay of generative artificial intelligence and the internet.
\newblock In {\em International Workshop on Epistemic Uncertainty in Artificial Intelligence}, pages 59--73. Springer, 2023.

\bibitem{briesch2023large}
Martin Briesch, Dominik Sobania, and Franz Rothlauf.
\newblock Large language models suffer from their own output: An analysis of the self-consuming training loop.
\newblock {\em arXiv preprint arXiv:2311.16822}, 2023.

\bibitem{wyllie2024fairness}
Sierra Wyllie, Ilia Shumailov, and Nicolas Papernot.
\newblock Fairness feedback loops: training on synthetic data amplifies bias.
\newblock In {\em The 2024 ACM Conference on Fairness, Accountability, and Transparency}, pages 2113--2147, 2024.

\bibitem{dohmatobtale}
Elvis Dohmatob, Yunzhen Feng, Pu~Yang, Francois Charton, and Julia Kempe.
\newblock A tale of tails: Model collapse as a change of scaling laws.
\newblock In {\em Forty-first International Conference on Machine Learning}, 2024.

\bibitem{gerstgrasser2024is}
Matthias Gerstgrasser, Rylan Schaeffer, Apratim Dey, Rafael Rafailov, Tomasz Korbak, Henry Sleight, Rajashree Agrawal, John Hughes, Dhruv~Bhandarkar Pai, Andrey Gromov, Dan Roberts, Diyi Yang, David~L. Donoho, and Sanmi Koyejo.
\newblock Is model collapse inevitable? breaking the curse of recursion by accumulating real and synthetic data.
\newblock In {\em First Conference on Language Modeling}, 2024.

\bibitem{dohmatob2024model}
Elvis Dohmatob, Yunzhen Feng, and Julia Kempe.
\newblock Model collapse demystified: The case of regression.
\newblock {\em arXiv preprint arXiv:2402.07712}, 2024.

\bibitem{gillmanself}
Nate Gillman, Michael Freeman, Daksh Aggarwal, HSU Chia-Hong, Calvin Luo, Yonglong Tian, and Chen Sun.
\newblock Self-correcting self-consuming loops for generative model training.
\newblock In {\em Forty-first International Conference on Machine Learning}, 2024.

\bibitem{alemohammad2024self}
Sina Alemohammad, Ahmed~Imtiaz Humayun, Shruti Agarwal, John Collomosse, and Richard Baraniuk.
\newblock Self-improving diffusion models with synthetic data.
\newblock {\em arXiv preprint arXiv:2408.16333}, 2024.

\bibitem{villani2009optimal}
C{\'e}dric Villani et~al.
\newblock {\em Optimal transport: old and new}, volume 338.
\newblock Springer, 2009.

\bibitem{futowards}
Shi Fu, Sen Zhang, Yingjie Wang, Xinmei Tian, and Dacheng Tao.
\newblock Towards theoretical understandings of self-consuming generative models.
\newblock In {\em Forty-first International Conference on Machine Learning}, 2024.

\bibitem{van2014probability}
Ramon Van~Handel.
\newblock Probability in high dimension.
\newblock {\em Lecture Notes (Princeton University)}, 2(3):2--3, 2014.

\bibitem{zuboff2019age}
Shoshana Zuboff.
\newblock The age of surveillance capitalism: The fight for a human future at the new frontier of power, edn.
\newblock {\em PublicAffairs, New York}, 2019.

\bibitem{ishihara2023training}
Shotaro Ishihara.
\newblock Training data extraction from pre-trained language models: A survey.
\newblock {\em arXiv preprint arXiv:2305.16157}, 2023.

\bibitem{alkhamissi2022review}
Badr AlKhamissi, Millicent Li, Asli Celikyilmaz, Mona Diab, and Marjan Ghazvininejad.
\newblock A review on language models as knowledge bases.
\newblock {\em arXiv preprint arXiv:2204.06031}, 2022.

\bibitem{hartmann2023sok}
Valentin Hartmann, Anshuman Suri, Vincent Bindschaedler, David Evans, Shruti Tople, and Robert West.
\newblock Sok: Memorization in general-purpose large language models, 2023.

\bibitem{nasr2023scalable}
Milad Nasr, Nicholas Carlini, Jonathan Hayase, Matthew Jagielski, A~Feder Cooper, Daphne Ippolito, Christopher~A Choquette-Choo, Eric Wallace, Florian Tram{\`e}r, and Katherine Lee.
\newblock Scalable extraction of training data from (production) language models.
\newblock {\em arXiv preprint arXiv:2311.17035}, 2023.

\bibitem{brown2022does}
Hannah Brown, Katherine Lee, Fatemehsadat Mireshghallah, Reza Shokri, and Florian Tram{\`e}r.
\newblock What does it mean for a language model to preserve privacy?
\newblock In {\em Proceedings of the 2022 ACM conference on fairness, accountability, and transparency}, pages 2280--2292, 2022.

\bibitem{schapire2013explaining}
Robert~E Schapire.
\newblock Explaining adaboost.
\newblock In {\em Empirical inference: festschrift in honor of vladimir N. Vapnik}, pages 37--52. Springer, 2013.

\bibitem{wyner2017explaining}
Abraham~J Wyner, Matthew Olson, Justin Bleich, and David Mease.
\newblock Explaining the success of adaboost and random forests as interpolating classifiers.
\newblock {\em Journal of Machine Learning Research}, 18(48):1--33, 2017.

\bibitem{feldman2020does}
Vitaly Feldman.
\newblock Does learning require memorization? a short tale about a long tail.
\newblock In {\em Proceedings of the 52nd Annual ACM SIGACT Symposium on Theory of Computing}, pages 954--959, 2020.

\bibitem{mireshghallah2020privacy}
Fatemehsadat Mireshghallah, Mohammadkazem Taram, Praneeth Vepakomma, Abhishek Singh, Ramesh Raskar, and Hadi Esmaeilzadeh.
\newblock Privacy in deep learning: A survey.
\newblock {\em arXiv preprint arXiv:2004.12254}, 2020.

\bibitem{yao2024survey}
Yifan Yao, Jinhao Duan, Kaidi Xu, Yuanfang Cai, Zhibo Sun, and Yue Zhang.
\newblock A survey on large language model (llm) security and privacy: The good, the bad, and the ugly.
\newblock {\em High-Confidence Computing}, page 100211, 2024.

\bibitem{dwork2014algorithmic}
Cynthia Dwork and Aaron Roth.
\newblock The algorithmic foundations of differential privacy.
\newblock {\em Foundations and Trends{\textregistered} in Theoretical Computer Science}, 9(3--4):211--407, 2014.

\bibitem{he2020tighter}
Fengxiang He, Bohan Wang, and Dacheng Tao.
\newblock Tighter generalization bounds for iterative differentially private learning algorithms.
\newblock {\em arXiv preprint arXiv:2007.09371}, 2020.

\bibitem{dwork2015preserving}
Cynthia Dwork, Vitaly Feldman, Moritz Hardt, Toniann Pitassi, Omer Reingold, and Aaron~Leon Roth.
\newblock Preserving statistical validity in adaptive data analysis.
\newblock In {\em Annual ACM Symposium on Theory of Computing}, pages 117--126, 2015.

\bibitem{nissim2015generalization}
Kobbi Nissim and Uri Stemmer.
\newblock On the generalization properties of differential privacy.
\newblock {\em CoRR, abs/1504.05800}, 2015.

\bibitem{oneto2017differential}
Luca Oneto, Sandro Ridella, and Davide Anguita.
\newblock Differential privacy and generalization: Sharper bounds with applications.
\newblock {\em Pattern Recognition Letters}, 89:31--38, 2017.

\bibitem{song2013stochastic}
Shuang Song, Kamalika Chaudhuri, and Anand~D Sarwate.
\newblock Stochastic gradient descent with differentially private updates.
\newblock In {\em 2013 IEEE global conference on signal and information processing}, pages 245--248. IEEE, 2013.

\bibitem{abadi2016deep}
Martin Abadi, Andy Chu, Ian Goodfellow, H~Brendan McMahan, Ilya Mironov, Kunal Talwar, and Li~Zhang.
\newblock Deep learning with differential privacy.
\newblock In {\em ACM SIGSAC Conference on Computer and Communications Security}, pages 308--318, 2016.

\bibitem{yu2021differentially}
Da~Yu, Saurabh Naik, Arturs Backurs, Sivakanth Gopi, Huseyin~A Inan, Gautam Kamath, Janardhan Kulkarni, Yin~Tat Lee, Andre Manoel, Lukas Wutschitz, et~al.
\newblock Differentially private fine-tuning of language models.
\newblock {\em arXiv preprint arXiv:2110.06500}, 2021.

\bibitem{ding2024delving}
Youlong Ding, Xueyang Wu, Yining Meng, Yonggang Luo, Hao Wang, and Weike Pan.
\newblock Delving into differentially private transformer.
\newblock {\em arXiv preprint arXiv:2405.18194}, 2024.

\bibitem{zhang2024dpzero}
Liang Zhang, Bingcong Li, Kiran~Koshy Thekumparampil, Sewoong Oh, and Niao He.
\newblock {DPZ}ero: Private fine-tuning of language models without backpropagation.
\newblock In {\em Forty-first International Conference on Machine Learning}, 2024.

\bibitem{ma2022dimension}
Yi-An Ma, Teodor~Vanislavov Marinov, and Tong Zhang.
\newblock Dimension independent generalization of dp-sgd for overparameterized smooth convex optimization.
\newblock {\em arXiv preprint arXiv:2206.01836}, 2022.

\bibitem{li2022does}
Xuechen Li, Daogao Liu, Tatsunori~B Hashimoto, Huseyin~A Inan, Janardhan Kulkarni, Yin-Tat Lee, and Abhradeep Guha~Thakurta.
\newblock When does differentially private learning not suffer in high dimensions?
\newblock {\em Advances in Neural Information Processing Systems}, 35:28616--28630, 2022.

\bibitem{malladi2023fine}
Sadhika Malladi, Tianyu Gao, Eshaan Nichani, Alex Damian, Jason~D Lee, Danqi Chen, and Sanjeev Arora.
\newblock Fine-tuning language models with just forward passes.
\newblock {\em Advances in Neural Information Processing Systems}, 36:53038--53075, 2023.

\bibitem{bagdasaryan2019differential}
Eugene Bagdasaryan, Omid Poursaeed, and Vitaly Shmatikov.
\newblock Differential privacy has disparate impact on model accuracy.
\newblock {\em Advances in neural information processing systems}, 32, 2019.

\bibitem{amin2019bias}
Kareem Amin, Alex Kulesza, Andres Munoz, and Sergei Vassilvtiskii.
\newblock Bounding user contributions: A bias-variance trade-off in differential privacy.
\newblock In {\em International Conference on Machine Learning}, pages 263--271. PMLR, 2019.

\bibitem{xu2021removing}
Depeng Xu, Wei Du, and Xintao Wu.
\newblock Removing disparate impact on model accuracy in differentially private stochastic gradient descent.
\newblock In {\em Proceedings of the 27th ACM SIGKDD Conference on Knowledge Discovery \& Data Mining}, pages 1924--1932, 2021.

\bibitem{zhang2023chatgpt}
Jizhi Zhang, Keqin Bao, Yang Zhang, Wenjie Wang, Fuli Feng, and Xiangnan He.
\newblock Is chatgpt fair for recommendation? evaluating fairness in large language model recommendation.
\newblock In {\em Proceedings of the 17th ACM Conference on Recommender Systems}, pages 993--999, 2023.

\bibitem{li2023survey}
Yingji Li, Mengnan Du, Rui Song, Xin Wang, and Ying Wang.
\newblock A survey on fairness in large language models.
\newblock {\em arXiv preprint arXiv:2308.10149}, 2023.

\bibitem{chu2024fairness}
Zhibo Chu, Zichong Wang, and Wenbin Zhang.
\newblock Fairness in large language models: A taxonomic survey.
\newblock {\em ACM SIGKDD explorations newsletter}, 26(1):34--48, 2024.

\bibitem{gallegos2024bias}
Isabel~O Gallegos, Ryan~A Rossi, Joe Barrow, Md~Mehrab Tanjim, Sungchul Kim, Franck Dernoncourt, Tong Yu, Ruiyi Zhang, and Nesreen~K Ahmed.
\newblock Bias and fairness in large language models: A survey.
\newblock {\em Computational Linguistics}, pages 1--79, 2024.

\bibitem{dai2024bias}
Sunhao Dai, Chen Xu, Shicheng Xu, Liang Pang, Zhenhua Dong, and Jun Xu.
\newblock Bias and unfairness in information retrieval systems: New challenges in the llm era.
\newblock In {\em Proceedings of the 30th ACM SIGKDD Conference on Knowledge Discovery and Data Mining}, pages 6437--6447, 2024.

\bibitem{calders2009building}
Toon Calders, Faisal Kamiran, and Mykola Pechenizkiy.
\newblock Building classifiers with independency constraints.
\newblock In {\em IEEE International Conference on Data Mining Workshops}, pages 13--18, 2009.

\bibitem{hardt2016equality}
Moritz Hardt, Eric Price, and Nati Srebro.
\newblock Equality of opportunity in supervised learning.
\newblock In {\em Advances in Neural Information Processing Systems}, pages 3315--3323, 2016.

\bibitem{singh2019fairness}
Harvineet Singh, Rina Singh, Vishwali Mhasawade, and Rumi Chunara.
\newblock Fairness violations and mitigation under covariate shift.
\newblock {\em arXiv preprint arXiv:1911.00677}, 2019.

\bibitem{chen2022fairness}
Yatong Chen, Reilly Raab, Jialu Wang, and Yang Liu.
\newblock Fairness transferability subject to bounded distribution shift.
\newblock {\em arXiv preprint arXiv:2206.00129}, 2022.

\bibitem{pham2023fairness}
Thai-Hoang Pham, Xueru Zhang, and Ping Zhang.
\newblock Fairness and accuracy under domain generalization.
\newblock {\em arXiv preprint arXiv:2301.13323}, 2023.

\bibitem{yoon2020joint}
Taeho Yoon, Jaewook Lee, and Woojin Lee.
\newblock Joint transfer of model knowledge and fairness over domains using wasserstein distance.
\newblock {\em IEEE Access}, 8:123783--123798, 2020.

\bibitem{endres2003new}
Dominik~Maria Endres and Johannes~E Schindelin.
\newblock A new metric for probability distributions.
\newblock {\em IEEE Transactions on Information theory}, 49(7):1858--1860, 2003.

\bibitem{barikeri2021redditbias}
Soumya Barikeri, Anne Lauscher, Ivan Vuli{\'c}, and Goran Glava{\v{s}}.
\newblock Redditbias: A real-world resource for bias evaluation and debiasing of conversational language models.
\newblock {\em arXiv preprint arXiv:2106.03521}, 2021.

\bibitem{rawte2023survey}
Vipula Rawte, Amit Sheth, and Amitava Das.
\newblock A survey of hallucination in large foundation models.
\newblock {\em arXiv preprint arXiv:2309.05922}, 2023.

\bibitem{ji2022survey}
Ziwei Ji, Nayeon Lee, Rita Frieske, Tiezheng Yu, Dan Su, Yan Xu, Etsuko Ishii, Yejin Bang, Delong Chen, Ho~Shu Chan, Wenliang Dai, Andrea Madotto, and Pascale Fung.
\newblock Survey of hallucination in natural language generation.
\newblock {\em arXiv preprint arXiv:2202.03629v6}, 2022.

\bibitem{liu2024survey}
Hanchao Liu, Wenyuan Xue, Yifei Chen, Dapeng Chen, Xiutian Zhao, Ke~Wang, Liping Hou, Rongjun Li, and Wei Peng.
\newblock A survey on hallucination in large vision-language models.
\newblock {\em arXiv preprint arXiv:2402.00253}, 2024.

\bibitem{dhuliawala2023chain}
Shehzaad Dhuliawala, Mojtaba Komeili, Jing Xu, Roberta Raileanu, Xian Li, Asli Celikyilmaz, and Jason Weston.
\newblock Chain-of-verification reduces hallucination in large language models, 2023.

\bibitem{huang2024opera}
Qidong Huang, Xiaoyi Dong, Pan Zhang, Bin Wang, Conghui He, Jiaqi Wang, Dahua Lin, Weiming Zhang, and Nenghai Yu.
\newblock Opera: Alleviating hallucination in multi-modal large language models via over-trust penalty and retrospection-allocation.
\newblock In {\em Proceedings of the IEEE/CVF Conference on Computer Vision and Pattern Recognition}, pages 13418--13427, 2024.

\bibitem{xu2024hallucination}
Ziwei Xu, Sanjay Jain, and Mohan Kankanhalli.
\newblock Hallucination is inevitable: An innate limitation of large language models.
\newblock {\em arXiv preprint arXiv:2401.11817}, 2024.

\bibitem{chen2024inside}
Chao Chen, Kai Liu, Ze~Chen, Yi~Gu, Yue Wu, Mingyuan Tao, Zhihang Fu, and Jieping Ye.
\newblock Inside: Llms' internal states retain the power of hallucination detection, 2024.

\bibitem{chuang2023dola}
Yung-Sung Chuang, Yujia Xie, Hongyin Luo, Yoon Kim, James Glass, and Pengcheng He.
\newblock Dola: Decoding by contrasting layers improves factuality in large language models.
\newblock {\em arXiv preprint arXiv:2309.03883}, 2023.

\bibitem{zheng2024novo}
Zheng~Yi Ho, Siyuan Liang, Sen Zhang, Yibing Zhan, and Dacheng Tao.
\newblock Novo: Norm voting off hallucinations with attention heads in large language models.
\newblock {\em arXiv preprint arXiv:2410.08970}, 2024.

\bibitem{wang2022self}
Xuezhi Wang, Jason Wei, Dale Schuurmans, Quoc Le, Ed~Chi, Sharan Narang, Aakanksha Chowdhery, and Denny Zhou.
\newblock Self-consistency improves chain of thought reasoning in language models.
\newblock {\em arXiv preprint arXiv:2203.11171}, 2022.

\bibitem{shi2022natural}
Freda Shi, Daniel Fried, Marjan Ghazvininejad, Luke Zettlemoyer, and Sida~I. Wang.
\newblock Natural language to code translation with execution, 2022.

\bibitem{cohen2023lm}
Roi Cohen, May Hamri, Mor Geva, and Amir Globerson.
\newblock Lm vs lm: Detecting factual errors via cross examination.
\newblock {\em arXiv preprint arXiv:2305.13281}, 2023.

\bibitem{lotfinon}
Sanae Lotfi, Marc~Anton Finzi, Yilun Kuang, Tim~GJ Rudner, Micah Goldblum, and Andrew~Gordon Wilson.
\newblock Non-vacuous generalization bounds for large language models.
\newblock In {\em Forty-first International Conference on Machine Learning}, 2024.

\bibitem{Tay_scaling}
Yi~Tay, Mostafa Dehghani, Samira Abnar, Hyung~Won Chung, William Fedus, Jinfeng Rao, Sharan Narang, Vinh~Q. Tran, Dani Yogatama, and Donald Metzler.
\newblock Scaling laws vs model architectures: How does inductive bias influence scaling.
\newblock {\em arXiv preprint arXiv:2207.10551}, 2022.

\bibitem{niu2024beyond}
Xueyan Niu, Bo~Bai, Lei Deng, and Wei Han.
\newblock Beyond scaling laws: Understanding transformer performance with associative memory.
\newblock {\em arXiv preprint arXiv:2405.08707}, 2024.

\end{thebibliography}
\end{document}